# AN INTEGRATED SOFT COMPUTING APPROACH TO A MULTI-BIOMETRIC SECURITY MODEL

**A THESIS**

**SUBMITTED TO**

## DAYALBAGH EDUCATIONAL INSTITUTE

**TOWARDS FULFILMENT OF THE REQUIREMENTS**

**FOR THE AWARD OF**

## DOCTOR OF PHILOSOPHY

**IN**

## PHYSICS AND COMPUTER SCIENCE

**BY**

## PREM SEWAK SUDHISH

**DEPARTMENT OF PHYSICS AND COMPUTER SCIENCE**

**DAYALBAGH EDUCATIONAL INSTITUTE**

**2015**

**THE THESIS HAS BEEN FORMATTED FOR DUPLEX PRINTING.**

# DEDICATION

**To**

**The shared vision and values**

**and the beatific affinity between**

**Dayalbagh Educational Institute**

**and**

**Michigan State University**

# ACKNOWLEDGMENT

*My humble obeisances before Thee, my Father, Mentor and Friend,*

*It is utterly impossible, for an iota of Thy infinite blessings to be penned;*

*Forgive my imprudence, oh Lord, for making this puerile attempt,*

*To bind the unbounded with vocabulary, I hope it is not contempt;*

*Individuals and opportunities facilitating this research, that I may call mine,*

*Are all Thy agents and instruments, in Thy assiduously orchestrated design;*

*With deepest gratitude towards my co-travelers, I offer this heartfelt prayer,*

*Forever and forever may we abide by Thee, under Thy protection and care;*

*Be at Dayalbagh Educational Institute or at Michigan State University,*

*Wherever ordained by Thee, may we serve Thy divine mission earnestly;*

*Supervisors, collaborators, colleagues, administrators – whether past or present,*

*Reviewers, well-wishers and family, grant us Thy boon, may we all transcend.*

***– Prem Sewak Sudhish***

# TABLE OF CONTENTS

# LIST OF TABLES

# LIST OF FIGURES

# ABSTRACT

## Motivation

The application of biometrics for human identification has multiplied manifold over the past decade, especially in large-scale national identification applications, both for security and social welfare initiatives. The substantial increase in scale and diversity of population that needs to be covered by biometrics has made single instances of biometric evidence (e.g., fingerprints) to be inadequate. Large scale biometric systems need to rely on multi-modal biometric techniques that fuse together multiple biometric traits (e.g., fingerprint and iris) to establish uniqueness of identity of an individual.

Several applications of person recognition in the civil and criminal domains have scenarios where users may have a motivation to claim a duplicate or false identity. Multi-modal biometrics are often used to detect such multiple or false instances of individual identity, known as the process of de-duplication. While offering several advantages including improvement in the overall accuracy, the use of multi-modal biometrics causes significant increase in computational overheads because of the requirement to compute matching scores for each trait followed by fusion. The accuracy of multi-modal biometric systems has extensively been studied but the computational efficiency of these systems has not been investigated scientifically.

The research addresses the following objectives outlined in the synopsis, towards creating an integrated soft computing based multi-biometric system:

- Mere usage of multiple biometrics does not necessarily imply better system performance; a poorly designed multi-biometric system can result in deterioration in performance of the individual modalities, increase the cost of the system, and present increased inconvenience to users/administrators (e.g., complex enrollment procedures). It is important, therefore, to consider the development of a model for a biometric system that would not only be robust and optimal but also efficient, through an appropriate and judicious combination of existing biometrics.
- While soft computing techniques have largely been applied in the field of biometrics for validation and identification (pattern recognition), it is possible to employ soft computing techniques in other stages of operation of biometric systems as well, and this line of study opens up new vistas for research.
- Search for individuals often include large databases containing traditional identifiers such as name, address, age, gender, etc. Even though several combinations of soft biometric traits such as hair color, eye color, etc. have been explored, a search based on fusion of biometric traits with traditional identifiers may lead to added robustness in methods for searching individuals.

## Chapter Organization

The first chapter of the thesis introduces the person recognition problem, along with several applications and the verification and identification recognition scenarios. A description of the several types of recognition systems, including biographical, possession based, knowledge based, biometric systems and multi-factor recognition systems has been provided. The fake identity problem, along with prior attempts at de-duplication of identities by combining biometric and biographical information is discussed, concluding with the contributions presented in the thesis.

The second chapter presents an introduction to biometric systems, describing the desired characteristics of biometric traits and a brief description of some commonly used biometric traits. The sensing, quality estimation, enhancement and segmentation, feature extraction, matching and decision, and database modules that constitute unimodal biometric systems have been described. Several limitations presented by unimodal biometric systems are overcome by multi-biometric systems but these systems have their own sets of challenges, requiring several pieces of information to be fused together. Multi-biometric systems may be of different types, such as multi-sensor, multi-algorithm, multi-instance, multi-sample, multi-modal or hybrid systems and the fusion of information in these systems may be at different levels, e.g., the sensor level, feature level, score level, rank level or decision level. It is well accepted that fusion at the score level presents the best trade-off between the amount of information and the complexity of fusion. The fusion strategy may be based on transformation and combination of scores, or may be based on the probability densities or may employ a classifier. The dynamic score selection strategy and quality based fusion have also been described. The chapter ends with a discussion on the various metrics for evaluating and comparing biometric systems and the state of the art in unimodal and multi-modal biometric recognition.

The third chapter, while introducing various soft computing approaches, e.g. logistic regression, support vector machines, neural networks, evolutionary computing and fuzzy logic, also presents the motivation why biometric recognition is an ideal candidate for soft computing approaches. Soft computing approaches may also be combined together to create hybrid systems. These model selection and model tuning in soft computing approaches is generally driven by the available data and the evaluation of these approaches consists of two distinct phases called training and testing. The standard best practices for training and testing of models have been described in this chapter, along with bootstrap aggregation as a meta-algorithm for improving the robustness of soft computing models.

An integrated approach towards multi-biometric and biographical information using a soft computing model has been presented in the fourth chapter. The application scenarios for the use of de-duplication systems have been described, along with some strategies for matching biometric and biographical information. A new algorithm for matching biographical information has also been proposed. This algorithm is shown to improve the accuracy of matching biographical information by about 4% in comparison to Levenshtein distance on US census data. The design objectives for a universal fusion mechanism for large-scale systems have been enlisted and a bootstrap aggregating ensemble of veto-wielding logistic regression classifiers is presented as a framework for improving the accuracy as well as efficiency of information fusion in de-duplication systems. An analysis of the proposed framework has also been presented in this chapter.

The fifth chapter contains a rigorous experimental analysis of the proposed framework on multiple large-scale benchmark datasets and a comparison with prior attempts demonstrates favorable results for the proposed framework, not only in terms of accuracy and enhanced robustness of the system, but also a reduction in computational requirements.

Some conclusions, along with the learnings from the research are presented in the sixth chapter. This chapter also presents some avenues for future expansion and further research on the work presented in the thesis.

## Salient Contributions

The research presents the following salient contributions:

i. A novel technique has been developed for comparing biographical information, by combining the average impact of Levenshtein, Damerau-Levenshtein, and editor distances. The impact is calculated as the ratio of the edit distance to the maximum possible edit distance between two strings of the same lengths as the given pair of strings. This impact lies in the range [0, 1] and can easily be converted to a similarity (matching) score by subtracting the impact from unity.

ii. A universal soft computing framework is proposed for adaptively fusing biometric and biographical information by making real-time decisions to determine after consideration of each individual identifier whether computation of matching scores and subsequent fusion of additional identifiers, including biographical information is required. This proposed framework not only improves the accuracy of the system by fusing less reliable information (e.g. biographical information) only for instances where such a fusion is required, but also improves the efficiency of the system by computing matching scores for various available identifiers only when this computation is considered necessary.

iii. A scientific method for comparing efficiency of fusion strategies through a predicted effort to error trade-off curve.

# CHAPTER 1

# INTRODUCTION

## 1.1. Person Recognition

Humans are known to efficiently recognize fellow humans they have known or encountered before. In fact, the task of effectively recognizing or authenticating individuals (termed *person recognition*) forms an integral part of routine activities in our daily lives, e.g., to initiate conversations and transactions with the intended individuals. However, in the present era, with millions of transactions being effected automatically via the use of technology, automatic recognition of individuals is important to ensure the authenticity and integrity of transactions. In such a scenario, manual recognition by human operators before every transaction is neither efficient, nor practical.

Historically, several methods have been used for assertion of individual identity or affiliation to a certain group. Many of these methods continue to be used to this day, e.g., possession of certain tokens such as an identity document; knowledge of a secret passphrase; tattoos, body marks or other anatomical characteristics such as fingerprints. Person recognition or authentication is a necessary pre-requisite before granting authorization for use of resources that have controlled or limited access, e.g., a bank account.

Person recognition is required for a variety of applications from surveillance, law enforcement, physical and logical access control to time and attendance systems, mobile user authentication and social welfare programs. Some of these applications are discussed in section 1.2. These applications present different scenarios for person recognition, which are discussed in section 1.3. The mechanisms and systems used for person recognition are described in section 1.4 while section 1.5 discusses systems that combine these mechanisms. Certain applications have incentives and motivation for individuals to falsify their true identity and to assume duplicate or multiple identities. This problem is examined in section 1.6. Some recent attempts for resolution of the duplicate identity problem are presented and compared in section 1.7. The contributions being presented through this thesis conclude the chapter in section 1.8.

## 1.2. Applications

The applications of person recognition may be categorized into several domains, some examples of which are presented below.

### 1.2.1. Civil and Social Welfare

It is estimated that about 10 million out of the 26 million annual childbirths in India are unregistered [1]. The absence of such legal identity not only makes the child more susceptible to child trafficking and child labor, but also leads to denial of certain other privileges later as an adult, such as a formal job, bank account, driving license, marriage certificate or a passport. For such civil and social welfare applications, it is required that a person be recognized accurately before providing them access to certain benefits and resources. Examples of such applications include disbursal of monetary benefits and subsidies to certain individuals, and recognition of candidates appearing for a large scale recruitment examination.

Reliable person recognition is also essential for non-cash financial transactions that are primarily accomplished through electronic means. It is estimated that 365.6 billion global non-cash transactions occurred during 2013 [2]. Moreover, the number of retail banking customers who use internet and mobile banking channels weekly has risen to 61.5% and 30.5%, respectively, in 2015 [3].

Another application of interest involves recognition of children for tracking their vaccination schedules. This is important especially in the developing economies to increase vaccination coverage as well as to prevent corrupt practices involving false claims of the number of children with unverifiable identities vaccinated [4].

### 1.2.2. Crime and Law Enforcement

Identifying the person of interest is of critical importance for law enforcement agencies and the judiciary, e.g., to shortlist and apprehend suspects and to ensure appropriate penalties in the case of repeat offenders. A majority of crimes do not result in a conviction because of failure to indisputably recognize the perpetrator of the crime. National Crime Records Bureau’s statistics indicate that the conviction rate for crimes committed in India in 2013 was only about 25% for most crimes [5]. Internationally, on an average, only one offender is found for every two crimes registered [6].

### 1.2.3. Border Protection

An estimated 6.968 million international tourists travel to India every year [7, 8]. The United States has 69.768 million tourists annually [8], and including international visitors for other purposes, the U.S. Customs and Border Patrol screens about a million travelers each day [9]. Therefore, border protection is another important application for person recognition. Individuals crossing international borders have to be reliably and uniquely identified before being granted entry. This is required for protecting the borders against entry of unwanted individuals.

### 1.2.4. Personal

The number of mobile devices exceeds the population of the world [10]. For many users, the sensitive and personal information accessible through these devices is far more valuable than just the explicit cost of the device.

The Automated Search Facility-Stolen Motor Vehicles (ASF-SMV) database of the INTERPOL had more than 6.8 million records of stolen vehicles at the end of 2014 [11]. A suitable and accurate authentication of the driver is likely to prevent several of these thefts.

Several personal applications requiring person recognition span over a multiplicity of personal assets that may require access, e.g. the place of residence, automobile or data stored on a laptop computer or mobile phone.

## 1.3. Recognition Scenarios

The aforementioned applications primarily operate in two different scenarios, verification and identification.

### 1.3.1. Verification

When an identity claimed by an individual needs to be verified, the scenario is termed as verification. To verify the identity, typically, the credentials presented by the individual are matched against the known credentials of the identity being claimed. An example of the verification scenario is border protection, when a person presents immigration documents at the port of entry, claiming a certain identity, and the immigration officer verifies the claim.

The outcome in the verification scenario is generally a binary response, indicating whether the claim of identity has been confirmed or declined.

### 1.3.2. Identification

Certain applications, on the other hand, require that the person be recognized using their credentials, without them explicitly claiming an identity. This is termed identification. In the identification scenario, the credentials presented by, or obtained from the person being recognized are compared against all known credentials available in the system to ascertain their identity. An example of the identification scenario is recognition of a perpetrator of crime from a CCTV footage from the scene of crime by comparing the obtained face image to known mugshot face images.

The outcome of the identification scenario is generally a list of identities, usually ranked by the likelihood of that being the identity of the person being recognized.

The identification problem could be either open or closed set. In *closed set* identification, it is assumed that the identities of all potential users are already known to the system, e.g., an automated attendance management system, while in *open set* identification, a potential user of the system may or may not already be known to the system. For example, at the time of first crime, a lawbreaker would not already be known to a system that recognizes offenders.

The identification scenario is further categorized as positive and negative identification scenario.

#### 1.3.2.1. Positive Identification

The identification scenario is said to be positive identification if the purpose of identification is to determine the privileges or authorization *granted* to the person, e.g. being allowed access to a lecture theatre. In the positive identification scenario, it is expected that the users are willing to be identified, similar to verification, except that no identity needs to be explicitly claimed.

#### 1.3.2.2. Negative Identification

The negative identification scenario is when the purpose of identification is to determine the privileges or authorizations *denied* to them, e.g., screening passengers who are considered to be potential threat from boarding an aircraft [12] or preventing users from assuming multiple identities [13].

## 1.4. Person Recognition Systems

As stated earlier, person recognition systems utilize the credentials of the user to verify the identity claimed by them or to ascertain their identity. These credentials could be based on biographical or textual personally identifiable information; what the user possesses (possession based systems); what the user knows (knowledge based systems);

or, tied directly to their anatomical or behavioral traits (biometric systems). Each of these categories is described below and examples from these categories is shown in Fig. 1.1.

| PART NO | Serial No | Elector Name | Relative Name |
|---|---|---|---|
| 106 | 1186 | SWADEEP SINGH | PRITPAL SINGH |
| 235 | 713 | SWADESH BALA | ANIL KUMAR SINGH |
| 239 | 1481 | SWADESH KUMAR GAUTAM | JIVAN KUMAR GAUTAM |
| 95 | 1163 | SWADESH KUMAR SHUKLA | MUNNOO LAL SHUKLA |
| 214 | 259 | SWADESH MITTAL | BISHAN SWARUP MITTAL |
| 208 | 844 | SWADESH UPADHYAY | RAMAN KISHOR UPADHYAY |
| 231 | 1053 | SWAMI DAS KHANNA | LEKHRAJ KHANNA |
| 233 | 335 | Swami Dayal Satsangi | Kunwar Lal |
| 225 | 941 | SWAMI DAYAL SINGH | JAGJIT SINGH |
| 104 | 1384 | SWAMI PRASAD SINGH | VIJAY SINGH |
| 229 | 977 | SWAMI PRASAD SINGH | RAGHUNANDAN PRATAP SINGH |
| 229 | 883 | Swami Prasad Taneja | Amrit Lal |
| 237 | 37 | SWAMI PYARI | SITARAM SWAMI |
| 235 | 239 | Swami Pyari | Rajapal |
| 209 | 510 | SWAMI PYARI | SITA RAM SWAMI |
| 225 | 926 | SWAMI PYARI | SUBHASH CHAND |
| 225 | 998 | SWAMI PYARI SINHA | AJAY KUMAR SINHA |
| 174 | 1134 | SWAMI RAVINDRA BHARTI | CHANDRABHAN AGARWAL |
| 237 | 1116 | SWAMI SARAN | RAMSAROOP PAL |
| 231 | 1015 | SWAMIDAS SINHA | RAMCHANDRA SINHA |
| 229 | 905 | Swamy Prasad Vinjamuri | Subba Rao Vinjamuri |
| 226 | 80 | Swanti Devguptapu | Prem Sewak Sudhish |
| 261 | 1016 | SWAPAN KUMAR DAS | SUSHIL KUMAR DAS |

**a.**

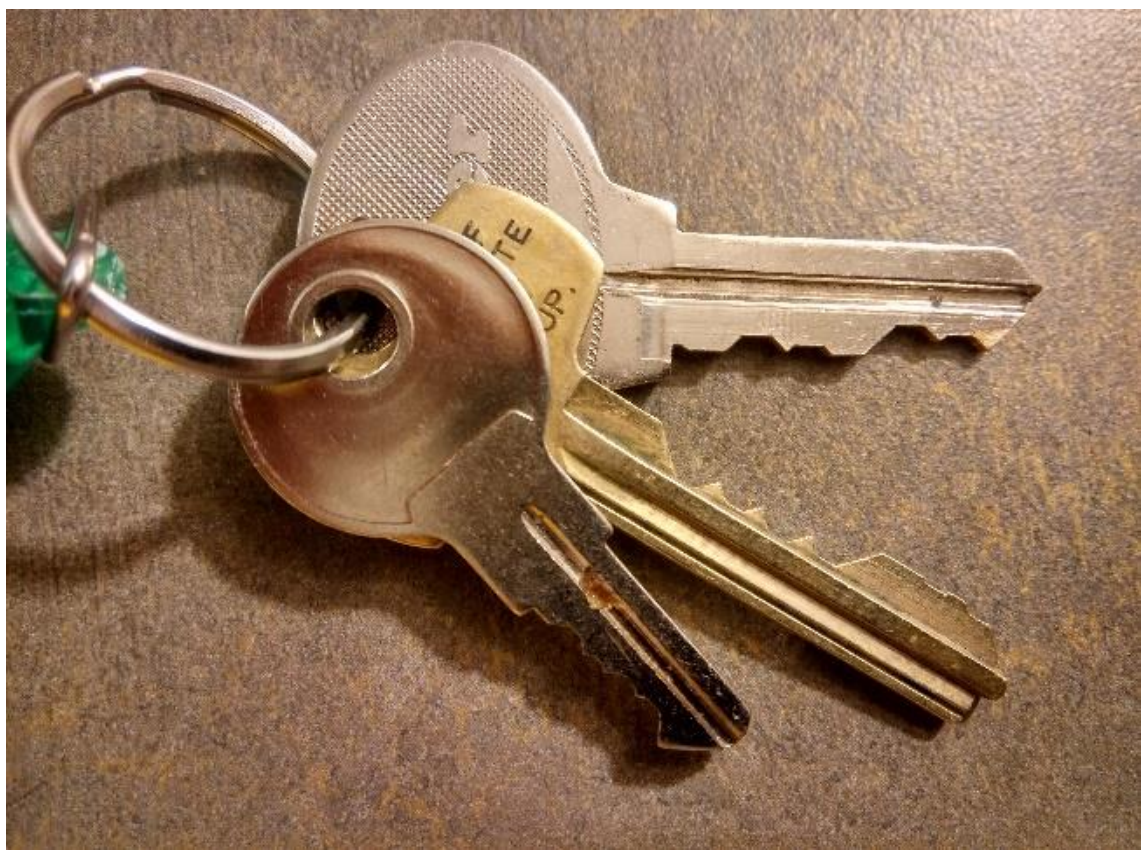

**b.**

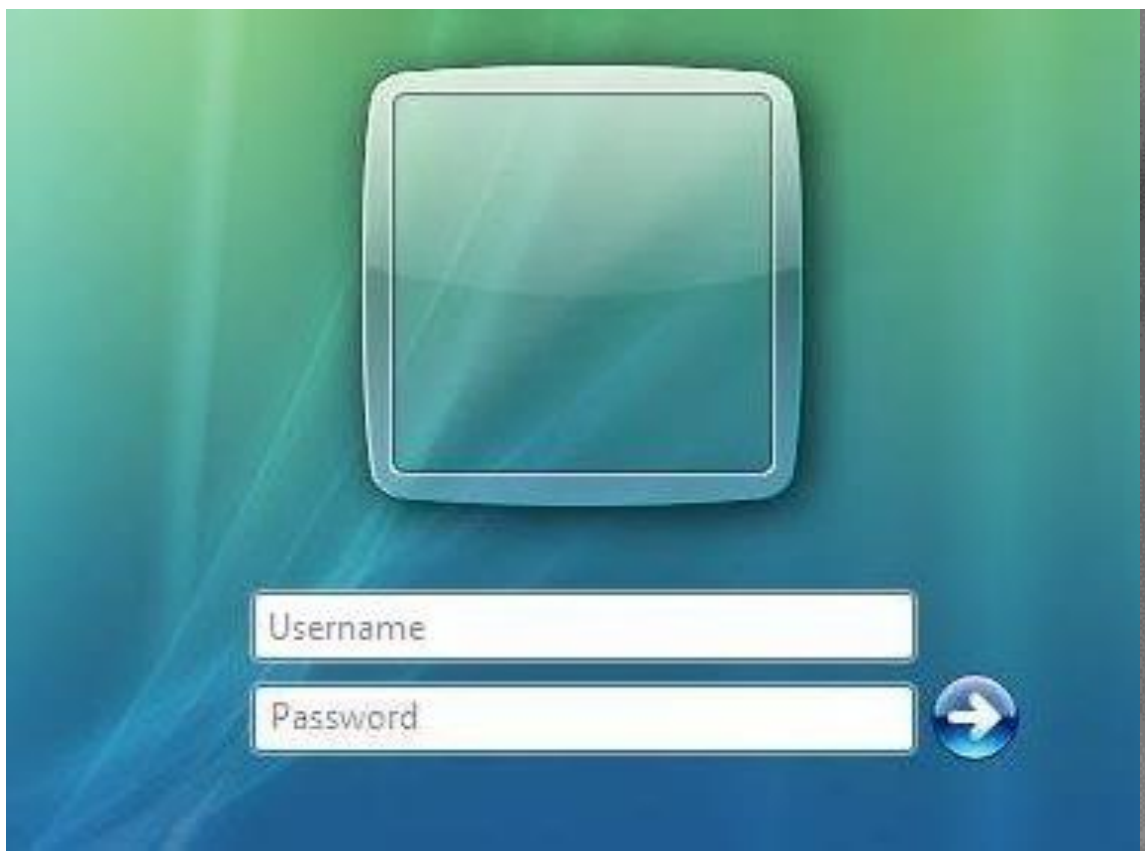

**c.**

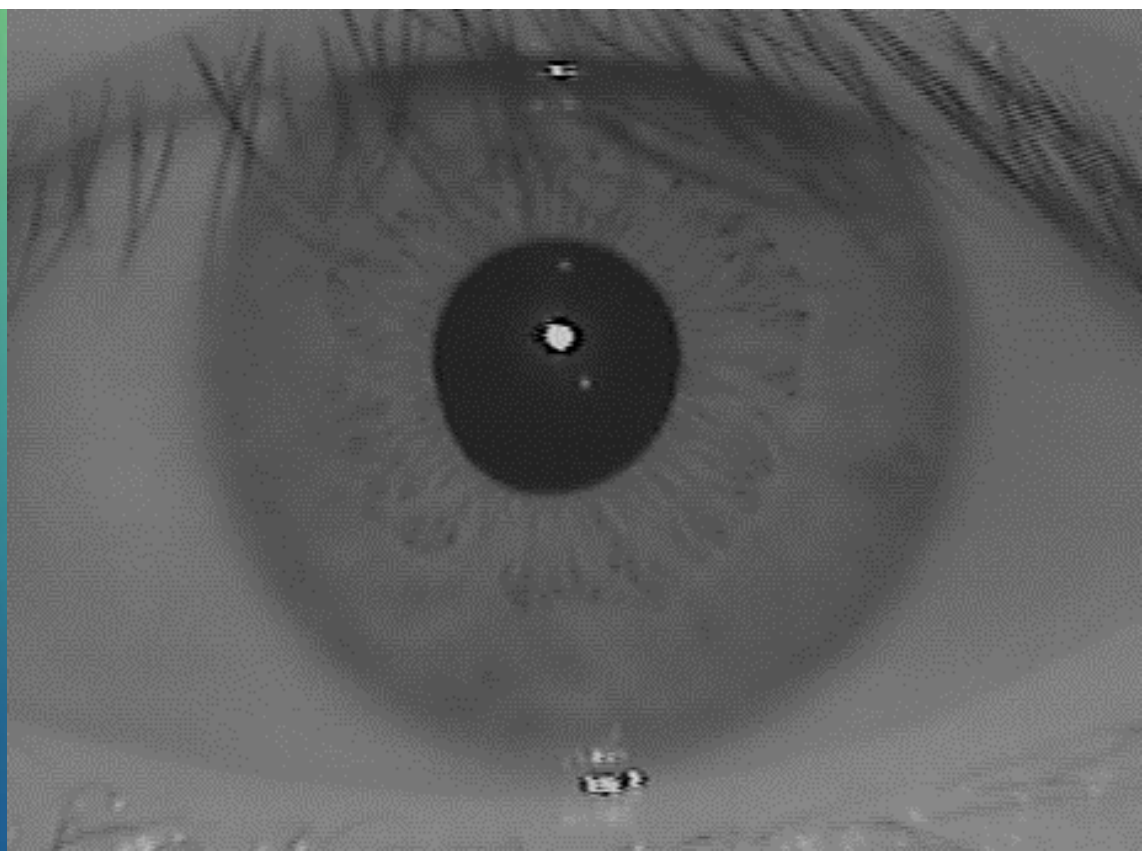

**d.**

**Fig. 1.1 Examples of credentials for person recognition systems.**
**a. Biographical or textual; b. Possession based; c. Knowledge based; d. Biometric**

## 1.4.1. Biographical or Textual Personal Information

Biographical or textual personally identifiable information has traditionally been the most common way of associating identities to individuals. Examples of biographical information include the person’s first and last name, name of their close relatives such as a parent, guardian or spouse, their current and previous residential and office addresses, date of birth or age, mobile phone number and other identifiable information such as their social security number [14].

There are certain limitations of using textual or biographical information as an effective means of person recognition [15, 16], e.g.

i. errors may be introduced by human operators entering information to the system,
ii. lack of a standard format and transliteration leading to multiple ways in which same information may be represented,
iii. lack of uniqueness (e.g. two individuals with the same name),
iv. lack of permanence (e.g. change of address).

Further, in the absence of other documentary evidence or methods of recognition, it is difficult to ascertain whether the biographical or textual personally identifiable information being presented by the user is their own. While an important association to an individual's identity for everyday transactions, this information is not a reliable means for authentication.

### 1.4.2. Possession Based Systems

Possession based systems use a token which the user possesses to establish their individual identity or affiliation to a group. These systems address the recognition problem based on "what you have". Examples of possession based systems include proximity cards or keys to traditional locks.

A notable disadvantage of possession based systems is that the tokens used to authenticate users may be lost, stolen or copied resulting in denial of access to the authorized user, while granting access to an unauthorized user.

### 1.4.3. Knowledge Based Systems

Knowledge based systems utilize a secret piece of information to determine the user's identity. These systems address the recognition problem based on "what you know". Examples of knowledge based systems include systems that use username and password combinations, passphrases, cryptographic keys or personal identification numbers (PIN) to authenticate the users.

A disadvantage of using knowledge based systems is that the secret piece of information or credential should be sufficiently complex to prevent unauthorized access by an impostor through guesswork or a brute-force search, while the genuine user is expected to remember this complex piece of information for authentication.

### 1.4.4. Biometric Systems

Biometric systems address the person recognition problem based on "who you are" instead of "what you know" or "what you possess". Biometric systems are based on the measurement of physiological, behavioral or cognitive traits that are expected to be unique and may be used to determine a person's identity. Examples of biometric traits include fingerprint, face and iris. Biometric systems have the following advantages over other categories of person recognition systems [17]:

i. biometric traits cannot easily be circumvented
ii. they cannot be lost, stolen or guessed, and
iii. there is no need for the user of biometric systems to create and recall passwords

Biometric systems, however, suffer from certain limitations. The matching of biometric traits is generally not exact but approximate in nature and no known biometric trait fulfils all the desirable characteristics expected from an ideal trait. A detailed discussion on these ideal characteristics, the commonly used biometric traits, their limitations along the various components of biometric systems has been presented in the next chapter.

## 1.5. Multi–factor Recognition Systems

Although the aforementioned recognition systems can be used for addressing the person recognition problem, some applications (e.g. those involving financial transactions) require a higher level of security. . For such applications, multi-factor recognition systems that combine more than one of these systems to provide an improved level of security are used. The other advantage of using multi-factor authentication systems is that the complimentary nature of each individual authentication system reduces the chances of simultaneous failure of each individual system.

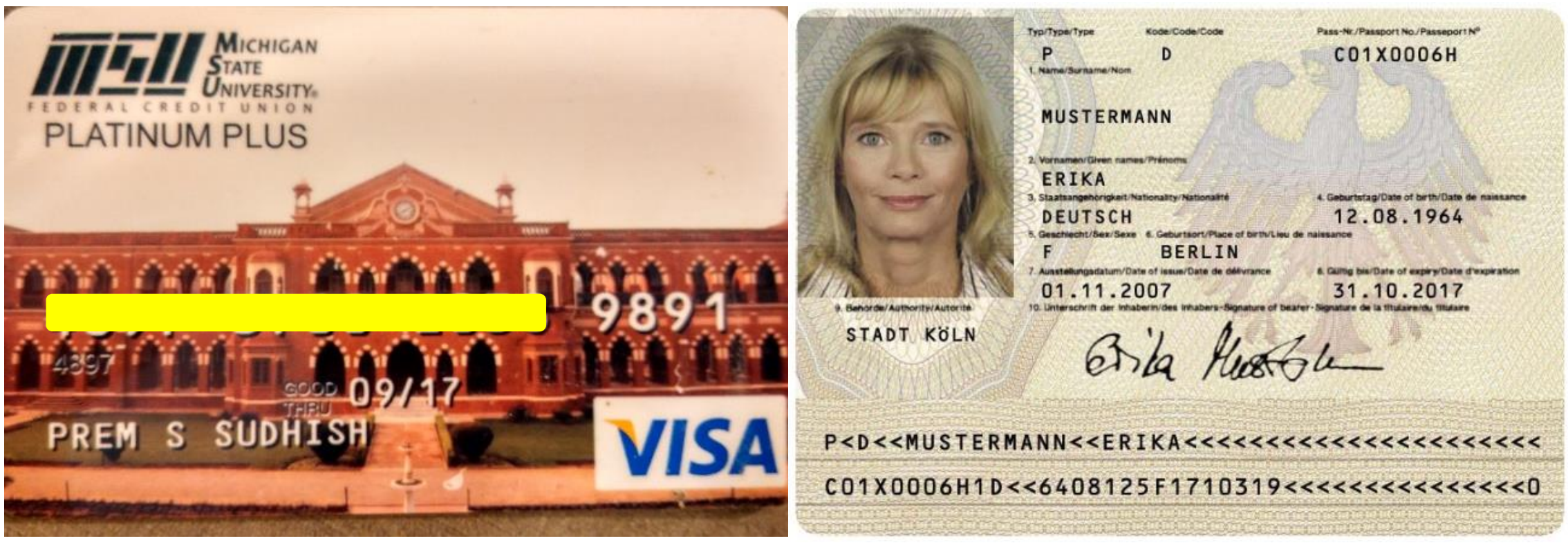

a. b.

**Fig. 1.2 Examples of multi-factor authentication systems.**
**a. Debit card requires possession, along with knowledge of secret key;**
**b. Passport requires possession, along with a biometric match.**

An example of a multi-factor authentication system that requires a combination of possession and knowledge based credentials is as debit cards issued by financial institutions. For withdrawing cash from an automated teller machine (ATM), the user is required to present the card physically to establish possession and enter a secret personal identification number (PIN) to establish knowledge.

Another example is logging on a secure online portal using the user's username and password to establish knowledge and then entering a one-time password (OTP) sent to the user's mobile device to establish possession. These systems may also include biometrics as one of the authentication factors.

An example that also includes biometric credentials as a factor is a user's passport whose possession must be established and the facial image on the passport should match with the appearance of the user.

A few examples of multi-factor authentication systems are shown in Fig. 1.2.

## 1.6. The Duplicate or False Identity Problem

There are several situations where an individual may be motivated to assume more than one identity to circumvent a person recognition system. These multiple identities assumed by a person, whether belonging to another individual or imaginary, are called duplicate or false identities.

A user may assume duplicate or false identities in *border protection* or *criminal* applications, to evade immigration control, law enforcement [18] or to avoid harsher penalties for repeat offenders [19]. In a *civilian* scenario, multiple identities may be used by an individual for tax evasion [20] or for claiming multiple subsidies and other social benefits [21].

The duplicate identity problem, while challenging, is required to be solved effectively by a person recognition system. The process of detecting and removing duplicate identities is commonly referred to as de-duplication.

## 1.7. Review of Identity De-duplication Systems

The primary purpose of using biometrics in large scale applications is to ensure that no individual is able to assume more than one identity. Some prior studies [22, 23] have demonstrated that fusion of biographical information with biometrics can also improve the de-duplication accuracy, although biographical information is not as reliable as biometric identifiers [24].

Although better results can be expected when more information is involved, the de-duplication in large scale applications suffers from complexity of computation even such systems are deployed with the support of high performance parallel computation infrastructure at the backend [25]. For each new query, current methods compare all its biometric and biographical information against the background database and fuse these scores to make a final decision. So, adding even one biometric trait results in a large increase of computational cost and the delays get pronounced with time as more individuals get enrolled on the system. Besides, while the de-duplication accuracy for these identification systems is high, even a small percentage error translates to big absolute numbers on a large population [26], bringing to question the credibility of claims regarding “unique identification” [27]. It is of utmost importance, therefore, for the identification system to not only be robust and optimal but also efficient, through an appropriate and judicious combination of available biometric and biographical information. Examples of data typically collected at the time of enrolment of users in such systems are shown in Fig. 1.3.

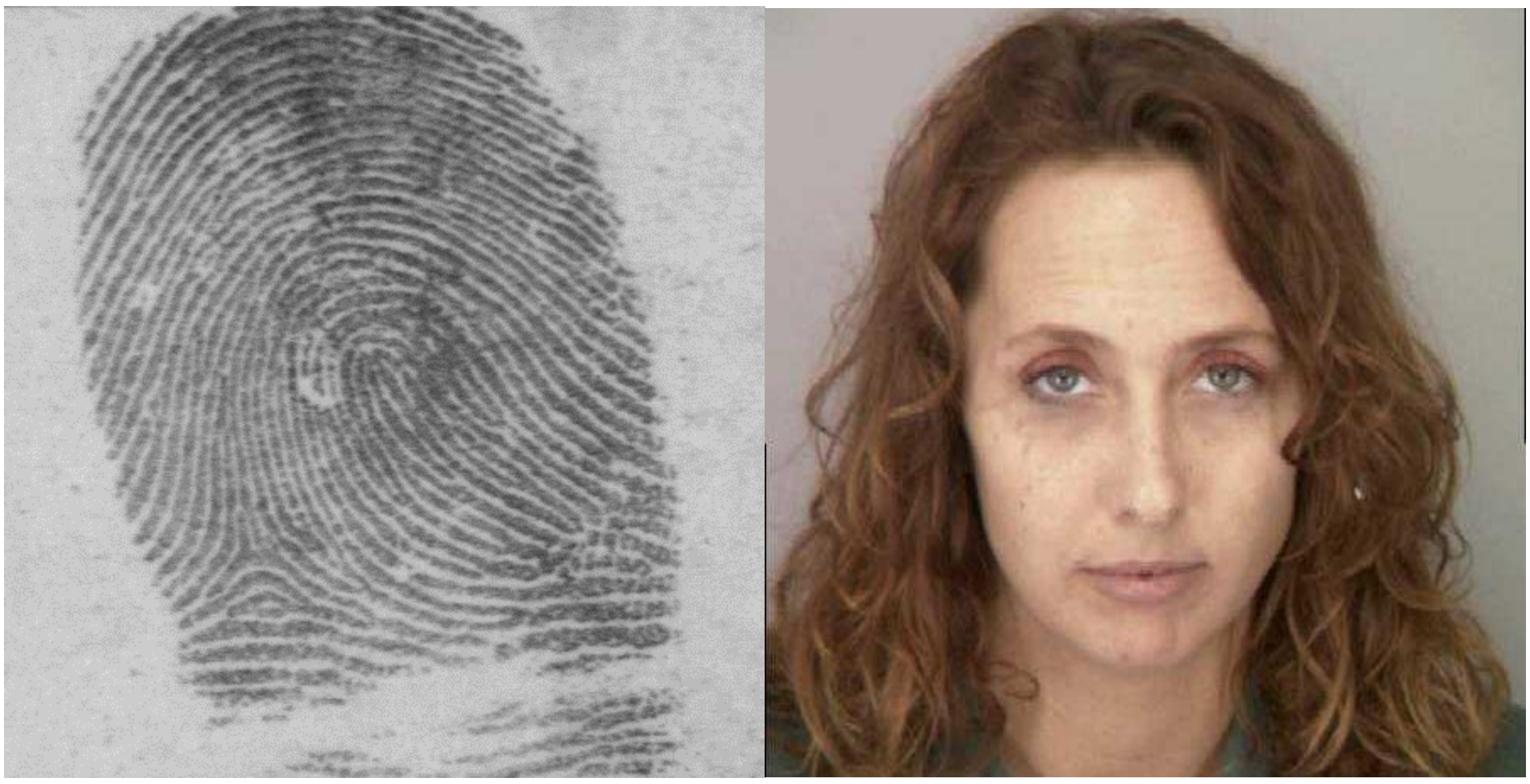

a. b.

**KAROLE ABLEMAN** **ALONSO ABLEMAN**

c. d.

**Fig. 1.3 Examples of biometric and biographical information collected during enrollment.**
**a. Fingerprint [28]; b. Face [29]; c. Name [30]; d. Father's name [30].**

Some commercial systems that fuse biometric and biographical data [31] have been developed but the algorithmic details and results on benchmark datasets for these systems are not publically available. Preliminary studies discussed above [22, 23] have been pioneering in establishing the proof of concept that fusion of biometric and biographical information improves the accuracy of the person recognition system. However, state of the art biometric matchers have not been employed in the experimental evaluation, leading to reported percentage accuracies that are not of practical use in large scale systems. Besides, these studies have also only studied fusion from an accuracy but not from an efficiency perspective. A comparative evaluation of previous attempts at combining biometric and biographical information for person identification, along with results of the present work is presented in Table 1.1.

**Table 1.1 Comparative Summary of Different Works Pertaining to Fusion of Biometric and Biographical Information**

| Study | Target Application | Biometric information used and database | Biographical information and database | Matching algorithm and fusion strategy | Accuracy* | Comments |
|---|---|---|---|---|---|---|
| Bolme et al. [32] | Person (celebrity) identification | 1,331 face images from 118 celebrities (93 men, 25 women) | Textual Information with 400 or more words from celebrity websites | **Biometric**: EBGM for face **Biographical**: Cosine of angle between word frequency vectors **Fusion**: Weighted sum | **Biometric**: 22% **Biographical**: 22% **Fusion**: 35% | Matching scores of all traits are computed and fused for every query. Small database |
| Tyagi et al. [22] | Identity De-duplication | A subset of the fingerprint matching scores from NIST BSSR1 dataset [33] | Name and addresses from an electoral records dataset | **Biometric**: Precomputed fingerprint scores **Biographical**: Matching algorithm not specified **Fusion**: Log-likelihood ratio | **Biometric**: 94.73% **Biographical**: 84.40% **Fusion**: 98.93% | Matching scores of all traits are computed and fused for every query. Small database |
| Bhatt et al. [23] | Identity De-duplication | Fingerprints from 5,734 subjects. Gallery augmented with additional 10K fingerprints. | Name, father's / husband's name, address | **Biometric**: NIST NBIS [34] **Biographical**: Levenshtein distance for string matching **Fusion**: SVM | **Biometric**: 76.6% **Biographical**: 69.4% **Fusion**: 86.5% | Matching scores of all traits are computed and fused for every query. Reported accuracy not adequate for large scale deployment |
| Proposed study | Identity De-duplication | Fingerprints of 27,000 subjects from NIST SD 14 [35] Augmented with face images of 27,000 subjects from the PCSO [29] dataset. | Gender, name and father's name. Names follow the statistics based on the US Census data [30]. The gender is the same as that in the PCSO face dataset. | **Biometric**: COTS** matchers for fingerprint and face. **Biographical**: Combination of Levenshtein [36], Damerau-Levenshtein [37] and editor distances [38]. **Fusion**: Proposed adaptive sequential fusion algorithm. | **Biometric**: 99.64% **Biographical**: 97.47% **Fusion**: 100.0% | Fingerprint alone is adequate for 63.18% of the queries; face required for only for 36.82% of queries; biographical information required only for 8.13% of the queries. |

***Accuracy is the percentage of subjects for whom the true mate is retrieved at rank 1;**
****COTS matcher stands for Commercial Off-the-Shelf matcher.**

## 1.8. Thesis Contributions

The de-duplication of individual identities is of utmost importance in large scale person recognition systems consisting of hundreds of millions of users, such as India's Aadhaar program [39]. De-duplication of identities is also required for other civil and criminal applications and when several identification databases are merged together, e.g. databases from multiple law enforcement agencies [40]. The research embodied in this thesis presents and evaluates a framework for accurate and efficient de-duplication of identities using biometric traits and biographical information.

First, a novel approach towards determination of similarity scores for biographical information is proposed to improve biographical matching.

Second, with the dual objective of speeding up the de-duplication process while also improving the de-duplication accuracy, an algorithm for adaptive fusion of biometric and biographical identification information is proposed. This algorithm offers the following four-fold benefit:

i. A real-time decision is made on whether additional pieces of evidence would be required for fusion, thereby saving computation of matching scores and subsequent fusion for various identifiers;
ii. The robust decision is based on the evaluation of a simple mathematical expression;
iii. The biographical information, which is not very reliable for reasons mentioned earlier (subsection 1.4.1), is considered for fusion only for a small fraction of queries, where the available biometric identifiers are deemed to be insufficient.
iv. In a unique identification scenario, where socio-economic benefits are linked to individual identities, false identities may be created with the connivance of an enrolment operator, causing leakages in incentives provided through social welfare programs. If the operator agrees to capture biometric traits from different individuals for creating a false identity, this algorithm would be significantly superior in flagging that identity as a duplicate when compared to fusion methods that combine matching scores from all available traits.

Experiments on benchmark databases show significant savings in computational effort, e.g., on NIST Special Database 14 [35], combined with mugshot face images from the PCSO database [29] and biographical information derived from US Census [30], it is shown that a unimodal match with only one fingerprint is correctly predicted to be sufficient for 63.18% of the queries, and with the fusion of face scores, no further information is required to be fused for 91.87% of the subjects, without any deterioration in accuracy.

Finally, a predicted effort to error trade-off curve is proposed as a tool for scientifically comparing the efficiency and accuracy of fusion algorithms.

The remainder of this thesis is organized as follows. The next chapter presents an overview of the biometric systems while the third introduces soft computing techniques and their application to biometric systems. The proposal for a soft-computing based integrated security system is described in fourth chapter and the results from evaluation and analysis of the proposed system on benchmark datasets are presented in the fifth chapter. The conclusions are drawn in the sixth chapter, along with proposals for avenues for further research.

# CHAPTER 2

# BIOMETRIC SYSTEMS

## 2.1. Introduction

Biometric systems are increasingly being deployed ubiquitously for automated person recognition. These systems are based on the measurement of purportedly unique anatomical and behavioural traits of an individual. Estimates of future trend indicate an increase in the growth and outreach of these systems with a global spread [41]. The applications of biometric systems range from traditional security applications that include forensics, surveillance, law enforcement, physical and logical access control to the more recent civil applications such as time and attendance systems, mobile user authentication and social welfare programs.

The growing prevalence of biometrics is evident from the large scale deployments at the national level, such as Biometric Identity Management System of the US Department of Homeland Security [42] and the Aadhaar project of the Government of India that facilitates access to benefits and services by providing unique identification to all residents based on biometrics [39]. The International Civil Aviation Organization (ICAO) mandates the use of electronically enabled travel documents with biometric identification capability for passports [43] and visas [44]. Several other biometrics-based identity programs are also being implemented or contemplated in countries such as Brazil [45], Indonesia [46], Norway [47] and Israel [48].

### 2.1.1. Biometric Traits

The human traits that are used for identification using biometrics must possess certain characteristics [49]:

i. *Universality*: Every person in the target population possesses the trait.
ii. *Distinctiveness*: The trait should be sufficiently different between any two individuals or, in other words, unique to each individual.
iii. *Permanence*: The trait should be persistent i.e. sufficiently invariant, over a period of time.
iv. *Collectability*: It should be possible to acquire and perform quantitative measurements on the trait.
v. *Performance*: The trait should be amenable to a high accuracy with a good computational speed.
vi. *Acceptability*: The extent to which the target population is willing to accept the use of a particular biometric trait should be sufficiently high.
vii. *Circumvention*: It should not be possible to fool the biometric person recognition system based on the trait using fraudulent methods.

A practical biometric system should meet the specified accuracy, speed, and resource requirements, be harmless to the users, be accepted by the intended population, and be sufficiently robust to thwart various fraudulent methods and attacks [49].

These characteristics are fulfilled to varying degrees by the traits that have been subdivided as physiological, behavioural and cognitive traits in Fig. 2.1.

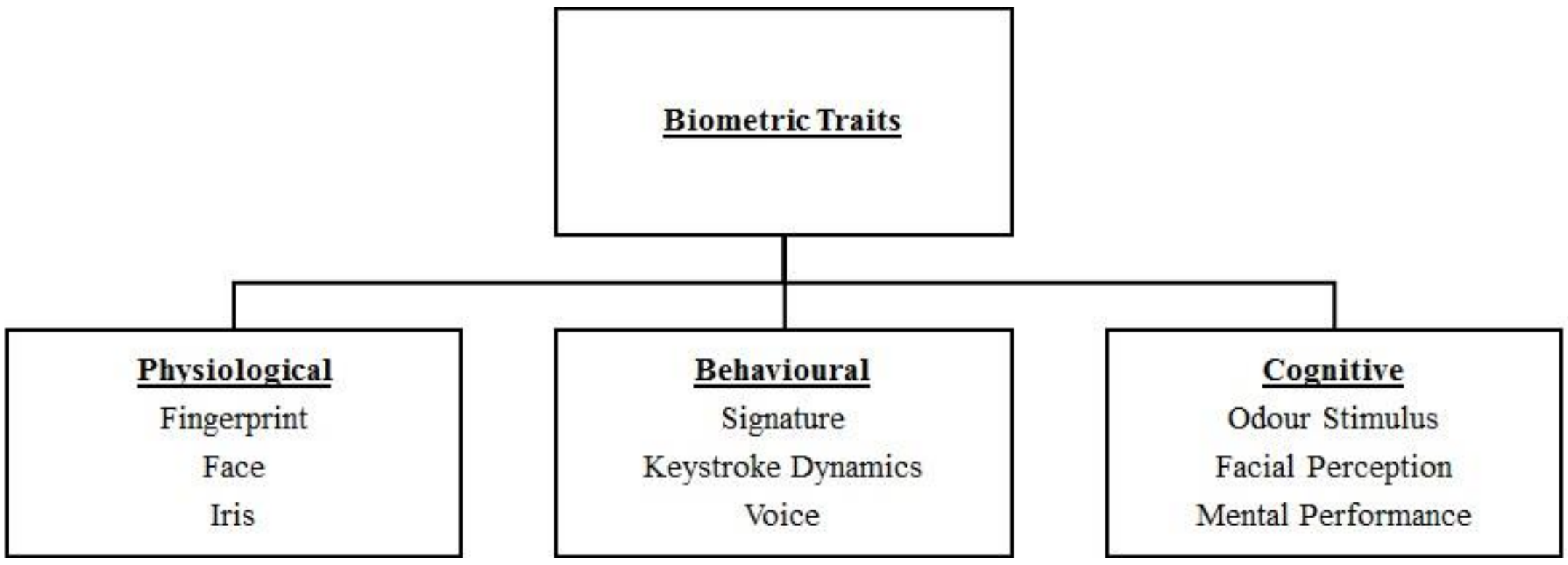

**Fig. 2.1 Categories of biometric traits.**

Biometric traits may also be classified as primary and secondary traits. Primary traits (e.g., fingerprints, face) are commonly used whereas secondary traits (e.g., odour stimulus, mental performance) are not used so frequently.

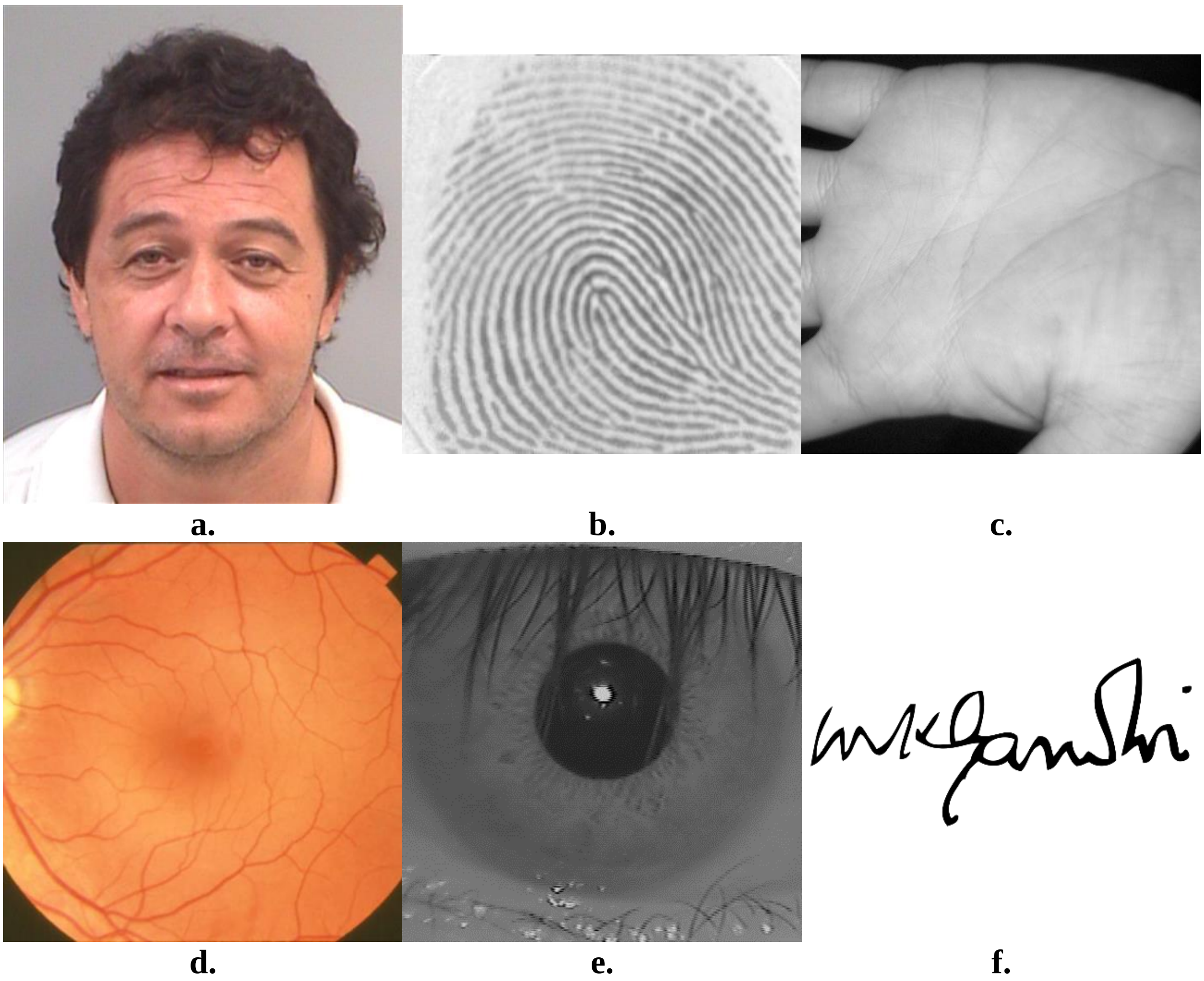

**Fig. 2.2 Examples of some biometric traits.**
**a. Face; b. Fingerprint; c. Palm print; d. Retina; e. Iris; f. Signature**

Some of the most commonly used biometric traits for person recognition are discussed below, and some examples of biometric traits are illustrated in Fig. 2.2.

*Face* is the most common biometric trait used by humans for person recognition. Facial recognition is non – intrusive and the two commonly adopted approaches in automated recognition are based on (i) the location and shape of facial attributes such as eyes, eyebrows, nose, lips and chin, and their spatial relationships, or (ii) the overall (global) analysis of the face image that represents a face as a weighted combination of a number of canonical faces [49].

*Fingerprints* are patterns of ridges and furrows located on the tip of each finger and have been used for person recognition for a long time. Fingerprint recognition is based on features extracted from these patterns of ridges and furrows, such as singular points, ridge endings and ridge bifurcations. The distribution of these fingerprint features is distinct not only between individuals but also between different fingers of the same individual. With the development of high resolution scanners, some fingerprint recognition systems also consider highly detailed features, e.g., the distribution of sweat pores, on the surface of fingerprint ridges. One of the challenges in fingerprint recognition is working with latent fingerprints that are accidentally deposited on the surface of objects when they are touched or handled, e.g. fingerprints that are obtained from a crime scene. Latent fingerprints are partial impressions of the finger and often have a high degree of noise content. It has been shown that fingerprint pattern is persistent and recognition accuracy of fingerprints is stable, even when the time difference between acquisitions of fingerprint samples spans over several years [50].

*Palm prints* have a large number of creases that form a purportedly unique pattern. The endpoints of some prominent principal lines, commonly referred to as the heart–line,

head–line, and life–line in palmistry are rotation invariant. Some approaches use these endpoints and midpoints for the registration of geometrical and structural features of principal lines for palm print matching. In addition, similar to fingerprints, friction ridge patterns present on the palm are also used for matching.

*Retina* scan has low acceptability because of its intrinsically intrusive nature. However, it has low circumvention because it is not easy to change or replicate the vascular configuration of the retina, considered to be a characteristic of each individual and each eye, much like the fingerprint.

*Iris* contains a complex pattern that contains many distinctive features such as arching ligaments, furrows, ridges, crypts, rings, corona, freckles and a zigzag collarette [51]. Iris based biometrics is considered to be less intrusive than retina based methods. Non-intrusive methods for extraction of iris features from facial images have also been proposed [52].

*Keystroke* is the way that each person types on a keyboard and is considered to be a characteristic of the individual. This behavioral biometric is not expected to be unique to each individual but it is expected to offer sufficient discriminatory information that permits verification of identity [53].

*Signature* is widely accepted as a method of authentication in government, legal, and commercial transactions. The way a person signs their name is known to be a characteristic of that individual [54]. Signatures are a behavioral biometric that change over a period of time and are influenced by physical and emotional conditions of the individual [55].

*Voice* is a combination of physical and behavioral biometrics. The features of an individual's voice are based on the shape and size of the appendages (e.g., vocal tracts, mouth, nasal cavities, and lips) that are used in the synthesis of the sound [56]. These physical characteristics of human speech are invariant for an individual, but the behavioral part of the speech of a person changes over time due to age, medical conditions (such as common cold), emotional state, etc. [55].

In addition to the above, several other biometric traits have been considered for person recognition. It is, however, noteworthy that these other traits do not measure favorably in comparison to the traits discussed above, on the criteria described at the beginning of this subsection.

## 2.2. Unimodal Biometric Systems

A system that relies on a single acquired sample of a single biometric trait is known as a unimodal biometric system. Even though an ideal single biometric trait that satisfies all the characteristics required for a practical biometric system (subsection 2.1.1) is not known yet, unimodal biometric systems have widely been used for applications where the size of the target population is limited.

The choice of a particular biometric trait is dependent on the application scenario, in particular on factors such as the size and demographics of the target population, the level of security expected from the application and the cost considerations.

## 2.3. Biometric System Modules

The process of biometric recognition is usually divided into two distinct phases – enrolment and recognition.

In the enrollment phase, the biometric trait of an individual is first captured using a sensor. The captured representation of the biometric trait is further processed by a feature extractor to generate a distinctive feature representation, called a template. Depending on the application, the enrolled template is either stored in the central database of the biometric system or recorded on a smart card issued to the individual. Often, multiple templates are enrolled to account for variations observed in the biometric trait. Templates in the database are sometimes also updated over time to account for variations in the trait that occur over time.

In the recognition phase, the biometric trait is captured and a probe template is generated from the captured biometric analogous to the enrolment phase. The generated template is then compared to the templates stored in the database, for the purpose of recognition.

In summary, biometric systems, irrespective of the trait being used, generally consist of the following subsystems [57]:

i. A *sensor* module which captures the biometric trait as raw data.
ii. A *quality estimation* module to determine whether the raw data captured is of a sufficiently high quality, such that further processing on the data would produce meaningful results, or whether another attempt should be made to capture the raw data.
iii. An *enhancement* module that improves the raw data by applying signal processing operations and a *segmentation* module that separates the useful part of the raw data from the background that may have been captured with the data.
iv. A *feature extraction* module which extracts features from the captured biometric data,
v. A *database* module where the extracted features are stored in the form of a template, along with other identifying information corresponding to the various users.

vi. A *matching* module which compares the extracted features to the features stored in the database (templates) to generate matching scores, and a *decision* module which determines or verifies the identity based on the scores generated.

A more detailed discussion on the modules of biometric systems is presented below. The interconnections between the various modules of a biometric system is illustrated in Fig. 2.3.

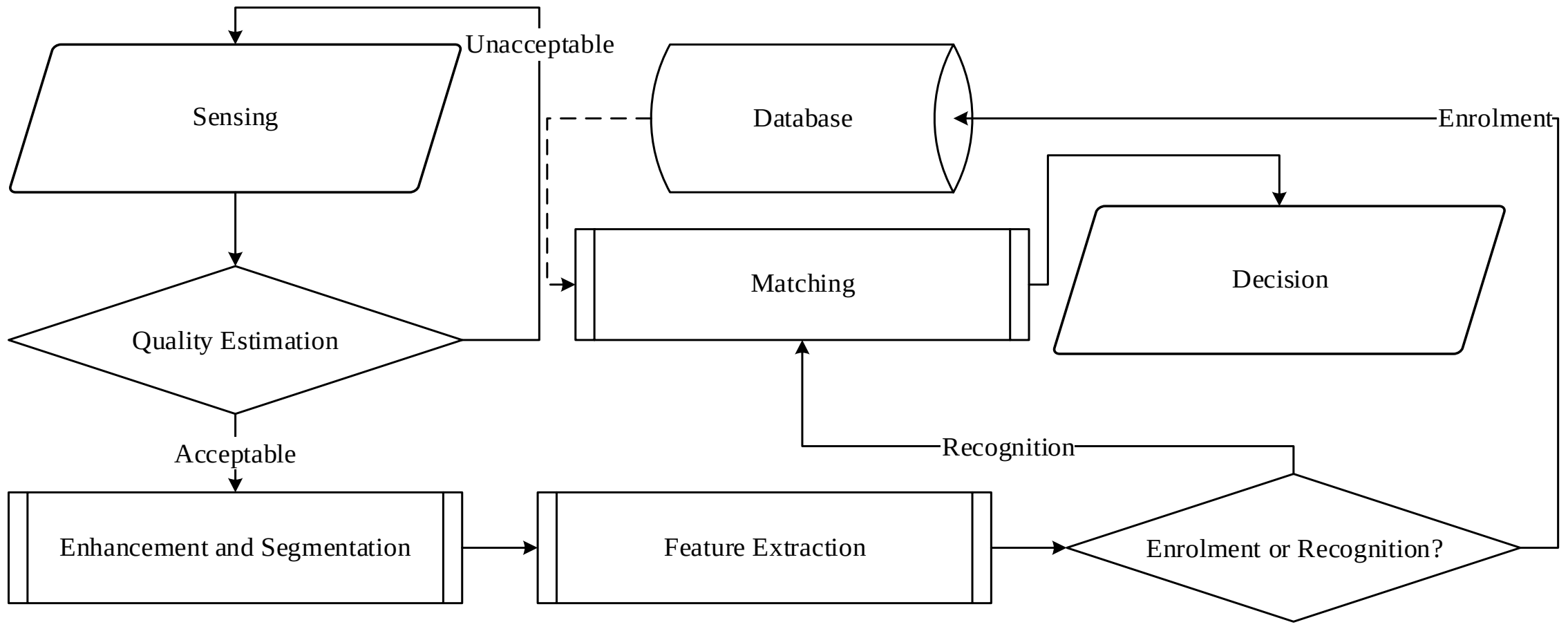

**Fig. 2.3 Modules of a biometric system.**

## 2.3.1. Sensing

The sensing module captures a representation of the biometric trait of an individual, generally by sampling the analog nature of the trait to a digital format that is amenable to further digital computations.

For example, a fingerprint sensor captures an image of the ridge-valley patterns present on the human finger. Other examples of the sensing module include a digital camera for capturing a face image or a digital audio recorder for acquiring a speech sample.

### 2.3.2. Quality Estimation

The quality of the acquired biometric sample determines to a great extent the accuracy of the recognition. The captured biometric samples are therefore first processed through a quality estimation module to determine whether the quality of the acquired sample is reasonable and adequate for further processing, or whether another sample needs to be captured.

Quality estimation also corroborates the confidence in the results of the matching module for the biometric sample being considered.

### 2.3.3. Enhancement and Segmentation

The acquired biometric sample is generally enhanced using one or more signal processing techniques. For example, contrast improvement on a face image, or fingerprint image enhancement using directional image filtering techniques. The captured sample may also have other information, besides the biometric trait of interest. For example, a face image may also contain other objects in the background. Therefore, region of interest is segmented from the acquired sample. This is referred to as segmentation.

### 2.3.4. Feature Extraction

The feature extraction module processes the segmented and enhanced biometric data to extract a set of salient or discriminatory features. An example of a set of features is the position and orientation of minutiae points (local ridge endings and bifurcations) in a fingerprint based biometric system.

For some biometrics traits and associated algorithms, the number of features generated may be very large, leading to what is commonly referred as the "curse of dimensionality". For such algorithm, the feature extraction is often followed by a dimensionality reduction technique to retain only those features that contain discriminatory information.

### 2.3.5. Database

The biometric templates of the enrolled users are stored in the background database for the purpose of matching at the time of recognition. As discussed in section 1.3, the biometric sample acquired at the time of recognition is matched with all templates stores in the database for the identification application and with the template of the claimed identity for the verification application.

The database module sometimes also stores the acquired biometric sample, besides the stored templates. Even though the acquired sample is not used for matching directly, its storage enables updating templates without requiring re-enrolment of users.

### 2.3.6. Matching and Decision

The matching module compares the features of the probe template captured for recognition against the enrolled templates to generate matching scores. For example, the number of matching minutiae between the input and the template fingerprint images is determined and a matching score is computed in a fingerprint based system.

The matcher scores are provided to an encapsulated decision making module, where a user's identity is confirmed or established based on the matching scores. Since biometric matches are not perfect but approximate, the matching scores are usually distributed over a wide range, with the score distributions for genuine and impostor matches overlapping over certain regions.

A sample score distribution for genuine and impostor matches for a typical biometric system is shown in Fig. 2.4.

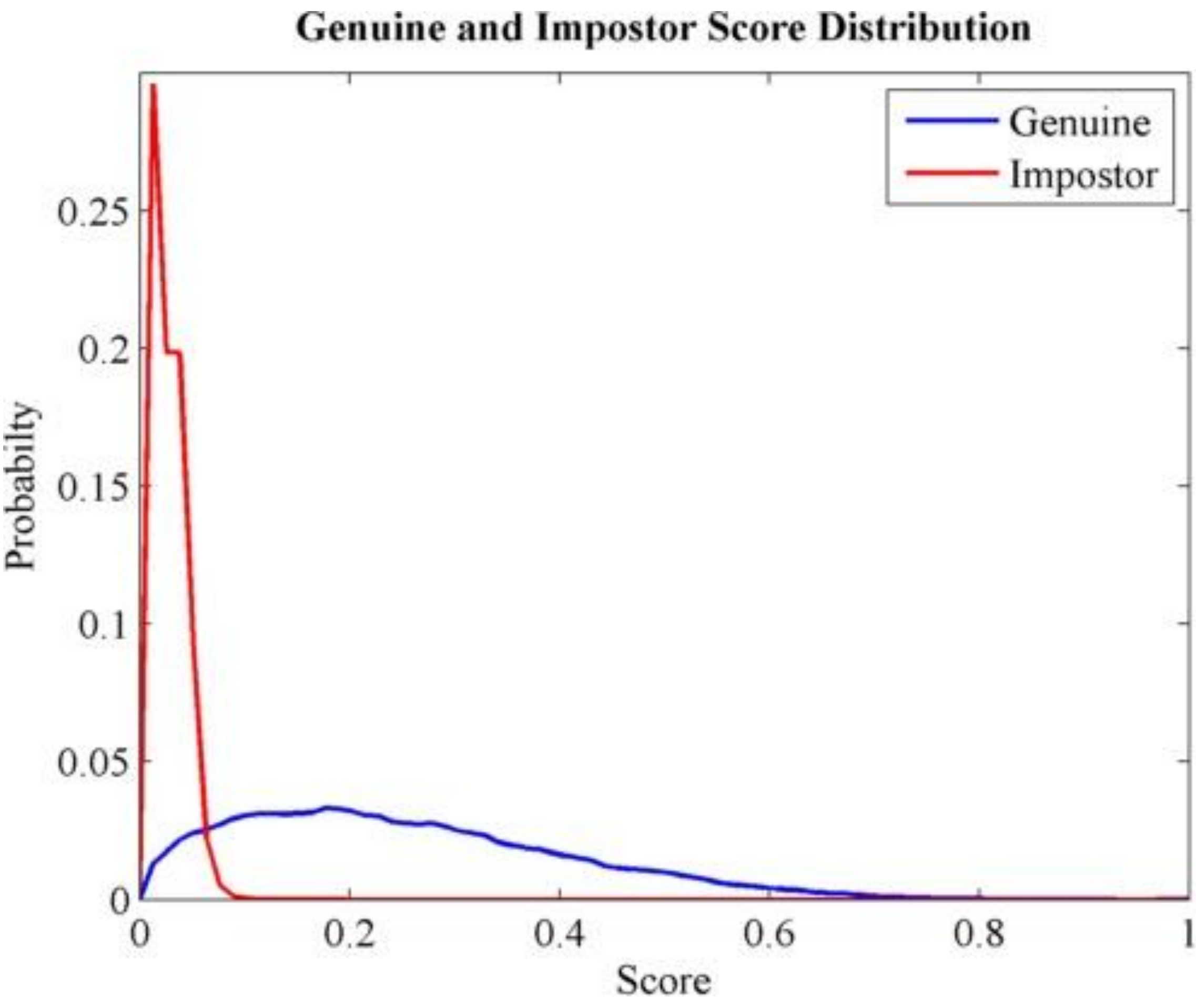

**Fig. 2.4 Distribution of genuine and impostor match scores.**

## 2.4. Multi-biometric Systems

Biometric systems that rely on a single instance of biometric evidence (unimodal systems) may suffer from limitations such as lack of uniqueness (large intra-class variations and large inter-class similarities), non-universality, noisy data, and spoof attacks [17]. These inadequacies get more pronounced with scale, but can be overcome by fusing information from multiple biometric traits [58] (multi-modal systems) for large scale applications [59, 60]. Multi-biometric systems, are expected to be more reliable due to the presence of multiple, independent pieces of evidence. These systems combine the information presented by multiple biometric sensors, algorithms, samples, units, or traits. Besides improving matching performance, these systems improve population coverage, deter spoofing and impart fault tolerance to biometric applications. Multi-modal biometric systems also have improved accuracy over unimodal systems [13].

Multi-biometric systems offer several advantages over unimodal biometric systems, including but not limited to addressing non-universality (that is, biometric trait for a subject is missing or cannot be enrolled), filtering and indexing of large databases, spoofing, noisy data, malfunctioning of sensors and monitoring of individuals.

Since most large scale programs are expected to be inclusive for the entire population, that is, no individual should be denied an identity in the system due to the absence of a biometric trait, or the inability of the system to capture that trait. In other words, the problem of biometric exceptions is handled elegantly with the use of multi-modal biometric systems [61].

All the advantages of multi-biometric systems notwithstanding, these benefits do not come without associated costs. The multiple pieces of evidence demand additional human and computational effort in acquiring and processing information from multiple sources, while also causing an increased inconvenience to the user.

## 2.5. Information Fusion in Multi-biometric Systems

Even though a multi-biometric system uses multiple pieces of evidence, it is expected to provide a single response about the identity of a user. This requires fusing information obtained from different sources to generate a consolidated result. Information may be fused in different ways and at different levels in multi-biometric systems. These are discussed briefly in the following two subsections.

### 2.5.1. Types of Fusion

Fusion in multi-biometric systems is typically classified into six different types, depending on how multiple pieces of evidence are captured or processed [62]:

i. *Multi-sensor*, employing multiple sensors to capture a single biometric trait,
ii. *Multi-algorithm*, invoking multiple feature extraction and/or matching algorithms,
iii. *Multi-instance*, using multiple instances of same body trait, e.g. using both irises for person recognition,
iv. *Multi-sample*, using a single sensor to acquire multiple samples of the same trait,
v. *Multi-modal*, utilizing evidence from multiple biometric traits, and,
vi. *Hybrid*, integrating a subset of the five scenarios above.

### 2.5.2. Levels of Fusion

Biometric traits may be fused at different levels, corresponding to the subsystems described above.

Fusion at the *sensor level* combines raw biometric data. It is not particularly useful with different biometric traits. However, sensor level fusion is used when either multiple sensors are used to capture the same biometric (multi-sensor system) or when multiple samples are collected (multi-impression system). Sensor level fusion has been used to create composite templates by mosaicking [63] and for creating 3D face models from 2D face images [64, 65, 66].

*Feature level* fusion is used to combine feature sets obtained from:

i. different feature extraction algorithms, or
ii. samples of the same or different [67, 68] biometric traits into a single feature set.

Fusion at this level suffers from the limitation that most Commercial Off-The-Shelf (COTS) systems do not provide access to feature sets. Furthermore, feature sets may be incompatible (e.g., variable-length vs. fixed-length) and may require dimensionality reduction techniques [69].

The matching module of biometric systems generally outputs a similarity or a distance score between the probe template biometric evidence presented (or the feature set extracted therefrom) and each of the templates stored in the database. Fusion at the *score level* is also known as fusion at the measurement level or confidence level. If the distance or similarity scores are available, score level fusion is the most preferred since it contains rich information about the match while abstracting the details of feature representation and matching algorithm. The ease with which scores from different biometric matchers can be combined makes the matching score an optimal choice for information fusion [70].

Several commercial systems, when deployed in the identification mode rank the possible candidates based on their matching scores but without providing a direct access to these scores. Fusion at the *rank level* is particularly useful in such scenarios. The final rank is derived by fusing the ranked candidate lists provided by individual systems [62] with the use of techniques such as highest rank, Borda Count or logistic regression [71]. While the first two methods can be employed on datasets of any size, learning based methods, e.g., logistic regression [72], require a reasonably sized training data set.

Fusion at the *decision level* provides the least amount of flexibility but it may be the only available fusion alternative when commercial systems do not provide access to the extracted feature set and matching scores Decision level fusion techniques include majority voting, OR and AND logical operators, or clustering methods, such as the k-means clustering [73], fuzzy k-means [74], fuzzy vector quantization [75], and fuzzy clustering of fuzzy data [76]. Note that an optimal combination of matchers at the decision level can significantly enhance matching accuracy if the classifiers are independent [77].

### 2.5.3. Transformation and Combination Based Fusion

Biometric fusion at the score level presents the best trade-off between the amount of information available and the ease with which the information can be fused [78], making it by far the most popular choice. Main methods for score-level fusion are discussed in this and the following subsections.

#### 2.5.3.1. Normalization

The scores obtained from different matchers may have different ranges, (e.g., [0,100], [-1, 1], etc.), and may be of different types (similarity vs. distance). Therefore, it is important to normalize the scores from different matchers to a common domain, say [0,1], before combining them using a fusion rule. The normalization techniques may be linear or non-linear based on the function used to normalize the scores. Examples of linear normalization techniques include min-max normalization that is used to normalize the scores to a [0,1] range, and decimal scaling to normalize scores to the same order of magnitude. Non-linear score normalization techniques use normalization functions with tuneable parameters, e.g., double sigmoid and tanh functions. These techniques are discussed in detail below.

*Min-max* normalization is used to normalize the scores to a [0,1] range. The minimum and maximum bounds on the possible scores may be known a priori or may be estimated from the data. Given the original score $s_k$, the normalized score $s'_k$ is given by

$$s'_k = \frac{s_k - \min(s)}{\max(s) - \min(s)} \tag{1}$$

Min-max normalization is not robust and is sensitive to outliers, especially when the bounds on the scores are estimated from data.

*Decimal scaling* is useful when the scores from different matchers are different by orders of magnitude. The normalized score $s'_k$ is calculated as

$$s'_k = \frac{s_k}{10^n} \tag{2}$$

where $n = \log_{10}[\max(s)]$. Decimal scaling is not robust and is sensitive to outliers.

*z-score* is the most commonly used normalization technique and is useful when the scores have a Gaussian distribution. The normalized score $s'_k$ may be computed as

$$s'_k = \frac{s_k - \mu}{\sigma} \tag{3}$$

where $\mu$ and $\sigma$ are the mean and standard deviation of the distribution of $s$, either known or estimated from the data. The z-score normalization is sensitive to outliers, but comparatively lesser than the earlier two methods. Note that z-score normalization may not be optimal if the actual underlying score distribution is not Gaussian [70].

*Median and median absolute deviation (MAD)* has been proposed as an alternative to z-score normalization. In this method, the normalized score $s'_k$ is computed as

$$s_k' = \frac{s_k - \text{median}(s)}{MAD} \tag{4}$$

where $MAD = \text{median}(|s_k - \text{median}(s)|)$. This method of normalization is not sensitive to outliers but performs worse than z-score normalization when the underlying distribution is not Gaussian.

Both the z-score and the median and MAD normalization schemes do not return normalized scores in a pre-determined bounded interval.

Non-linear normalization techniques that use tunable parameters have also been proposed. These include double sigmoid [79] and tanh [80] normalization.

The *double sigmoid* function has three tunable parameters $t, r_1$ and $r_2$, where $r_1$ and $r_2$ denote the region around operating point $t$ where normalization is approximately linear. The normalized score $s_k'$ is computed as follows

$$s_k' = \begin{cases} \dfrac{1}{1 + \exp\left(-2(s_k - t/r_1)\right)}, & s_k < t \\ \dfrac{1}{1 + \exp\left(-2(s_k - t/r_2)\right)}, & otherwise \end{cases} \tag{5}$$

This normalization is especially useful in amplifying the difference in the region of overlap between genuine and impostor scores $(t - r_1, t + r_2)$. An example of double sigmoid normalization that maps scores to a (0, 1) domain is shown in Fig. 2.5. The values of $t, r_1$ and $r_2$ in this example are 600, 160 and 120 respectively. Note that the region corresponding to the scores in the range (440, 720) is approximately linear and is mapped over a large domain.

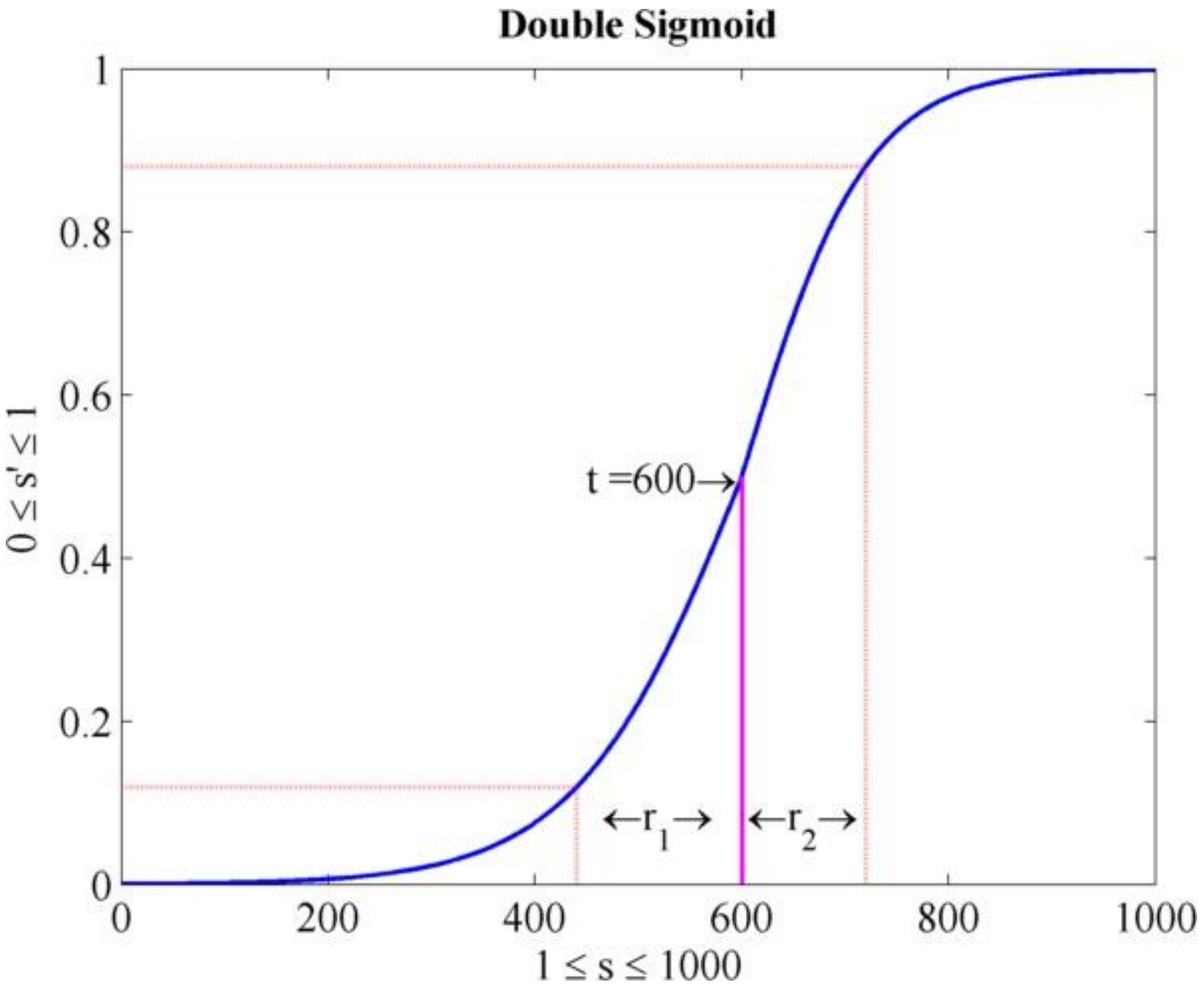

**Fig. 2.5 An example of double sigmoid normalization input - output curve. The value of the operating point t is 600, while $r_1$ and $r_2$ that denote the region around the operating point where normalization is approximately linear are 160 and 120 respectively.**

*Tanh* normalization [80] is robust and is insensitive to outliers. The normalized scores may be computed as

$$s'_k = \frac{1}{2}\left(\tanh\left(\alpha\left(\frac{s_k - \mu_H}{\sigma_H}\right)\right) + 1\right) \qquad (6)$$

where α is a tunable parameter that determines the spread of the normalized scores and $\mu_H$ and $\sigma_H$ are the robust Hampel mean and standard deviation estimates respectively of the distribution of $s$, based on the Hampel influence function that itself has three tunable parameters $a$, $b$, $c$ and is given by

$$\psi(u) = \begin{cases} u, & 0 \leq |u| < a \\ a\text{sign}(u), & a \leq |u| < b \\ a\text{sign}(u)\left(\dfrac{c - |u|}{c - b}\right), & b \leq |u| < c \\ 0, & |u| \geq c \end{cases} \tag{7}$$

#### 2.5.3.2. Combination Based Fusion

Once the scores from different matchers have been transformed to a common domain, they are fused together using a fusion rule, such as one of the following [81]. These fusion rules work best when the scores obtained from different matchers are independent.

The *sum rule* [17] finds the weighted arithmetic mean of the normalized matching scores obtained from different classifiers.

$$score^{(j)} = \sum_i w_i^{(j)} s_i^{(j)} \tag{8}$$

where $score^{(j)}$ is the final score after fusion for the $j^{th}$ subject and $w_i^{(j)}$ and $s_i^{(j)}$ are the weights and normalized scores respectively for the $i^{th}$ matcher and the $j^{th}$ subject. The weights corresponding to different matchers may be subject specific, Note that the weighted average is subject to the condition $\sum_i w_i^{(j)} \stackrel{\text{def}}{=} 1\ \forall j$.

The *product rule* [82] finds the weighted geometric mean of the normalized matching scores.

$$score^{(j)} = \left(\prod_i w_i^{(j)} s_i^{(j)}\right)^{\frac{1}{\sum_i w_i^{(j)}}} \tag{9}$$

The *median rule* is a robust alternative for the sum rule, because the arithmetic mean is sensitive to outliers.

$$score^{(j)} = \operatorname*{median}_{i}\left(s_i^{(j)}\right) \tag{10}$$

The *max rule* chooses the normalized score from the matcher that presents the greatest degree of confidence.

$$score^{(j)} = \max_{i}\left(s_i^{(j)}\right) \tag{11}$$

The *min rule* chooses the normalized score from the matcher that presents the least degree of confidence.

$$score^{(j)} = \min_{i}\left(s_i^{(j)}\right) \tag{12}$$

### 2.5.4. Density Based Fusion

Density based fusion techniques require knowledge of the densities for genuine and impostor scores, either a priori, or through estimation [77]. The scheme, based on the Neyman-Pearson lemma [83], uses a likelihood-ratio test which rejects the null hypothesis $H_0$ in favor of the alternate hypothesis $H_1$ when

$$\Lambda(x) = \frac{L(\theta_0|\boldsymbol{x})}{L(\theta_1|\boldsymbol{x})} \leq \eta \tag{13}$$

where $P(\Lambda(\boldsymbol{X}) \leq \eta|H_0) = \alpha$ is the most powerful test of size (false accept rate) α for the threshold η.

This technique, therefore, directly provides a way to create a system that approximates a pre-specified error rate. However, since the densities for the genuine and impostor scores are not known a priori, these are usually estimated from the available training data.

Some biometric matchers may output specific score values under certain conditions, and therefore, modeling with continuous density functions may not present an accurate representation. Consequently, the use of generalized density functions, consisting of both discrete and continuous parts has been proposed. One of the methods involves using kernel density estimators (KDE), however, requires a careful selection of the kernel type and bandwidth. The distributions for individual traits are combined to obtain a joint density using copula functions [84].

The distribution of the genuine and impostor scores can also be represented using finite Gaussian mixture model (GMM) that incorporates biometric sample quality [85]. The estimate for the density may be obtained as

$$\hat{L}(\theta_i|\boldsymbol{x}) = \sum_{j=1}^{M_i} p_{i,j} \emptyset^K\left(\boldsymbol{x}; \boldsymbol{\mu}_{i,j}, \boldsymbol{\Sigma}_{i,j}\right) \tag{14}$$

where $i \in \{0,1\}$ corresponding to genuine and impostor classes, $\emptyset^K(\boldsymbol{x}; \boldsymbol{\mu}, \boldsymbol{\Sigma})$ is the K-variate Gaussian density with mean vector $\boldsymbol{\mu}$ and covariance matrix $\boldsymbol{\Sigma}$, $M_i$ is the number of mixture components used to model the score density and $p_{i,j}$ is the weight assigned to the jth mixture component in $\hat{L}(\theta_i|\boldsymbol{x})$, and $\sum_{j=1}^{M_i} p_{i,j} \stackrel{\text{def}}{=} 1 \; \forall i$.

The selection of the number of mixture components is critical in this model as it may lead to an over-fitting or under-fitting density function. An elegant algorithm that automatically determines the number of mixture components to be used has been proposed in [86].

Normalization of scores and selection of optimum weights for score fusion for individual subjects not required for density based fusion, Furthermore, density based fusion is also able to handle discrete valued score distributions. However, the fusion results are sub-optimal when the estimates for the score distribution deviate from the true distribution. In such a case, a reasonably large number of representative training samples are required

## 2.5.5. Classifier Based Fusion

Classifier based fusion considers the scores produced by the different matchers as a feature vector for the two-class (genuine vs. impostor) classification problem. Several classification approaches based on different soft computing techniques have been proposed for example, based on support vector machines [87], fuzzy logic [76, 88], and neural networks [89].

However, it is to be noted that when machine learning methods are used, the performance depends not only on the choice of classifier but also on how representative the training examples are.

## 2.5.6. Dynamic Score Selection

The dynamic score selection [90] strategy chooses one of the unimodal scores over others as the final score and is especially useful when only low quality scores are available. An ideal dynamic score selection strategy would be one that would use the max rule for genuine subjects and the min rule for imposters. However, since it is not known a priori whether or not the subject is genuine, the following two-stage approach has been proposed:

i. Estimate whether or not the subject is genuine using a classifier, as discussed in subsection 2.5.5.

ii. Select the normalized score based on the output of the classifier.

$$score^{(j)} = \begin{cases} \max_i\left(s_i^{(j)}\right), & s^{(j)} \mathrel{\widehat{\in}} \text{genuine} \\ \min_i\left(s_i^{(j)}\right), & \text{otherwise} \end{cases} \tag{15}$$

The effectiveness of this dynamic score selection strategy is largely dependent on efficiency and accuracy of the classifier.

### 2.5.7. Quality Based Fusion

The accuracy of biometric matching is dependent not only on the matching algorithm, but also on the quality of the biometric sample captured by the sensor [91]. The quality of the sample is contingent on the subject (e.g. skin condition, subject cooperation), the sensing equipment and the ambient conditions (e.g. lighting) [92]. Several measures for assessing quality have been proposed for fingerprint [93, 94], face [95, 96] and iris [97, 98] biometric traits. This has fostered the development of dynamic score selection and fusion algorithms that adapt themselves to the quality of individual biometric samples by adjusting weights or choosing classifiers based on the quality of the captured sample [99, 100, 101].

## 2.6. Evaluation of Biometric Systems

Both the accuracy and the speed of a biometric system is of great importance in a practical application scenario. The success or failure of any unimodal or multi-biometric system is required to be measured quantitatively for the purpose of comparison between existing systems or to evaluate the effectiveness of a new algorithm. Commonly used metrics for performance evaluation of biometric systems are discussed in subsection 2.6.1 and some curves used for comparative evaluation of biometric systems have been described in subsection 2.6.2. The current state of the art for biometric systems, based on the evaluations using these metrics and curves is presented in subsection 2.6.3. Since the evaluation parameters are not completely independent of the biometric data used for evaluation, benchmark databases are generally used to compare performance between systems. A description of benchmark systems with some examples is discussed in subsection 2.6.4.

### 2.6.1. Evaluation Metrics

Some of the metrics that are commonly used for quantitative determination of effectiveness of a biometric system are described below. The acceptance and rejection rates are used in the context of verification scenario while the positive and negative identification rates are used for the identification scenario.

i. *True Acceptance Rate* or *True Positive Identification Rate* (TAR or TPIR) measures the probability that the system correctly indicates a successful match between the input pattern and a mated template in the database.
ii. *True Rejection Rate* or *True Negative Identification Rate* (TRR or TNIR) is the probability that the system does not indicate a match between the input pattern and a non-mated template in the database.
iii. *False Acceptance Rate* or *False Positive Identification Rate* (FAR or FPIR) is the probability that the system incorrectly indicates a successful match between the input pattern and a non-mated template in the database.

iv. *False Rejection Rate* or *False Negative Identification Rate* (FRR or FNIR) is a measure of the probability that the system incorrectly indicates a non-match between the input pattern and a mated template in the database.

v. *Equal Error Rate* (EER) is the value of the false acceptance and false rejection rates when the two are equal. This is a result of the trade-off between the false acceptance and false rejection rates, due to the overlap in the genuine and impostor score distributions. Improvement in one of these error rates by attempting to vary the threshold on matching scores worsens the performance in terms of the other error rate.

vi. *Rank-1 Accuracy* is used in the identification scenario and is the probability that the system will retrieve the correct mate from the background database at rank-1 with the highest probability.

vii. *Failure To Enroll Rate* (FTE or FTR) is the percentage of data captured that is considered invalid or that fails to enroll into the system.

viii. *Failure To Capture Rate* (FTC) is generally applicable to automated systems and is the probability that the system fails to detect a biometric characteristic (such as a face in a picture) when presented correctly.

### 2.6.2. Evaluation Curves

Some of the evaluation metrics discussed in the previous subsection are mutually related and this relationship is often plotted graphically to enable comparison of performance of biometric systems at different operating points. The commonly used plots for this purpose are the Receiver Operating Characteristic (ROC) or Detection Error Trade-off (DET) [102] and the Cumulative Match Characteristic (CMC) curves for the verification and the identification scenarios respectively.

i. *Receiver Operating Characteristic* or *Detection Error Trade-off* (ROC or DET) curves are used in the verification scenario. These are plots between the values of TAR or FRR respectively against the FAR. An example DET curve is shown in Fig. 2.6. The EER may be obtained from the DET curve as the point where the values of FAR and FRR are equal. The ROC and DET are generally plotted against semi-logarithmic (log-linear) and logarithmic (log-log) axes respectively.

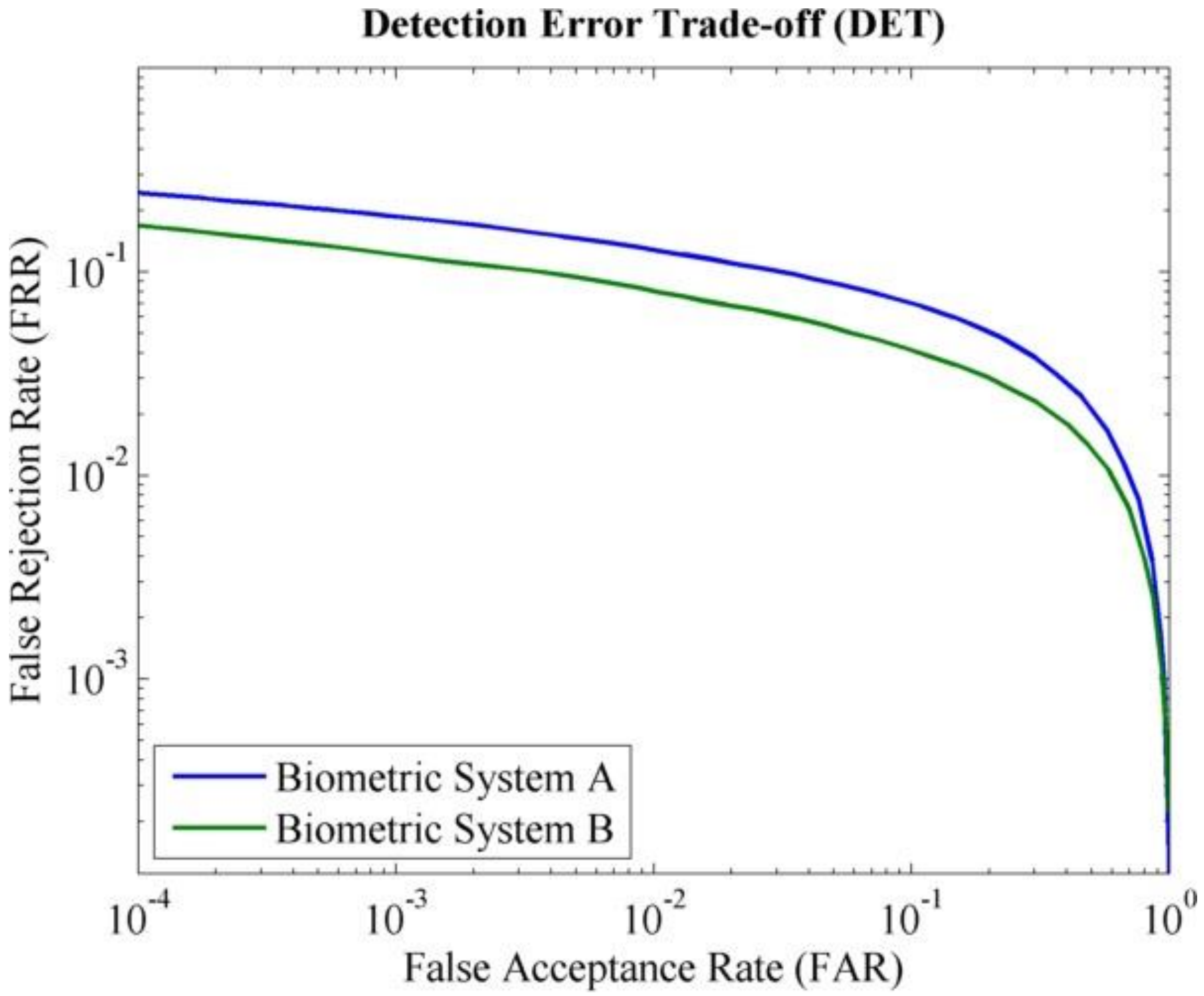

**Fig. 2.6 An example of Detection Error Trade-off (DET) curve.**
**It may be inferred from the plot shown that Biometric System B has a better performance than Biometric System A.**

ii. *Cumulative Match Characteristic* (CMC) curve is used in the identification scenario and is a plot of the identification rate, that is, the percentage of the subjects correctly identified at or before a particular rank. An example CMC curve for top 100 ranks is shown in Fig. 2.7. The CMC curve is generally plotted against linear axes.

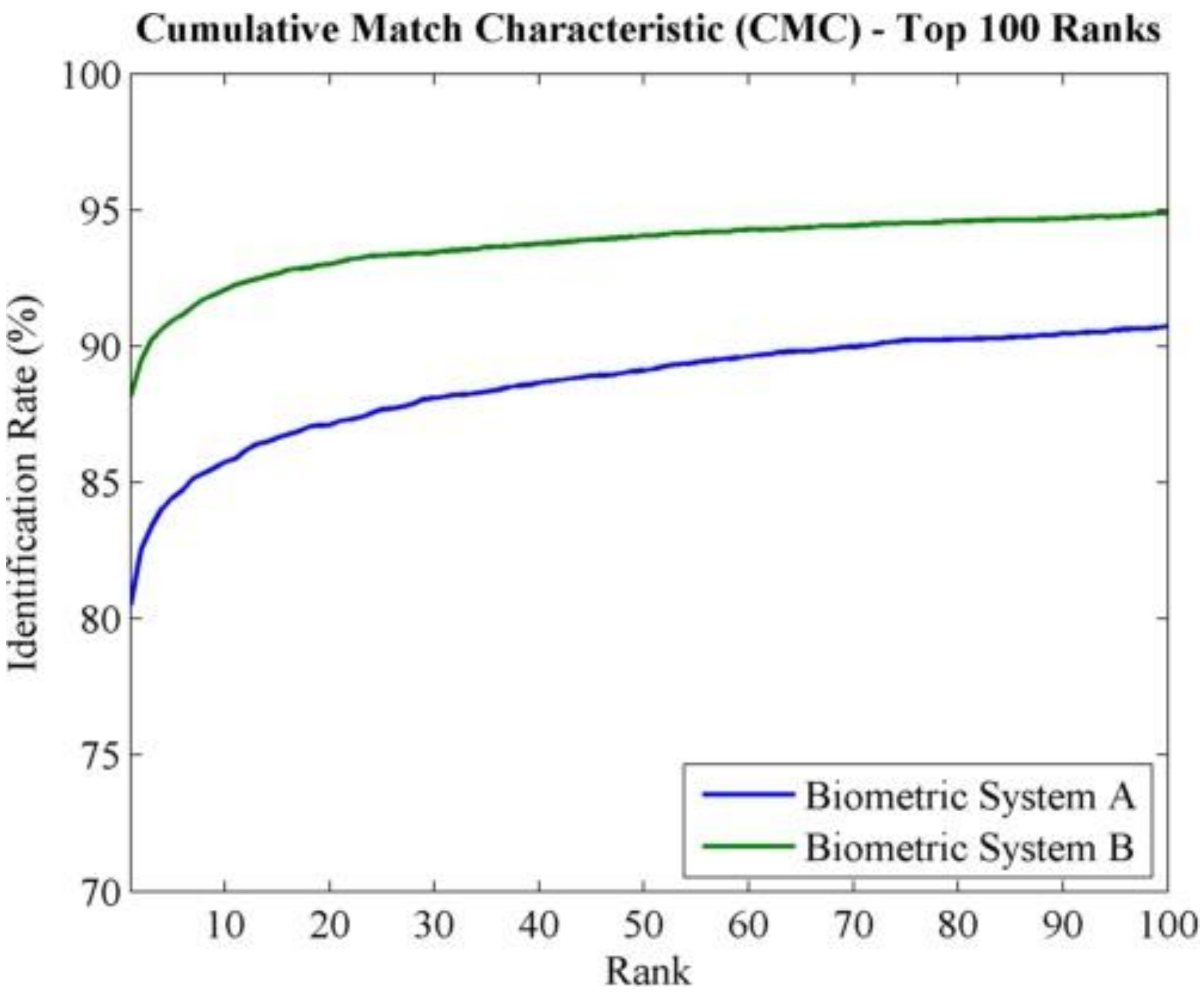

**Fig. 2.7 An example of Cumulative Match Characteristic (CMC) curve. It may be inferred from the plot shown that Biometric System B has a better performance than Biometric System A.**

The performance metrics and curves discussed above are generally not constant for a system and are dependent not only on the biometric trait or the algorithm being used, but on several factors. For example, a face matching algorithm that works very well for a certain group of subjects may have a poor performance for another set of subjects. In order to obtain practical values of performance metrics in an actual deployment scenario and to compare across systems, benchmark databases that present a reasonable representation of the population of interest are therefore used. Some public evaluations or competitions for ascertaining and improving the state of the art for various unimodal and multi-biometric identifiers using benchmark datasets are also organized periodically.

### 2.6.3. State of the Art

Some representative efforts that present an overview of the current state of the art, as determined through the evaluations conducted by the National Institute of Standards and Technology (NIST) of various biometric traits are presented in Table 2.1.

Even though several biometric traits and multi-biometric systems have been discussed in the preceding sections, the evaluations on the state of the art for biometric modalities conducted by NIST are for the traits that are considered to be well established and are most commonly used in large scale recognition systems.

**Table 2.1 State of the Art for Biometric Systems**

| Biometric Trait | State of the Art | Reference and Comments |
|---|---|---|
| Face | Rank-1 identification accuracy of 95.9%. | FRVT 2013 [103]<br>Mugshot images from 1.6 million subjects. |
| Fingerprint | False negative identification rates (FNIR) of 1.9% at false positive identification rate (FPIR) of $10^{-3}$. | FpVTE 2012 [104]<br>30 000 search subjects (10,000 mates and 20,000 non-mates). |
| Latent fingerprint | Rank-1 identification accuracy of 63.4%. | ELFT-EFS Evaluation #2 [105]<br>1,066 latent fingerprint images from 826 subjects. Gallery of mated exemplar sets from all 826 latent subjects, as well as 99,163 non-mated exemplar sets from other subjects. |
| Iris | False negative identification rates (FNIR) of 2.0% at false positive identification rate (FPIR) of $10^{-4}$. | IREX IV [106]<br>Enrolled population size of 1.6 million. |
| Voice (Speaker) | False miss (false reject) probability of 5% at false alarm (false accept) probability of 0.1%. | 2012 NIST Speaker Recognition Evaluation [107]<br>1918 target speakers under multiple training and test evaluation conditions |

The Unique Identification Authority of India also conducted a proof of concept [108] study on a population size of 20,000 subjects. The experiment included fusion of information from fingerprint and iris under the following multi-biometric scenarios:

i. Iris, from both eyes;

ii. Fingerprint, all ten prints; and,

iii. Iris from both eyes and fingerprint tenprint combined.

The error rates under these scenarios, plotted as the Receiver Operating Characteristic (ROC) on a log-linear scale is illustrated in Fig. 2.8.

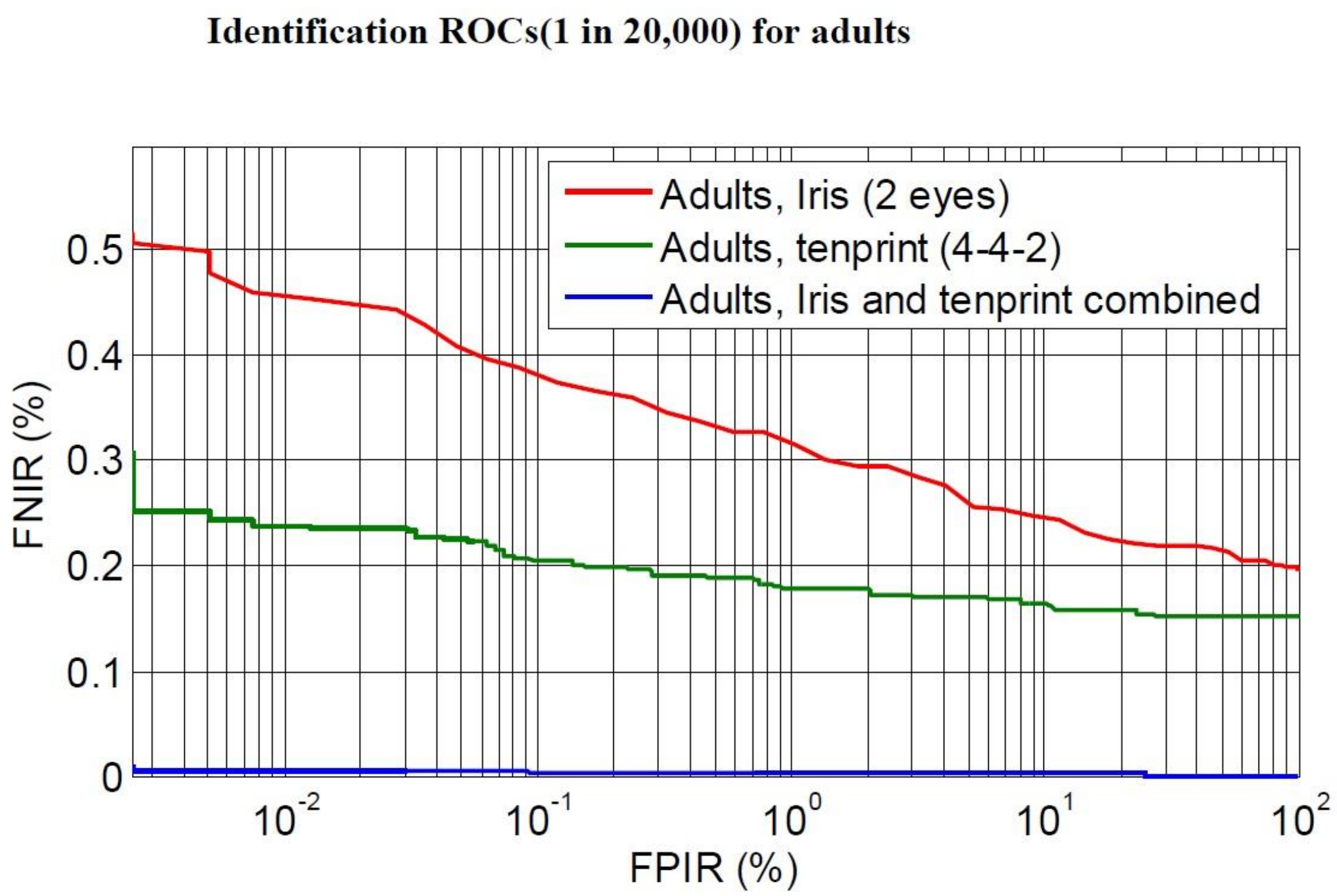

**Fig. 2.8 Receiver Operating Characteristic (ROC) curve comparatively illustrating the performance of multi-biometric systems [108].**

## 2.6.4. Benchmarks

Several unimodal and multi-modal biometric databases are available in the public domain for the purpose of benchmarking biometric systems. Some examples of unimodal biometric databases are NIST Special Database 4 [28], NIST Special Database 14 [35] for fingerprints and the NIST Special Database 18 [109] for face recognition. The benchmark databases often also specify training and testing protocols, further standardizing the conditions under which different systems are evaluated.

For benchmarking multi-biometric systems, some multi-modal biometric databases include the XM2VTS [110] database containing face and voice modalities and the BIOMET [111] database consisting of speech, hand geometry, fingerprint, signature and visual and infrared face image data. Comparison of fusion algorithms at the score level may not require the biometric sample and standard databases containing matching scores for different modalities may suffice for this purpose. An example of a multi-biometric score database is the NIST BSSR-1 [33] dataset that contains matching scores for fingerprint and face modalities.

Even though there is no evidence either for or against whether biometric traits are correlated, virtual multi-modal databases are often used to simulate large scale multi-biometric databases. A virtual database consists of virtual identities where the biometric traits have been combined together from different unimodal databases.

## 2.7. Summary

Biometric systems provide a reliable means of person recognition that provides a more robust solution to the person recognition problem when compared to other conventional recognition techniques. However, biometric match is non-exact and most unimodal systems do not fulfil all the criteria laid down for a good biometric trait. Therefore, use of multi-biometric systems has been proposed.

Mere usage of multi-modal biometrics, however, does not necessarily lead to improvement in system performance. Multi-modal systems incur computational overheads of matching and fusing evidence from multiple sources. A poorly designed multi-modal biometric system can deteriorate the performance of individual traits, increase the cost of the system, and present increased inconvenience to subjects and administrators, such as, complex enrolment procedures [55].

Multi-biometric systems also require information obtained from the various sources of evidence to be fused. Even though this information fusion may be of several types and various levels, fusion at the score level presents the best trade-off between complexity and the amount of information available. Once the scores have been transformed to a common domain using normalization, the scores from different matchers are fused together using a fusion rule, such as the sum rule [17], product rule [82], median, max or min rules [81]. The fusion could also be density based [77], classifier based [87] or quality based [100]. The fusion may also be based on a dynamic score selection [90] strategy that chooses one of the unimodal scores over others as the final score.

A description of the performance metrics and benchmarks used for evaluating the performance of biometric systems has also been presented in this chapter.

# CHAPTER 3

# SOFT COMPUTING APPROACHES

## 3.1. Introduction

*Soft computing* has emerged as a constructive and effective computing paradigm for solving several real world challenges. In contrast to *hard computing*, the soft computing paradigm lends itself well to problems that either do not present the necessary details for all aspects required for articulation as precise computational models based on logic and mathematical formulation, or, are computationally intractable. Besides, the soft computing paradigm is also known to be robust to imprecision, uncertainty, partial truth, and approximation, while providing computationally efficient solutions.

Several soft computing paradigms are inspired by the natural processes of computing and optimization. Some examples of soft computing paradigms are logistic regression, inspired by some early models of a neuron; artificial neural networks, that are inspired by the models of neurons in the human brain and their interconnections, also known as the nervous system; fuzzy logic, that draws its inspiration from the imprecision in human thought and natural languages; and evolutionary computation, which is based on the principles of biological evolution. Soft computing is also often viewed as an important ingredient of artificial intelligence.

A description of some of the most popular supervised soft computing paradigms has been presented in the section 3.2 and bootstrap aggregation as a meta-algorithm for soft computing paradigms is presented in section 3.3. The various soft computing paradigms may also be combined together to build more intricate or complex systems. Section 3.4 has a discussion on combination of soft computing paradigms and a summary of the chapter has been presented in section 3.5.

### 3.1.1. Motivation

The ability of the soft computing paradigms to handle imprecision and approximation efficiently naturally makes them amenable to applications in automatic person recognition systems. Specifically, for biometric recognition systems, the feature extraction and matching modules are generally both complex and non-linear, and can therefore be modelled better by soft computing approaches. Biometric recognition is approximate and is considered to be a human capability and therefore, adapts well to soft computing paradigms that are not only suited for approximate computation but also are generally inspired by and attempt to imitate biological function.

Soft computing paradigms are also robust in situations that are commonly encountered in biometric recognition where traditional hard computing paradigms would fail. These situations include noisy, occluded, misaligned or deformed biometric data, or the absence of certain biometric traits. While permanence is a desirable characteristic for an ideal biometric trait, it is accepted that biometric characteristics change with time and matching algorithms based on soft computing paradigms benefit from the capability of these paradigms to model and be adaptive to these changes.

## 3.2. Soft Computing Paradigms

Soft computing algorithms are natural candidates for solving several categories of problems or application domains, e.g.,

i. *Anomaly detection*, to identify observations that do not conform to an expected pattern;
ii. *Association rules*, to identify rules of interest discovered from datasets;
iii. *Classification*, to identify the category to which an observation belongs;
iv. *Clustering*, to group observations together based on some similarity;
v. *Feature learning*, to create useful representations from raw data;
vi. *Grammar induction*, to model a set of rules drawn from characteristics of observations;
vii. *Learning to rank*, to construct ranking models from partial information of order;
viii. *Online learning*, to continuously fine-tune model parameters based on a stream of data;
ix. *Regression*, to estimate the relationship between variables;
x. *Reinforcement learning*, for automating the actions in an environment through reward;
xi. *Semi-supervised learning*, when the target attribute may not be known for all examples used for training.
xii. *Structured prediction*, to predict structured objects instead of scalar values; and,
xiii. *Unsupervised learning*, when the target attribute is not known for the training examples.

Even though the above examples of application domains of soft computing cover a very broad area, the discussion in this section is rather limited to soft computing paradigms based on supervised learning in general, and the classification problem in particular, in the interest of brevity and relevance to this thesis.

Unlike hard computing, soft computing paradigms are not based on the standard algorithmic procedures in a precise mathematical framework, and therefore, the underlying models in these paradigms generally consist of tunable parameters. In supervised learning algorithms, the values of these parameters are determined based on some notion of *best fit* between the predicted outcome and the corresponding ground truth that is known *a priori* for the given examples. The process of tuning the parameters with a view to achieve this best fit is known as training.

Prior to the implementation of a prediction model based on the supervised soft computing paradigm for a certain application, the model is trained and its performance is evaluated. This process generally consists of three distinct stages [112]:

i. The *training* stage, to determine the values of the unknown parameters used in the soft computing paradigm,
ii. The *cross-validation* stage, for validating model and the tuneable parameters and hyper-parameters (e.g., the number of tuneable parameters) used in the model, and,
iii. The *testing* stage, to determine the accuracy of the trained and validated model, usually on examples that the model has previously not been exposed to.

For an accurate soft computing algorithm, the predicted value of the target attribute is expected to be a good approximation of the actual value, that is, the ground truth for the target attribute. The departure of the predicted value from the ground truth is generally represented mathematically as some cost function. The training phase of a supervised soft computing paradigm, therefore, generally involves a minimization of this cost function.

It is important for the designer to ensure that the model is not only a good fit for the given examples but also generalizes well during implementation. An estimate of this generalization is obtained during testing, by presenting examples to the model that have not been shown to the algorithm during the training.

A model that is not sufficiently complex to be able to represent a good fit between the predictions and the ground truth is said to *underfit*, while a model that fits well only on the training examples but does not generalize well to the test examples, is said to be *overfit*. A model may overfit either on account of being too complex or because of insufficient training examples.

## 3.2.1. Logistic Regression

Logistic regression is generally used for classification applications. The paradigm is based on the logistic (also known as sigmoid) function, given by:

$$f(z) = \frac{1}{1 + e^{-z}} \tag{16}$$

where $z$ is generally expressed as

$$z = \sum_i \theta_i x_i \tag{17}$$

Here $x_i$ is the $i^{th}$ independent variable (feature) and $\theta_i$ is the corresponding weight (parameter) associated with that feature. Equations (16) and (17) may be combined to construct the hypothesis for the value of the target variable as follows:

$$h_\theta(x) = \frac{1}{1 + e^{-\sum_i \theta_i x_i}} \tag{18}$$

A graphical plot of the logistic (sigmoid) function is shown in Fig. 3.1.

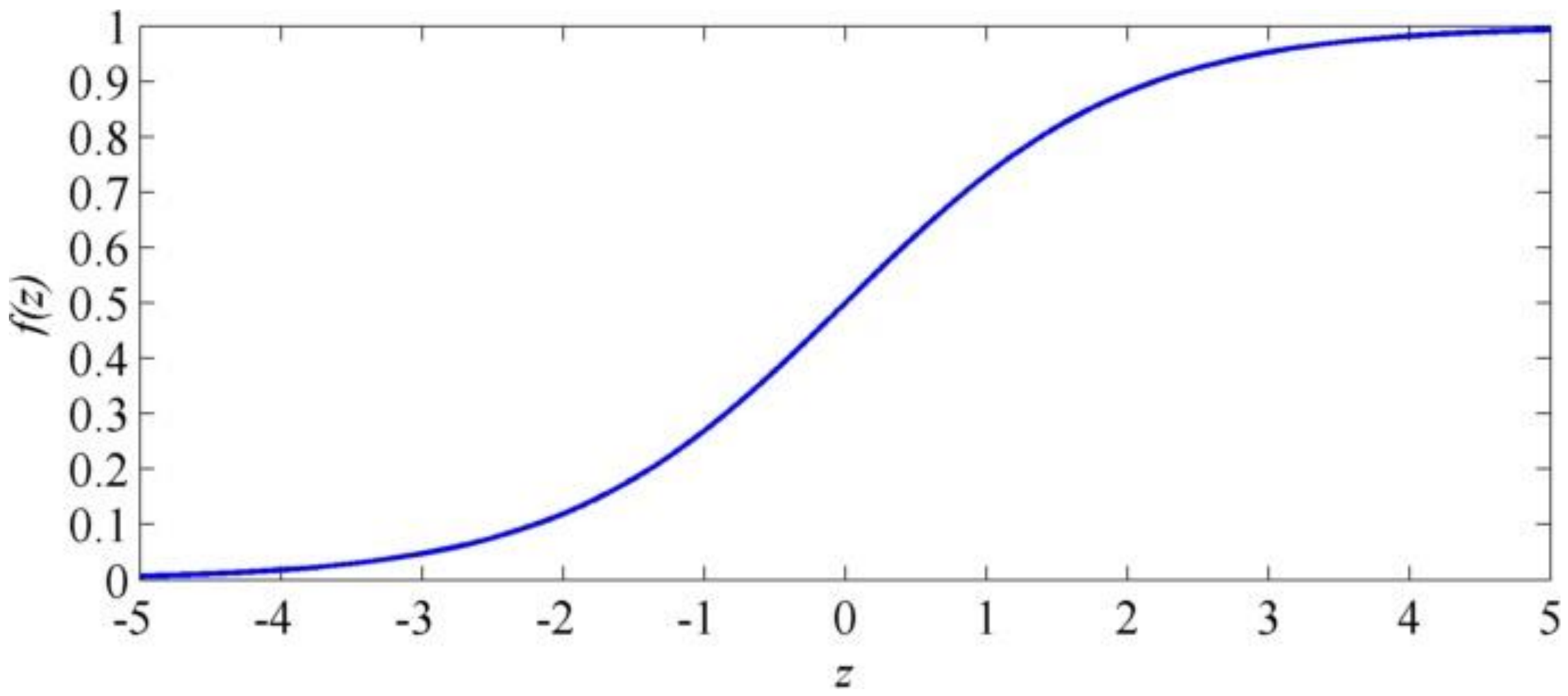

**Fig. 3.1 The logistic function.**

The cost function for the two-class classification problem, where the target variable y is expected to be either 0 or 1, when a certain parameter set $\theta$ is used is given by:

$$J(\theta) = -\frac{1}{m}\sum_{i=1}^{m} y^{(i)} log h_{\theta}\left(x^{(i)}\right) + \left(1 - y^{(i)}\right) log\left[1 - h_{\theta}\left(x^{(i)}\right)\right] \tag{19}$$

Here, $J(\theta)$ is the value of the cost function corresponding to a parameter set $\theta$, $m$ is the number of training samples used for evaluating the cost function, $x^{(i)}$ is the set of input attributes (also called target variables) for the $i^{th}$ sample and $y^{(i)}$ is the ground truth for the target attribute for the $i^{th}$ example.

It may be noted that the logistic function returns a real value in the (0,1) range, which may be converted to a class representation (true or false) based upon whether or not the output value of the logistic function is greater than a certain threshold. The output value of the target variable for the logistic function may also be interpreted as the probability of the example belonging to a certain class.

### 3.2.2. Support Vector Machines

Support vector machines [113] are a popular supervised learning soft computing paradigm for analysis of data and pattern recognition. This paradigm is useful for regression as well as non-probabilistic classification applications.

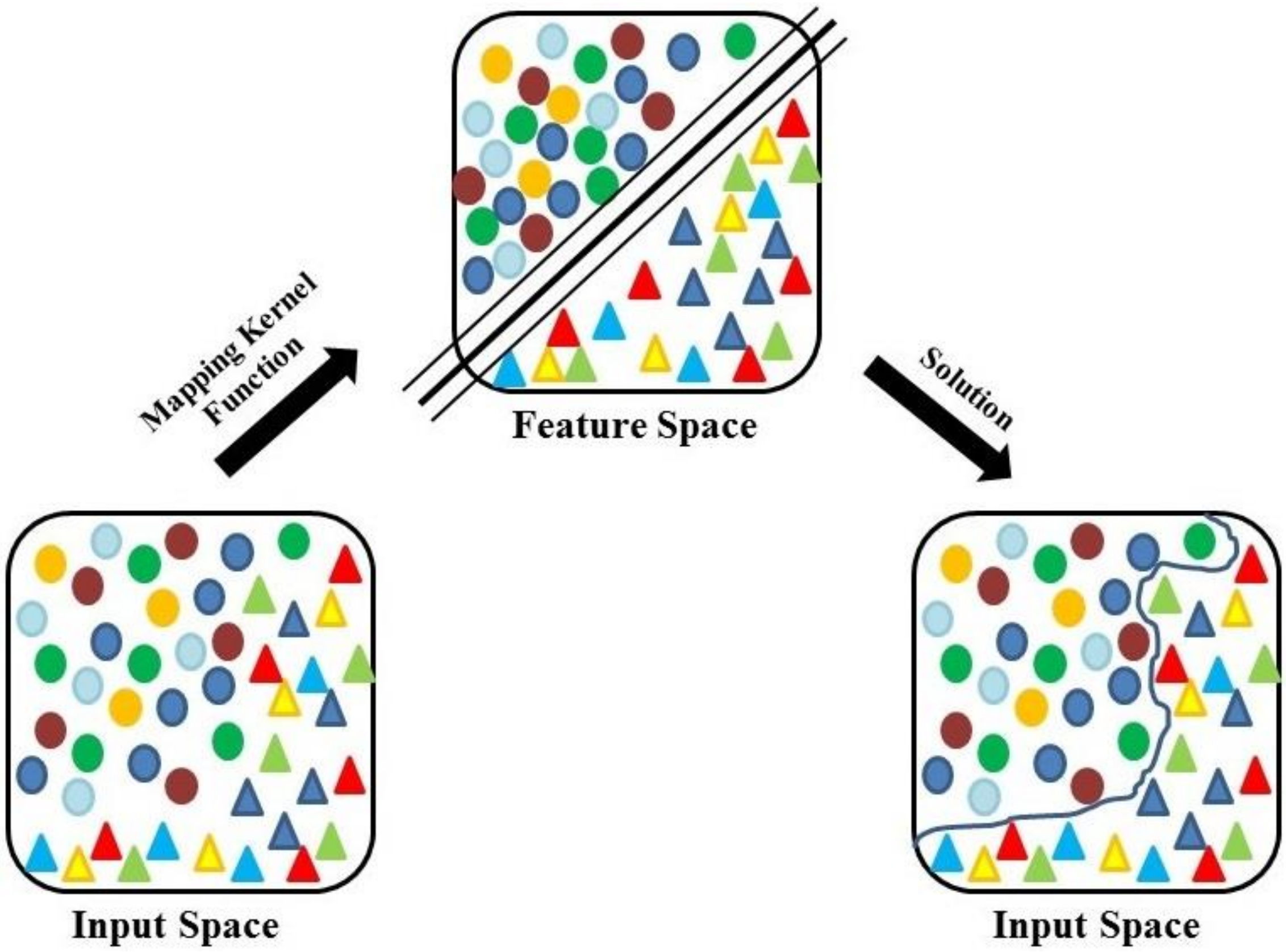

**Fig. 3.2 Principle of support vector machines.**
**Circles and triangles represent class annotations of the examples; multiple colors have been used to indicate higher dimensions besides the two planar dimensions.**

Besides linear classification, support vector machines also efficiently perform non-linear classification, based on the principle of maximal margin, through the use of kernels, as illustrated in Fig. 3.2. The mapping kernel function transforms the examples in the input space to an implicit feature space, based on the similarity between points in the input space. Some common kernel functions include homogeneous or inhomogeneous polynomials, Gaussian radial basis function, hyperbolic tangent, etc. [114].

In the classification application, given a set of landmark points with class annotations, a support vector machine builds a separating hyperplane between these class examples, such that the minimum distance of the hyperplane (margin) from the closest sample point is maximized. Through the process of maximizing the distance of the hyperplane from the closest points, a support vector machine determines the naturally most intuitive classification boundary.

### 3.2.3. Neural Networks

Neural networks [115] draw their inspiration from the human brain, which is a massively parallel, highly connected network of a large number neurons. Each of these neurons is an extremely simple processing unit but when combined together in a massive network, they are able to achieve very complex tasks.

These processing element in the human brain, the cell that is being referred to as a biological neuron in this discussion, makes dynamic connections to several other neurons, resulting in a well-structured interconnected topology of individual layers. The resulting network is called a neural network. An illustrative representation of the biological neuron is shown in Fig. 3.3. This neuron acts as inspiration for the computationally simple model of the artificial neuron, which combine together in different layers to create an artificial neural network. The computation model for the artificial neuron, and a possible topology for an artificial neural network is also illustrated.

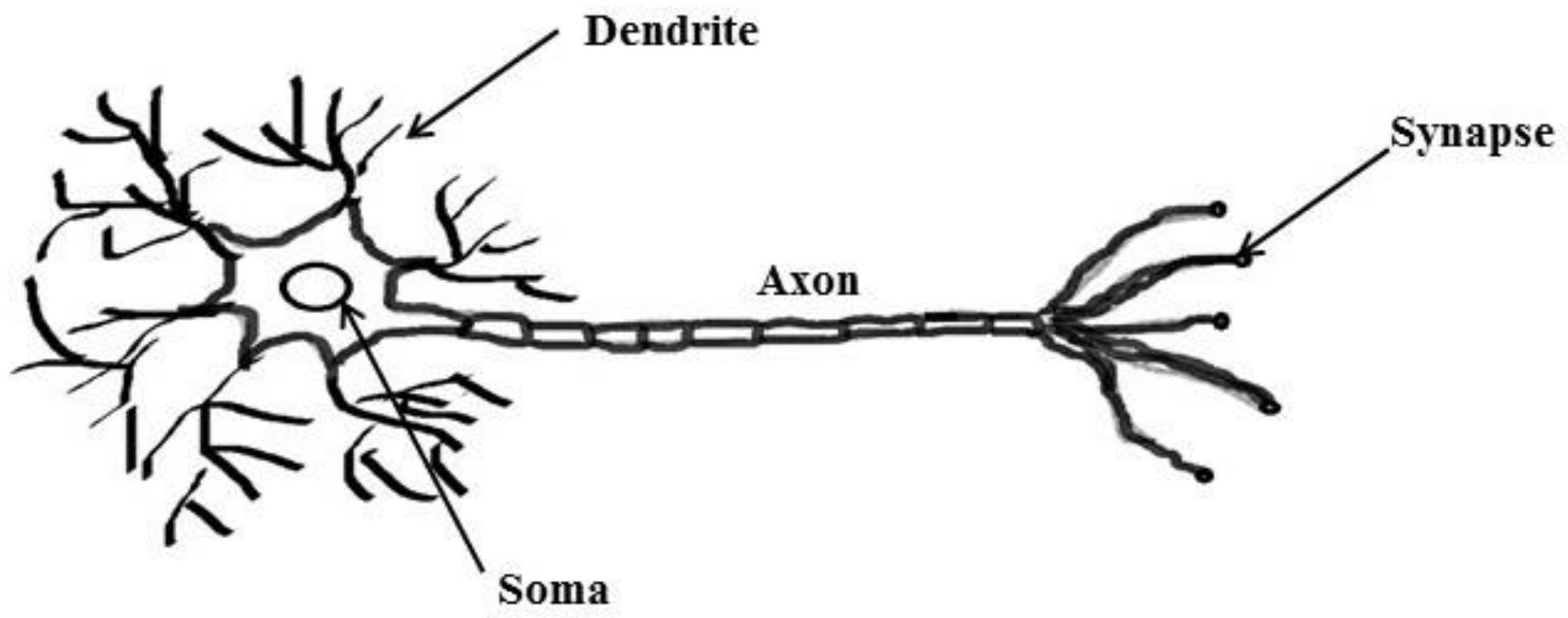

a.

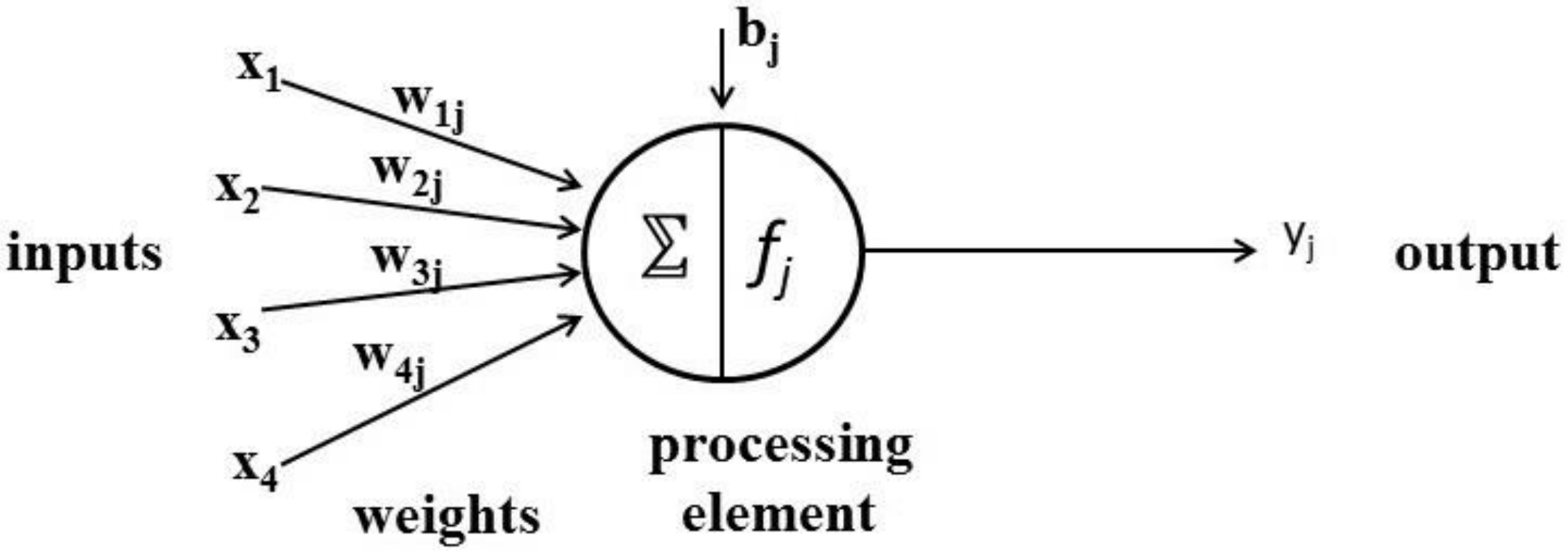

b.

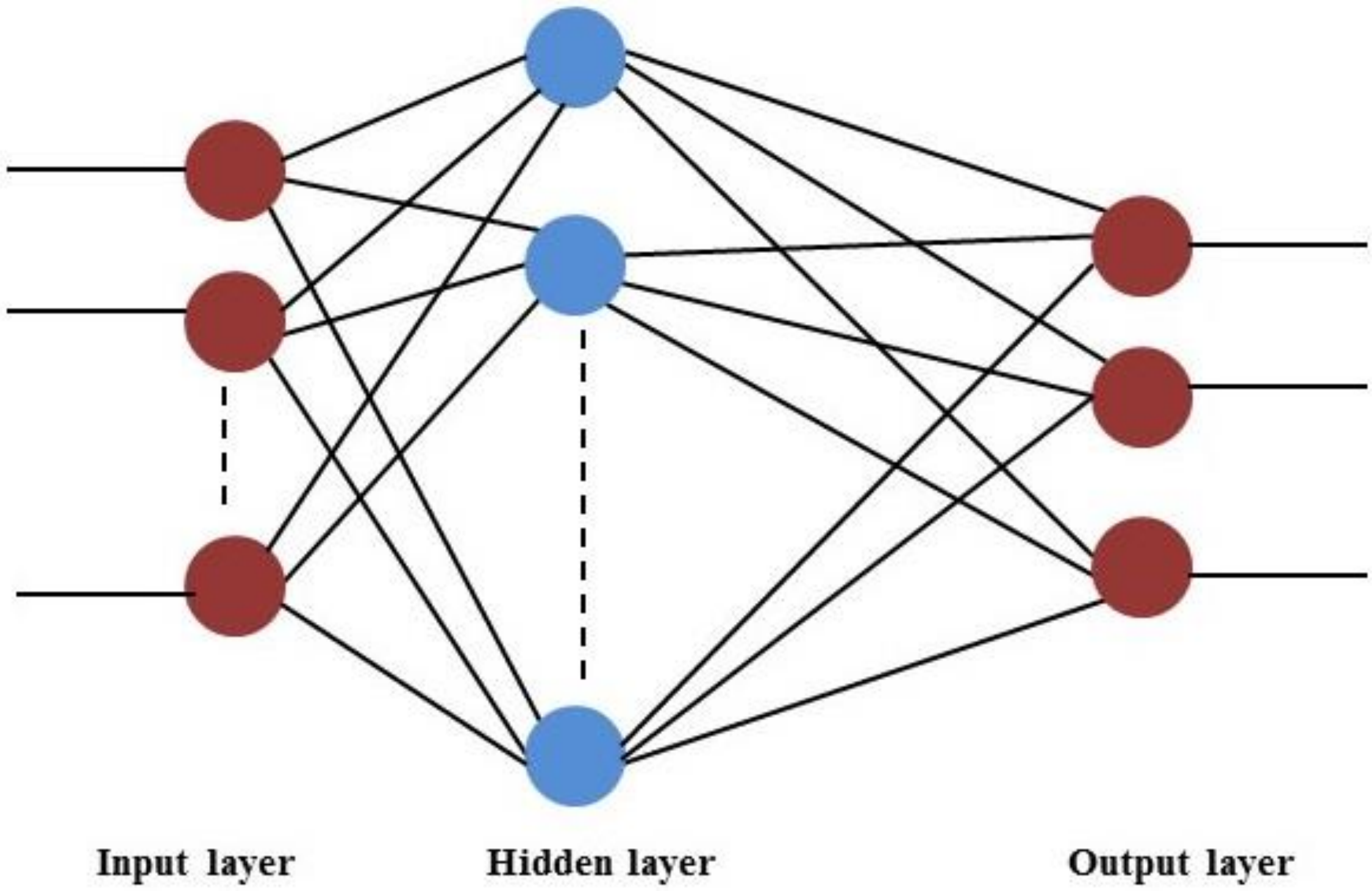

c.

**Fig. 3.3 The inspiration and topology of neural network.**
**a. Biological neuron; b. Computation model for an artificial neuron; c. An artificial neural network architecture.**

A typical neural network model consists of an input layer, several hidden layers in between and an output layer. Each neuron connection is associated with a weight which determines the effect of an input on the activation level of the neuron. The input signals are received by input neurons, and the neurons are then combined into a net input in the hidden layer using an integration function. The activation value of each neuron may be calculated by using a linear non-linear function of its aggregated input. The weights connecting neurons serve as the tunable parameters in the supervised learning model of the soft computing paradigm. The parameters in artificial neural networks may also be learnt using unsupervised or reinforcement learning.

Besides the learning of tunable parameters, the design of neural networks also requires a judicious selection of hyper-parameters, such as the number of layers and the number of processing units (also referred to as *nodes*) in each one of the layers.

Some well-known architectures for neural networks include feed-forward neural networks, radial basis function networks, self-organizing networks, recurrent networks, modular networks, associative networks, etc. These standard architectures also have further subcategories [115]. A detailed discussion on the various neural network architectures and the corresponding applications is beyond the scope of the present study.

#### 3.2.3.1. Deep Learning Networks

A recent advancement in neural networks has been the development of deep learning architectures [116]. These architectures have multiple layers of non-linear processing unit with a supervised or unsupervised learning mechanism at each layer for representation of features.

The several layers in the hierarchy of the architecture form different levels of features, ranging from local to global. The availability of high performance computing in recent times coupled with large and complex datasets for training has fostered the development of this paradigm.

Deep learning architectures have been applied across several domains, including speech recognition [117], image classification [118], drug discovery [119], etc. Recently, deep neural networks have also been applied in biometric face recognition [120].

### 3.2.4. Evolutionary Computation

Evolutionary computation [121] consists of optimization algorithms that have been inspired by the process of natural evolution. These algorithms have powerful search capabilities, making them a preferred choice for model development. The broad category of evolutionary algorithms may be further divided as genetic algorithms, genetic programming, ant colony optimization, honey bee colonies, artificial immune systems, swarm intelligence, etc. While a discussion on several variants of evolutionary computation algorithms will be outside the scope of the present work, a brief description of genetic algorithms is presented below as an example.

Genetic algorithms [122] is a search based heuristic technique inspired by the process of gene selection and survival in nature. The algorithm starts with a population of randomly generated genes and maintains a population of individual solutions for the application under consideration. The genes evolve iteratively through the application of a set of stochastic operators, such as *recombination*, *mutation* and *selection*. The simplest implementation of the recombination operator (also known as *one-point crossover*) selects two parents randomly from the population and after randomly choosing a position in each parent gene, exchanges the parts divided at the randomly chosen position, thus creating two new off-springs. The mutation randomly modifies a part of the gene while the selection operator decides the part of the population that will continue for the next generation, usually though a probabilistic selection process, based on a certain measure of fitness of the gene. An illustration of the phases and operators in genetic algorithm is shown in Fig. 3.4.

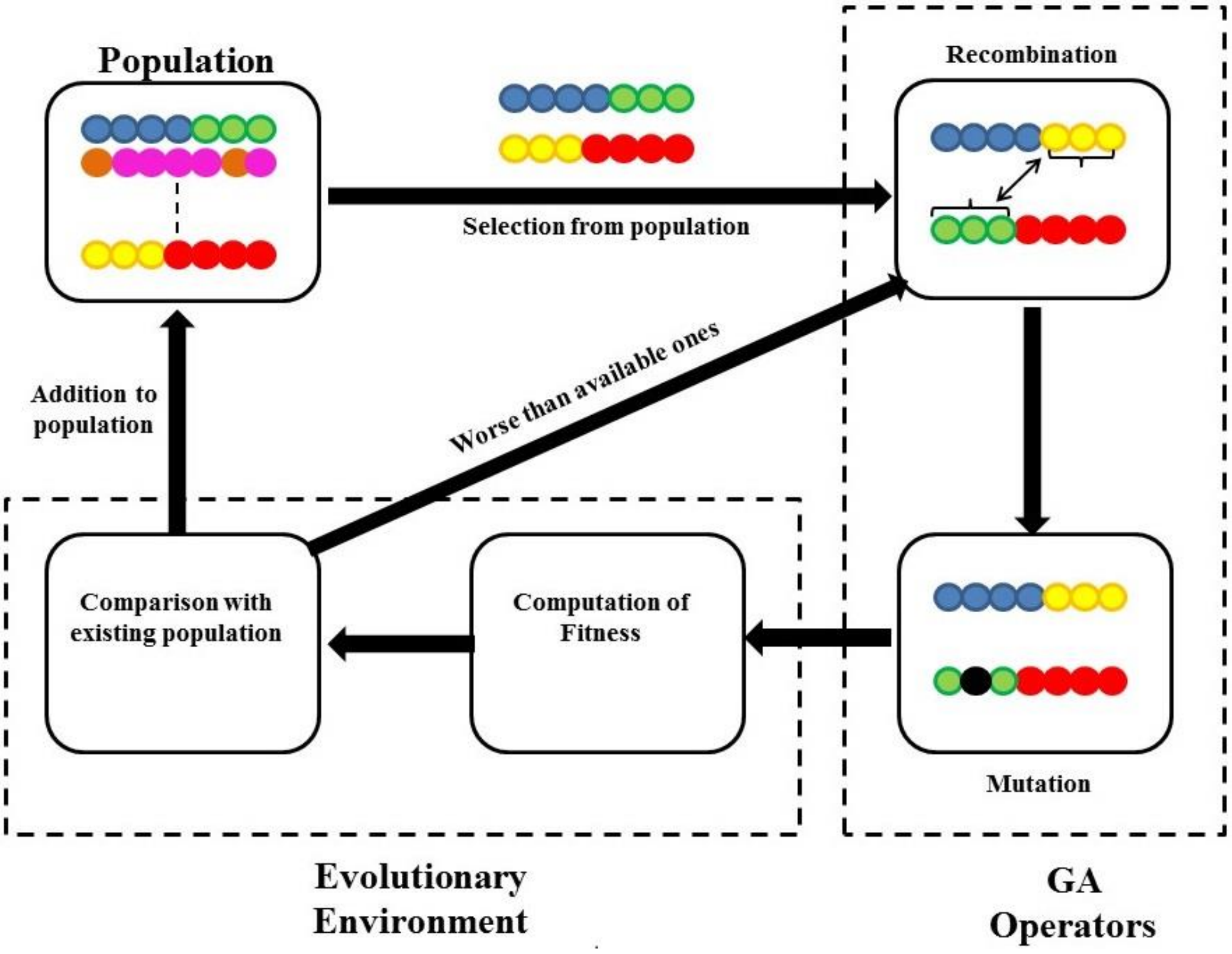

**Fig. 3.4 Principle of genetic algorithms.**

Evolutionary computation algorithms are inherently parallel in nature, making them amenable for implementation on high performance computing systems. These algorithms are often used towards solution of global optimization problems, e.g. routing and scheduling problems, and for the determination of hyper-parameters for other soft computing algorithms [123].

### 3.2.5. Fuzzy Logic

Fuzzy set theory [124] provides a mathematical framework for representing and handling uncertainty. A fuzzy set is represented through membership of its elements not as mere absence or presence of a member (0 or 1), as in Boolean sets, but in the continuous interval [0,1]. Thus, a fuzzy set, instead of being confined to *membership* or *non-membership* of an element, defines the *membership grade* for the element.

This fuzzy set theory affords to the fuzzy logic paradigm [125], the ability to deal with vagueness, imprecision, lack of information and partial truth, which are inherent in natural languages. Fuzzy logic provides the foundations for approximate reasoning with imprecise propositions using fuzzy set theory, by modeling *inexact* modes of reasoning that are fundamental to human decision making and expression, especially under uncertain and imprecise conditions.

While variables in traditional mathematics generally assume numerical values, fuzzy variables may assume non-numerical values, such as the linguistic notions of *low*, *medium* or *high*. Systems based on fuzzy logic are capable of function approximation, if an appropriate set of rules based on a set of linguistic labels and membership functions are either available from a human expert or may be developed.

The modules of a fuzzy system are illustratively shown in Fig. 3.5. These are:

i. *Fuzzifier*, that converts crisp inputs of the system to membership in fuzzy sets;
ii. *Rule base*, the set of fuzzy rules that represent approximate or imprecise reasoning;
iii. *Inference engine*, that provides the fuzzy output, based on the fuzzified inputs and the combination of relevant rules in the rule base; and,
iv. *Defuzzifier*, that converts the fuzzy output to a crisp output.

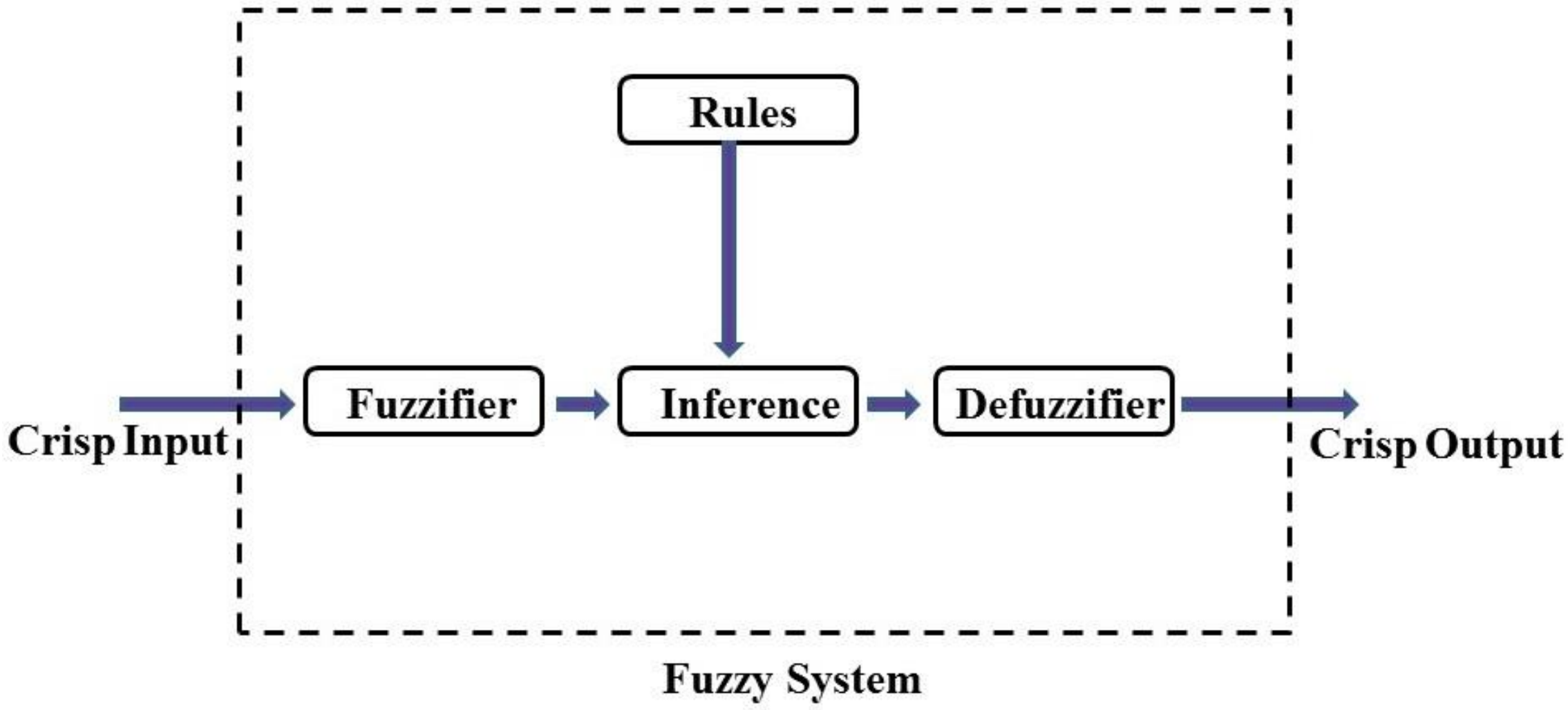

**Fig. 3.5 A fuzzy logic system.**

Fuzzy systems have been studied extensively for control engineering problems, such as altitude control of spacecraft [126], automatic transmission in vehicles [127], timing control in washing machines [128], etc.

## 3.3. Bootstrap Aggregation

Bootstrap aggregation [112] (also referred to as *bagging*) is a meta-algorithm that uses an ensemble to improve the stability and accuracy of algorithms based on the soft computing paradigm. Bootstrap aggregation provides an elegant solution to the problem of overfitting described in section 3.2 and reduces the variance in the solutions, leading to improvement in unstable procedures [129].

For the classification problem, given a training set, bootstrap aggregation consists of an ensemble of classifiers, each trained on a set drawn randomly from the given training set, with *uniform probability* and *with replacement*. This kind of a training sample, known as a *bootstrap sample*, results in a different model being fit to every classifier in the ensemble. The final decision may be based on majority voting between classifiers in the ensemble.

Bootstrap aggregation may also be used for regression problems, where the bootstrap samples are drawn in a way similar to the classification problem but the final outcome may be based on the average value of the regression output.

A variant of bootstrap aggregation has been used in the design of the algorithm in the present study, where the final decision is based on a *veto* (described in section 4.3) instead of a majority voting mechanism.

## 3.4. Combining Soft Computing Paradigms

Since most soft computing paradigms require a careful choice for both the tunable parameters, as well as hyper-parameters, hybrid approaches that combine more than one soft computing paradigms are often employed. In particular, instances where a soft computing paradigm is used for the determination of the value of meta-parameters for another soft computing algorithm have been extensively studied. These hybrid systems have also successfully been applied to several real world applications.

Some examples of hybrid approaches in soft computing systems [130] include:

i. Embedding a neural network as a part of a fuzzy system, to obtain fuzzy rules through training, instead of inferring the rules from the data [131];
ii. Use of fuzzy weights instead of crisp weights in a neural network [132];
iii. Use of evolutionary computing for determining the hyper-parameters in the determination of optimal architecture for a neuro-fuzzy system [133]; etc.

Several other combinations of two or more soft computing paradigms are possible in the design of a system. From the perspective of person recognition, soft computing paradigm can be used for feature selection, similarity score computation, fusion of multiple biometric traits, etc. Several computing paradigms may also be combined for effectively coalescing different tasks, e.g., combining the image quality information with the matching scores.

## 3.5. Summary

Pattern recognition is an important component of any biometric system and soft computing paradigms and given the approximate nature of biometric matching, soft computing approaches are well suited and have been applied for these applications [134, 135, 136, 137]. A comparison between different face recognition approaches demonstrates the use of soft computing paradigms not only for pattern recognition but also for feature extraction, reference determination, analysis and recognition [138].

This chapter presents an introduction to soft computing paradigms, introducing the terminology and the general steps in the implementation of systems based on soft computing approaches. A description of the various problem categories where soft computing approaches are used has been provided, along with the commonly accepted procedure for evaluation and testing of soft computing models.

The framework and application for some of the most commonly used soft computing paradigms, videlicet, logistic regression; support vector machines; neural networks; evolutionary computations; and fuzzy logic has also been provided.

The chapter also presents bootstrap aggregation as a meta-algorithm that employs an ensemble for improving the stability of solutions, followed by a brief discussion on combining different soft computing paradigms to build hybrid systems.

# CHAPTER 4

# ADAPTIVE INTEGRATION OF BIOMETRIC AND BIOGRAPHICAL INFORMATION

## 4.1. Introduction

One of the most pertinent problems in person recognition is to ensure that no individual is ascribed more than one identity by the system. The process of ascertaining such multiple identities and removing them from the system is commonly known as de-duplication of identities. While this may be rather easy to accomplish for small-scale systems, where the subjects may already well-known to all stake-holders and are generally well documented, the problem is very challenging for large scale systems, especially when the identity of the users are not pre-determined and there is an obvious incentive for a user to have multiple identities.

While biometrics are employed for de-duplication of individual identities, the collection of data at the time of enrolment often also includes soft biometric (e.g. gender, age, race) and biographical information (e.g. name, father's name, address) that identifies an individual in the operational world [13, 43, 44]. An integrated identification system that combines the available complimentary information judiciously, with the primary objective of automated de-duplication of identities in large scale systems, accurately and efficiently, has been proposed in this chapter.

## 4.2. Application Scenarios

De-duplication of identities may be considered under two different application scenarios:

i. Each subject has only one entry in the background database. As a new entry is presented, a matching process is adopted to determine if it is a potential duplicate.
ii. There are some subjects with multiple entries in the database and the de-duplication process is to eliminate the duplicates.

For the present study, the first application scenario is considered with a focus on large scale national identification applications.

### 4.2.1. Matching of Biometric Information

Various biometric traits, such as fingerprint, iris, face, etc. have been proposed for person recognition. Multi-modal biometrics combine complementary information from the same person to provide superior recognition performance. In a multi-modal biometric system, biometric traits can be fused at different levels: sensor level [63], feature level [67, 68], matching score level, rank level [62] and decision level [77]. Fusion at the score level presents the best trade-off between the amount of information available and the ease with which the information can be fused [78]. These techniques, along with the state of the art biometric recognition systems have already been discussed in the second chapter.

For the present study, state of the art Commercial Off-The-Shelf (COTS) matchers for fingerprints and face have been utilized under realistic de-duplication scenario. While the names of the COTS vendors are not being disclosed due to licensing restrictions, the matchers rank in the top three in recent NIST evaluations for fingerprint and face.

### 4.2.2. Matching of Biographical Information

The biographical information typically collected at the time for enrolment and that may be used for de-duplication consists of a person's name, their father's name and their address. While biographical information has been investigated to improve the de-duplication accuracy [22, 23], its use requires care due to errors that often creep in. The biographical similarity scores may be computed using several techniques, depending on the data type. For nominal data type (e.g., gender, race), the matching score may be binary ("same" or "not same"). Other textual data may be prone to variations for reasons already discussed in subsection 1.4.1, videlicet,

i. Errors introduced when information is entered in the system.
ii. Lack of a standard format and standard transliteration.
iii. Two individuals may share the same name.
iv. Change of address.

For such data, approximate string matching distance, such as the Levenshtein distance [36] is commonly used. For address, the geospatial distance may also be used as a possible metric.

The following distance measures have been considered for matching of biographical information:

i. Levenshtein [36];
ii. Damerau-Levenshtein [37]; and.
iii. Editor distances [38].

The Levenshtein distance between two strings is the minimum number of single-character insertion, deletion or substitution operations required to transform one string to the other. Damerau-Levenshtein distance also allows for transposition between two adjacent characters. Editor distance is similar to the Levenshtein distance except that substitutions are treated as two separate operations – insert and delete.

The edit distance is converted to a similarity as follows. First the edit distance is normalized in the [0,1] range by dividing it by the maximum possible edit distance between two strings of the same lengths as the given pair of strings. The corresponding similarity (matching score) between two strings is simply *(1 – normalized edit distance)*. The proposed similarity measure for the biographical information is the mean of similarities from Levenshtein, Damerau-Levenshtein, and Editor Distances.

### 4.2.3. Combining Biographical and Biometric Information

De-duplication of identities in large scale applications consists of matching of available biometric and biographical information and the subsequent fusion of the results from matching this information [13].

Even though experiments in [22, 23] have demonstrated that fusion of biographical information with biometrics can improve the de-duplication accuracy, ancillary information is not as reliable as biometric identifiers [24]. Especially, for the instances where individuals have deliberately provided inaccurate biographical information, the fusion of this information may not be expected to improve the accuracy of a system, but on the contrary, even deteriorate the accuracy. Therefore, an accurate and efficient de-duplication system therefore would require that biographical information be fused only when biometric information has reasonable been determined to be insufficient.

## 4.3. Design of Integrated Security System

The de-duplication process typically involves computation of matching scores for all biometric traits and biographical information of an enrolling subject against all subjects already enrolled in the database. These scores are then combined using a fusion strategy to make a decision whether the enrolling subject is a duplicate or not [70]. While fusion of complimentary information is helpful in reliable determination of identity, the process also introduces a significant computational expense in the determination of all matching scores followed by fusion. Besides, the effort in de-duplication keeps increasing with the increase in the size of the database as new subjects are enrolled.

### 4.3.1. Design Objectives

Although better results can be expected when more information is involved, the de-duplication in large scale applications suffers from complexity of computation. For each new query, current methods compare all its biometric and biographical information against the background database and fuse these scores to make a final decision. So, adding even one biometric trait results in a large increase of computational cost. It is of utmost importance, therefore, for the identification system to not only be robust and optimal but also efficient, through an appropriate and judicious combination of available biometric and biographical information.

With the dual objective of speeding up the de-duplication process while also improving the de-duplication accuracy, an algorithm for adaptive integration of identification information has been proposed in this chapter. The improvements in both imaging technology and matching algorithms have greatly enhanced the matching accuracy for unimodal systems. This diminishes the need for fusing all biometric and biographical information for all queries. Therefore, a sequential determination of duplicates through a soft computing approach is proposed as a universal model for predicting whether or not matching and fusion of additional biometric or biographical information is necessary. The proposed algorithm is designed in consideration of the following four objectives:

i. A real-time decision is made on whether additional pieces of evidence (biometric or biographical identifiers) would be required for fusion, thereby saving computation of matching scores for various identifiers.
ii. The decision is based on a simple evaluation, based on the information (matching scores) that have already been computed, without the explicit need for computation of any additional metric (e.g., quality).
iii. The missing pieces of information, caused by biometric system exceptions (e.g. failure to enroll) or otherwise (e.g. lack of documentation regarding proof of address) are automatically accounted for as the corresponding identifiers are never considered by the algorithm in the computation of the final fused score.
iv. The soft-biometric and biographical information, which is not very reliable for reasons listed earlier, is considered for fusion only for a small fraction of queries, where the available biometric identifiers are deemed to be insufficient.

The proposed system, therefore, not only improves efficiency by saving on the several computationally expensive operations required for matching of each individual identifier, but also further improves the de-duplication accuracy of an already highly-accurate system, by fusing ancillary information only for queries where the algorithm determines this to be necessary.

## 4.3.2. Proposed De-duplication Framework

The adaptive fusion algorithm is based on the principle of sequential fusion [62] as shown in Fig. 4.1. The order in which the biometric traits and biographical information is presented to the system is selected according to their discriminability.

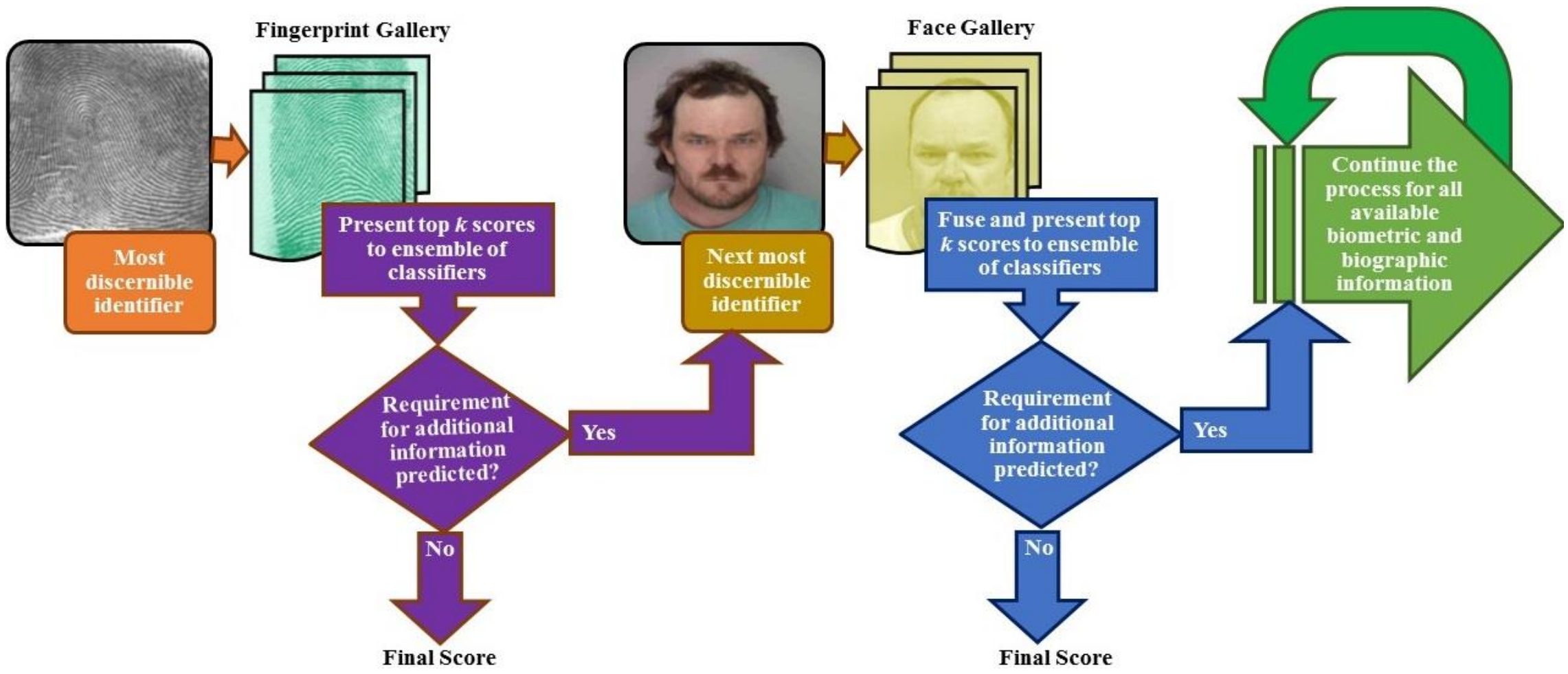

**Fig. 4.1 The proposed adaptive fusion framework.**
**The order of the biometric traits and biographical information could vary for different applications.**

After presenting a biometric trait or biographical information to the system, an ensemble of veto-wielding [139] logistic regression classifiers [140] is presented with the top k matching scores based on which the ensemble predicts whether the rank-1 score represents a genuine match. The decision threshold for the classifiers, based on the probability of misclassification of impostor as genuine, is set to an extremely low value.

In the veto-wielding ensemble, if all the classifiers in the ensemble agree that the rank-1 score represents a genuine match then no additional information is necessary for making a decision about the identity of the query.

The dual safeguard of using an abysmally small probability of misclassification of impostor as genuine as the threshold for prediction, coupled with the authority of each classifier to exercise a veto diminishes the chances of premature termination of fusion. A schematic diagram of this prediction model to decide whether additional information is required to determine a query's identity is shown in Fig. 4.2.

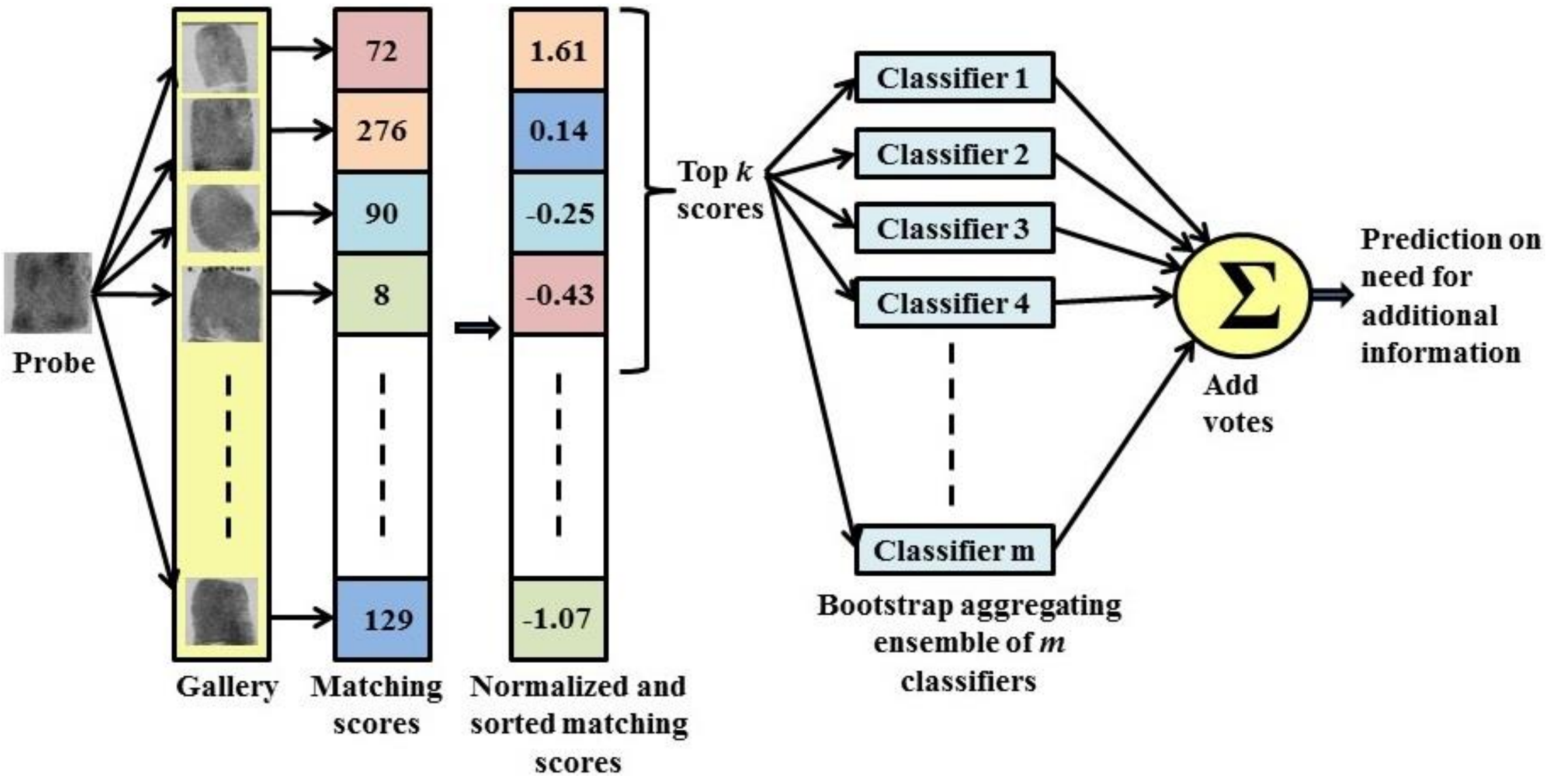

**Fig. 4.2 Prediction on requirement of matching and fusion of information from additional identifiers at each stage is made by an ensemble of classifiers. Fingerprint matching scores are used for illustration.**

### 4.3.3. Proposed Soft Computing Model for Adaptive Fusion

The proposed adaptive fusion algorithm evaluates whether or not additional pieces of evidence are required at each stage of fusion. In a prior work on matching of latent fingerprints [141], a strategy based on the "upper outlier" in the similarity score distribution, under the assumption that scores follow an exponential distribution.

Intuitively, the presence of a single upper outlier is a strong indication of a true mate at rank-1 (correct decision) because of the abysmally low probability of two events occurring simultaneously viz., a false match generates such high matching score that it is far removed from the rest of the similarity score distribution, and the true match generates such low score that it is within the distribution. This strategy has been used to determine whether additional mark-up is needed for the latent query.

However, it is not always possible to reliably determine a parametric distribution for the scores. Further, not all parametric score distributions are amenable to determining a single upper outlier. A detailed description of this algorithm has been provided in the appendix section A.3.

Towards a more generalized and universal approach to adaptive fusion of identifying information, the following soft computing approach is proposed as a model for predicting whether or not matching and fusion of additional biometric or biographical information is necessary.

The proposed model consists of an ensemble of *m* veto-wielding [139] logistic regression classifiers [140].

The logistic regression classifier has been described in subsection 3.2.1, but a concise outline is being presented here again for contextual articulacy.

The logistic function of a variable *z* is given by the following equation:

$$f(z)=\frac{1}{1+e^{-z}} \tag{20}$$

The value of the function $f(z)$ lies in (0,1), and for the purpose of two-class classification, is seen as the probability of alternate hypothesis ($H_1$: input represents class $w_1$); the probability of null hypothesis ($H_0$: input represents class $w_0$) is $(1 - f(z))$.

The input variable $z$ is a weighted sum of $k$ independent input features used for classification, as given by the equation

$$z = \sum_{i=0}^{k} \theta_i s_i \tag{21}$$

where $s_i$ is the $i^{th}$ feature and $\theta_i$ is the weight assigned to the $i^{th}$ feature. The bias term $s_0$ is set to unity.

The proposed algorithm uses the top $k$ highest currently known scores for the subject being de-duplicated as inputs $s_i$ to the logistic function for each classifier in the ensemble and the output $f(z)$ is interpreted as the probability that the rank-1 score does not represent a genuine match. A prediction on the requirement for consideration of additional information is made if this probability output is above a certain pre-determined decision threshold $\eta$.

The predictions of individual classifiers in the bootstrap aggregating [129] ensemble are combined for reaching a consensus decision. For most de-duplication scenarios, accuracy is a higher objective over efficiency. To ensure robustness, it is important to ensure that matching and fusion of information does not terminate prematurely. Each individual classifier in the ensemble, therefore, wields a veto and the process of further matching and fusion of available biometric and biographical information is terminated only when all classifiers in the ensemble unanimously [139] predict that additional information is not required to be considered. A high level algorithm of the proposed model is illustrated in Table 4.1.

**Table 4.1 High Level Description of the Proposed Adaptive Fusion Model**

**Input**: Training set $D = \left\{\left(\langle \boldsymbol{x}_{j\ raw}^{(i)} \rangle_{j=1}^{n}, y^{(i)}\right)\right\}$
**Output**: Whether or not $y^{(i)}$ is at rank-1
z-score normalize: $\boldsymbol{x}_j^{(i)} := \frac{x_{j\ raw}^{(i)} - mean\left(x_{j_{raw}}\right)}{standardDeviation\left(x_{j_{raw}}\right)}$
**Training**:
for $j \leftarrow 1\ to\ n$

$\boldsymbol{fusedScore}_j^{(i)} \leftarrow \frac{1}{j}\sum_1^j \boldsymbol{x}_j^{(i)}$

$\left(\boldsymbol{fusedScore}_j^{(i)}\right)_k \leftarrow$ top $k$ scores from $\boldsymbol{fusedScore}_j^{(i)}$

Train ensemble of $m$ logistic regression classifiers $h_j\left(\left(\boldsymbol{fusedScore}_j\right)_k\right)$

to predict $\begin{cases} 1, & y^{(i)}\ is\ not\ at\ rank-1 \\ 0, & otherwise \end{cases}$

end
**Implementation**:
Set logistic regression prediction threshold $\eta$ to an arbitrarily low value (e.g., $10^{-6}$)
Initialize: $j \leftarrow 0$
do:

$j \leftarrow j + 1$

$\boldsymbol{fusedScore}_j \leftarrow \frac{1}{j}\sum_1^j \boldsymbol{x}_j$

$\left(\boldsymbol{fusedScore}_j\right)_k \leftarrow$ top $k$ scores from $\boldsymbol{fusedScore}_j$

while $\sum_m h_j\left(\left(\boldsymbol{fusedScore}_j\right)_k\right) > 0$ and $j < n$
**Result**: Duplicate, if exists, is at rank-1.

Here, $\boldsymbol{x}_{j\ raw}^{(i)}$ is the vector consisting of raw (not normalized) matching scores for training example (probe) $i$ against the gallery for biometric trait or biographical information $j$. $y^{(i)}$ is the true identity of the subject. The number of identifiers (biometric trait or biographical information) available per subject is $n$.

The number of classifiers $m$ in the ensemble is chosen based on the trade-off between desired robustness and computational effort required for prediction. The value of $k$, representing the number of highest ranking scores is appropriately chosen to ensure that the system does not underfit or overfit the training data. The traits are chosen in order, starting with one that has the best discriminability when compared to the ground truth for the training data, *i.e.*, the trait that predicts the requirement for matching and fusion of additional information for the fewest number of subjects, without incorrectly predicting otherwise for any subject where the rank-1 score does not represent a genuine match.

A separate ensemble is trained for each stage of the algorithm using bootstrap aggregation with the same number of training examples for each classifier as the size of the training set, but randomly chosen with replacement. For example, if the available identifiers include a fingerprint, an iris and name and the training data indicates that fingerprint has the best discriminability for the algorithm, followed by iris and finally name, the ensemble for the first stage is trained using the fingerprint scores from the training set and for the second stage, using the fused scores for fingerprint and iris from the training set.

A comparative evaluation of the proposed soft computing model for adaptive fusion is presented in the next chapter.

## 4.4. Summary

A fundamental drawback of current fusion strategies is that scores from all the matchers are fused to get the final score. If some biometric or biographical information is not available, the de-duplication process has to be terminated or gets deteriorated. Besides, requirement for more information also means a larger cost of computation.

Most large scale de-duplication systems are deployed with the support of high performance parallel computation infrastructure at the backend [25]. However, the enormous amount of computation required, given the sheer volume of population, pushes processing delays beyond acceptable limits [142]. Besides, while the de-duplication accuracy for these identification systems is high, even a small percentage error translates to big absolute numbers on a large population [26], bringing to question the credibility of claims regarding “unique identification” [27]. Since de-duplication involves matching of the identity being enrolled with all previously enrolled identities, the delays get pronounced with time as more individuals get enrolled on the system.

This chapter presents a framework for identity de-duplication and a corresponding soft computing model based on adaptive fusion of identification information. The proposed system offers several advantages over existing de-duplication systems, not only in terms of computational efficiency, accuracy but also increased user convenience.

# CHAPTER 5

# EVALUATION OF ADAPTIVE INTEGRATED SOFT COMPUTING SECURITY MODEL

## 5.1. Description of Population and Datasets

As described in subsection 2.6.4, benchmark datasets are used for comparative evaluation of algorithms. The biometric database used for the experiment consists of two fingerprint images for each of the 27,000 subjects from NIST Special Database 14 [35] and two face images for each of the 27,000 subjects from the PCSO [29] dataset. Each subject in the NIST 14 database was randomly assigned a face in PCSO database to create a virtual bi-modal database of 27,000 subjects. For fingerprints, the first impression of each finger is used to form the gallery while the second impression is used as probe. For face, the image at a younger age was used as the gallery and a later image captured when the subject had grown older was used as probe. The biometric databases used here, as expected, have been anonymized, so there is no subject name associated with the images.

Since no large scale benchmark datasets for biographical information is available, the biographical information to each subject was assigned first using the gender information in the face database and then randomly drawing the first name, last name and father's name with the same likelihood as an actual national population, by mimicking the statistics from the US Census [30].

### 5.1.1. Multiple Instances of Biographical Data

In practice, biographical data across different instances of the same individual are not identical because of lack of standardization and possible human data entry errors. To replicate these dissimilarities between different instances of biographical data (identifiers), a statistical model that embodies the characteristics of variations and human errors was created. For the development of this model and to study these characteristics, several human operators were asked to enter the information typically present in identification documents. The crowdsourcing experiment [143] was conducted under payment conditions and time constraints similar to actual data entry applications.

## 5.2. Evaluation Metrics

The evaluation metrics for comparative evaluation in classical biometric recognition systems are have been presented in subsection 2.6.1. These error rates have been well studied [144]. For the de-duplication scenario, false de-duplication rate (FDDR) and false non-duplication rate (FNDR) for the dynamic matching error has been proposed in [145]. Even though FDDR and FNDR correctly model the de-duplication errors in a practical scenario, there are certain constraints that these parameters have, especially in the context of comparative evaluation on benchmark datasets:

i. The de-duplication error rates FDDR and FNDR are dependent on the sequence in which subjects are presented to the system. Therefore, for the same algorithm and on the same benchmark dataset, the results would be inconsistent for the purpose of comparison as they would depend on the sequence in which the subjects are presented to the system.

ii. The previous attempts [22, 23] on combination of biometric and biographical information for de-duplication have employed the CMC curves to estimate system accuracy. Using another metric in the present work would make results of the present study incompatible for comparison with prior art.

iii. The benchmark score datasets, e.g. [33], contain matching scores for all subjects in the gallery against all probes that consist of another impression of the same trait for the same set of subjects. However, these score datasets do not contain within-gallery matching scores, making it unfeasible to estimate FDDR and FNDR on these datasets.

Considering the above reasons, CMC curves have been used for estimating system accuracy in the present study.

## 5.3. Matching of Unimodal Information

Two state of the art commercial matchers (COTS A for fingerprint and COTS B for face) were used to compute the biometric match scores and the average impact of Levenshtein, Damerau-Levenshtein, and Editor Distances described in subsection 4.2.2 was used for matching of biographical information.

The average impact of edit distances proposed in subsection 4.2.2 outperforms the Levenshtein distance which was used in previous work [23]. A comparison of matching name and father's name and subsequent fusion of biographical scores on the dataset of 27,000 subjects, also used in later experimental analysis is shown in Fig. 5.1. The proposed algorithm for matching of biographical information outperforms the Levenshtein distance by about 4% on the data derived from US census information [30].

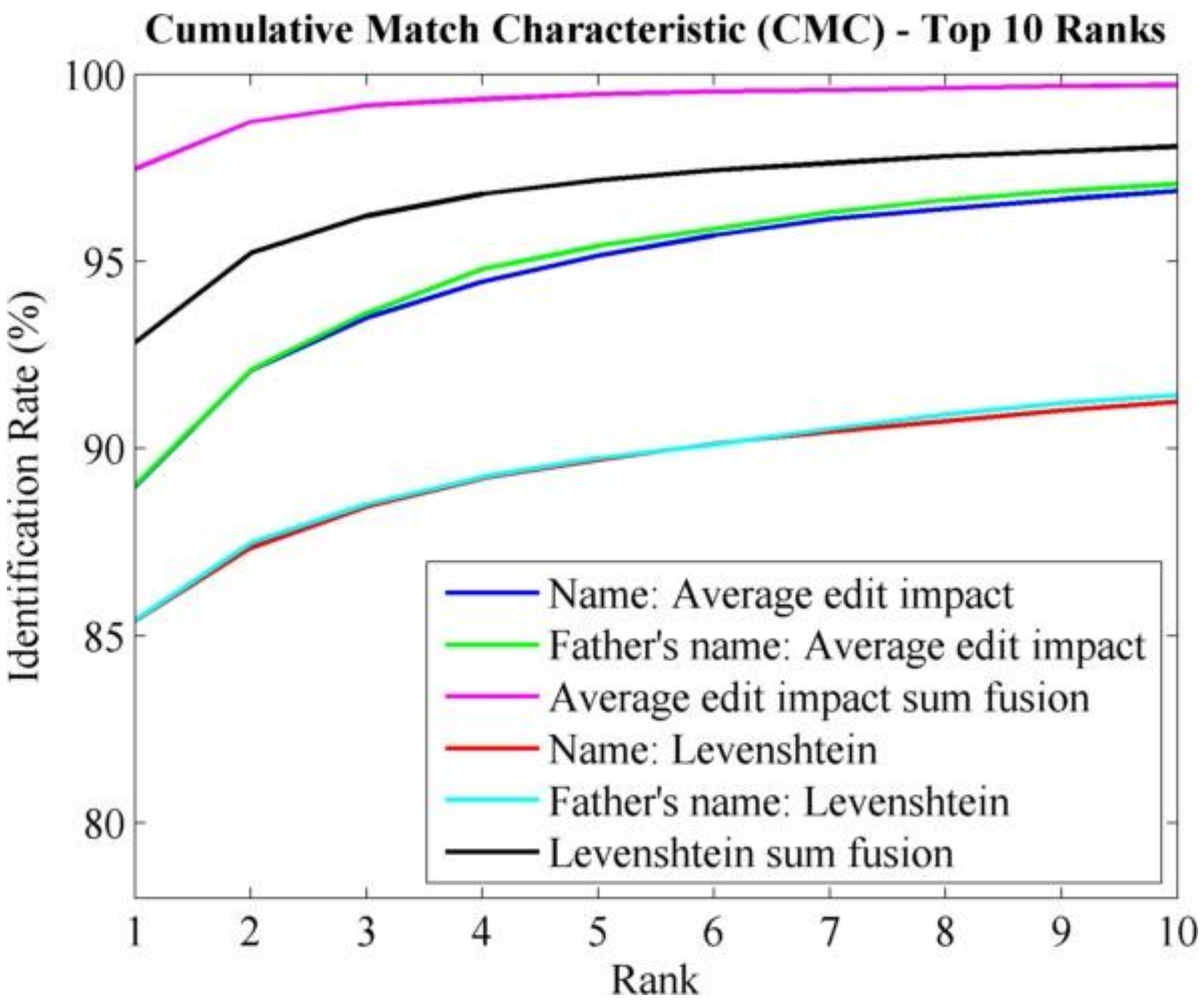

**Fig. 5.1 CMC curve representing the identification rate for top 10 ranks for biographical information using (i) average impact of edit distances, and (ii) Levenshtein distance.**

The cumulative match characteristic (CMC) curves that present a comparison between the various unimodal identifiers, videlicet,

i. Fingerprint;
ii. Face;
iii. Name; and,
iv. Father’s name

are shown in Fig. 5.2.

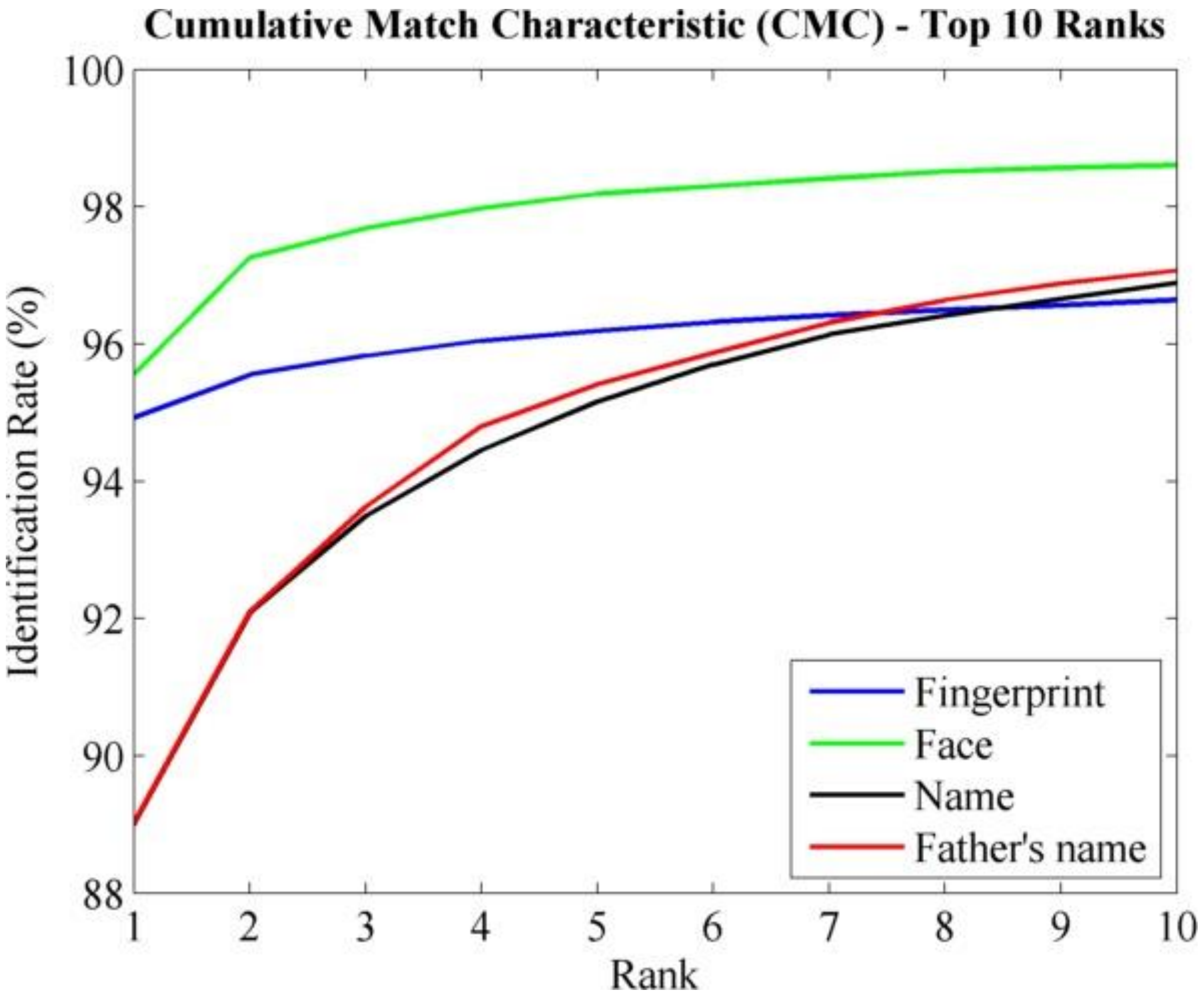

**Fig. 5.2 CMC curve representing the identification rate for top 10 ranks for unimodal identifiers.**

Two examples where the rank-1 fingerprint score represents a genuine match are shown in Fig. 5.3, while another two examples where fingerprint matching is not sufficient, i.e., the rank-1 fingerprint score does not represent a genuine match are shown in Fig. 5.4.

Two examples where the rank-1 face score represents a genuine match are shown in Fig. 5.5, while another set of two examples where face matching is not sufficient are shown in Fig. 5.6. The two face mugshot images per subject have varying time lapses; the age of the subject at the time of image acquisition is noted in the caption.

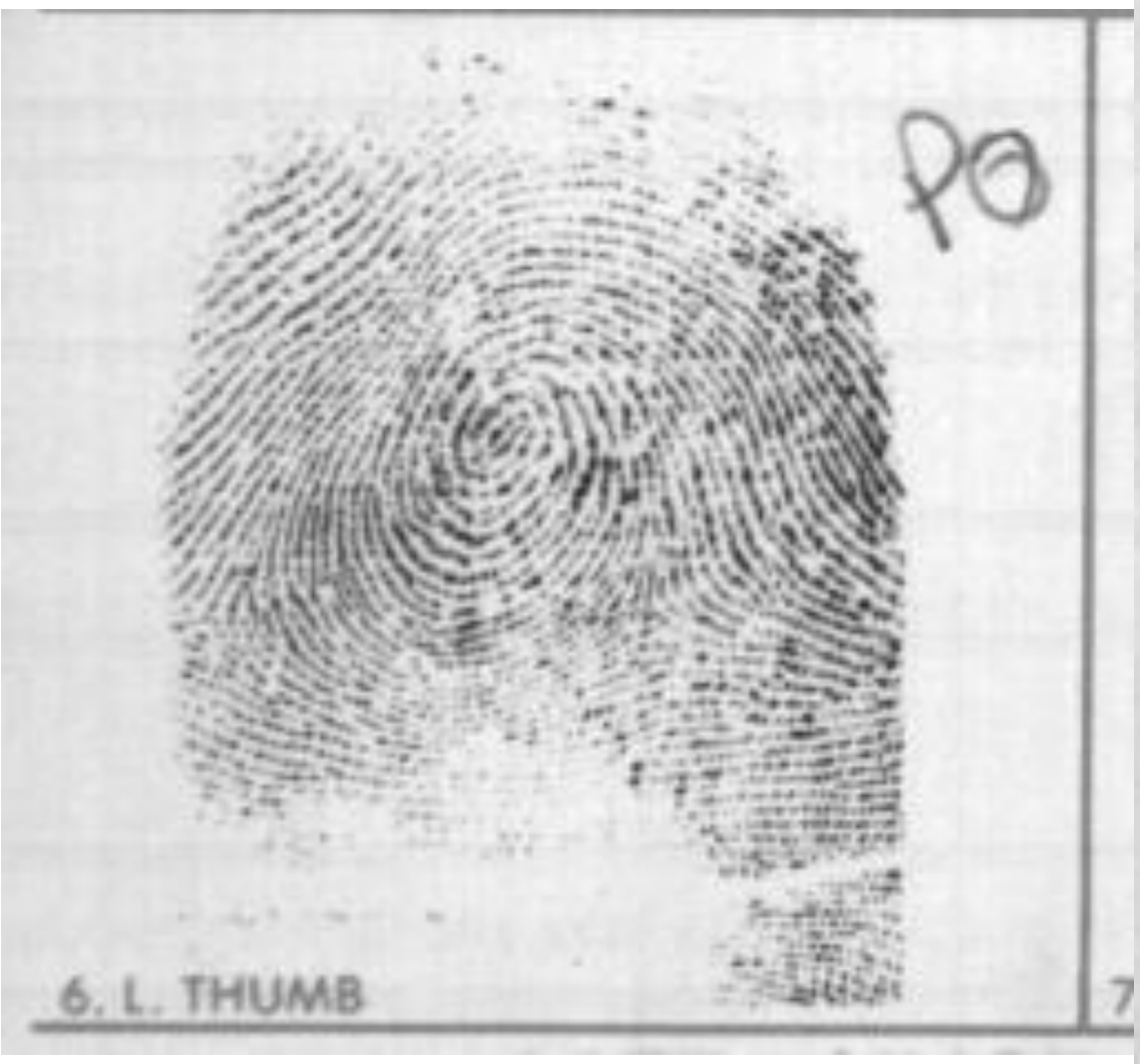

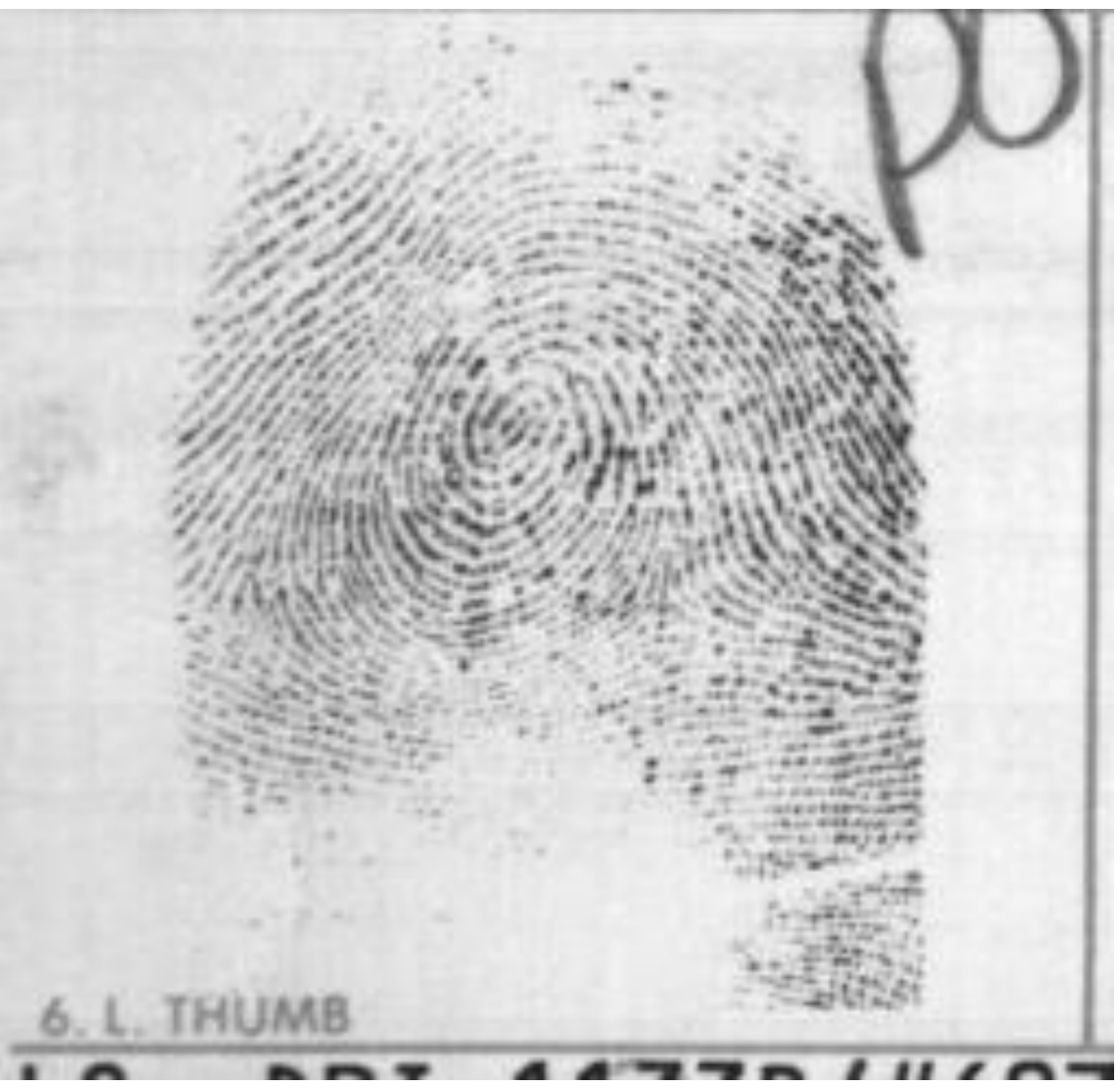

a. b.

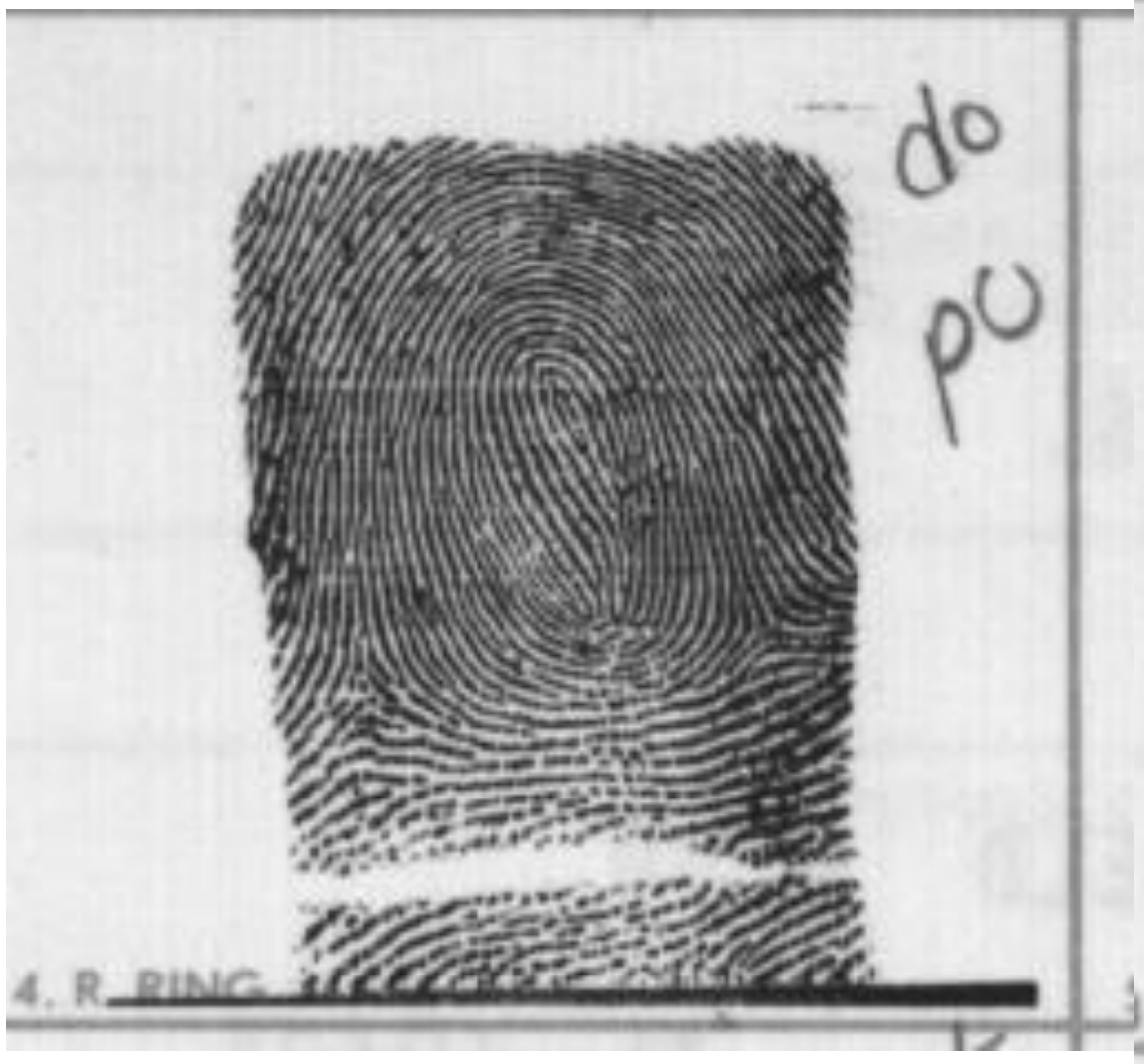

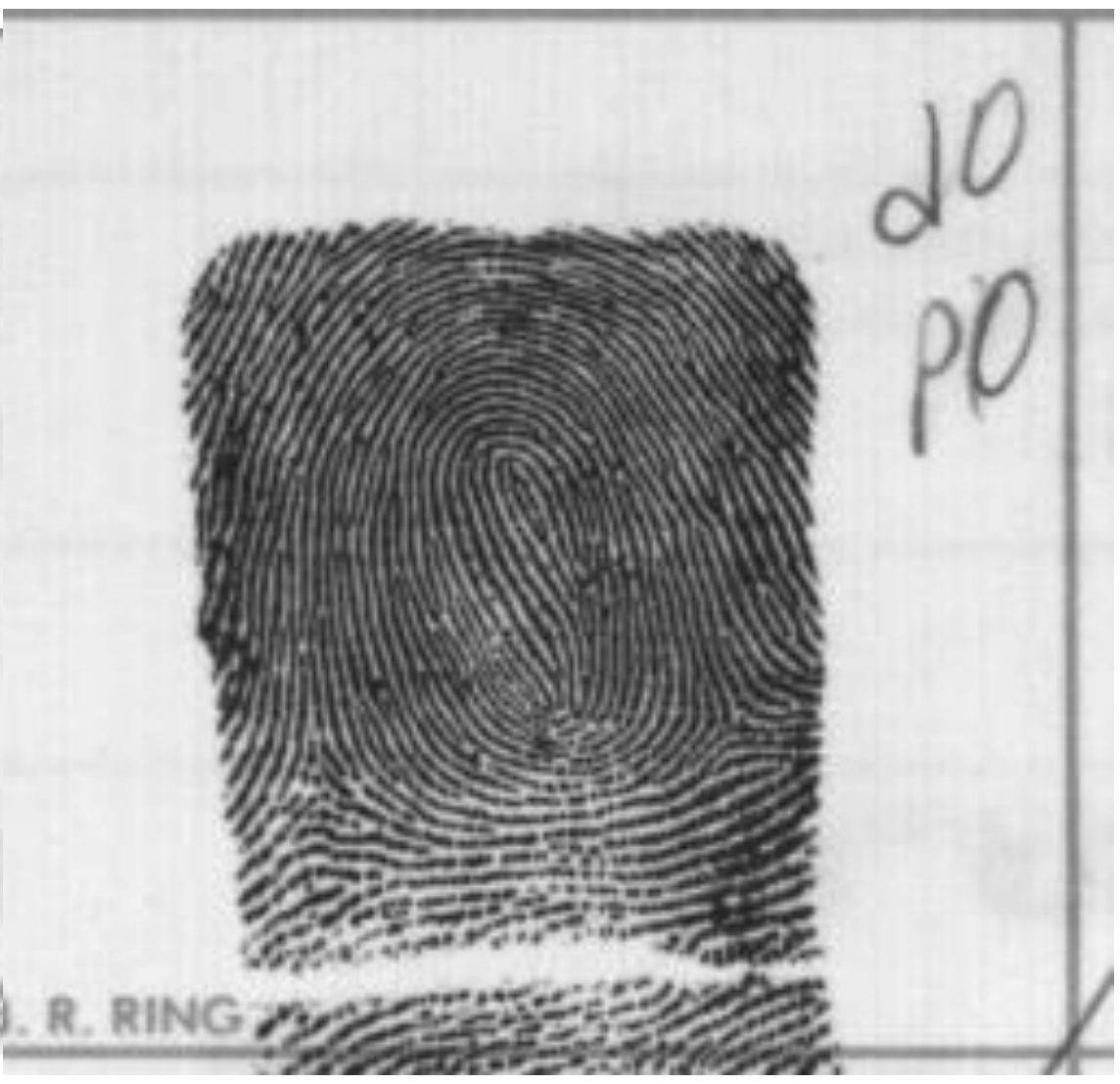

c. d.

**Fig. 5.3 Examples of successful fingerprint match where rank-1 scores represent a genuine match.**
**a. Probe image for fingerprint example 1; b. Rank-1 gallery image for fingerprint example 1.**
**c. Probe image for fingerprint example 2; d. Rank-1 gallery image for fingerprint example 2.**

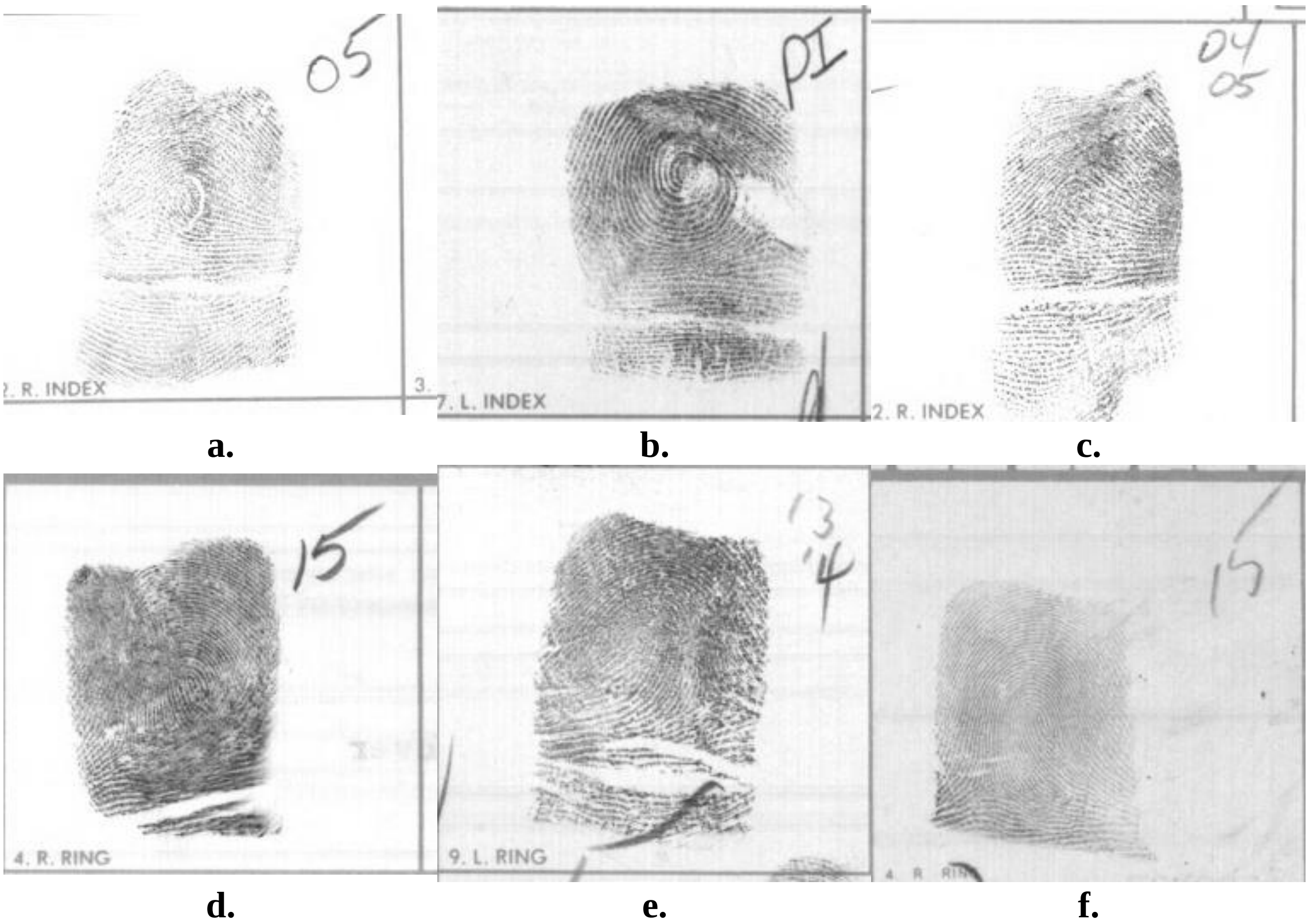

**Fig. 5.4 Examples where fingerprint match alone is not sufficient.**
**a. Probe image for fingerprint example 3; b. Rank-1 gallery image for fingerprint example 3; c. Gallery image of genuine subject for fingerprint example 3 retrieved at rank 24,684.**
**d. Probe image for fingerprint example 4; e. Rank-1 gallery image for fingerprint example 4; f. Gallery image of genuine subject for fingerprint example 4 retrieved at rank 25,268.**

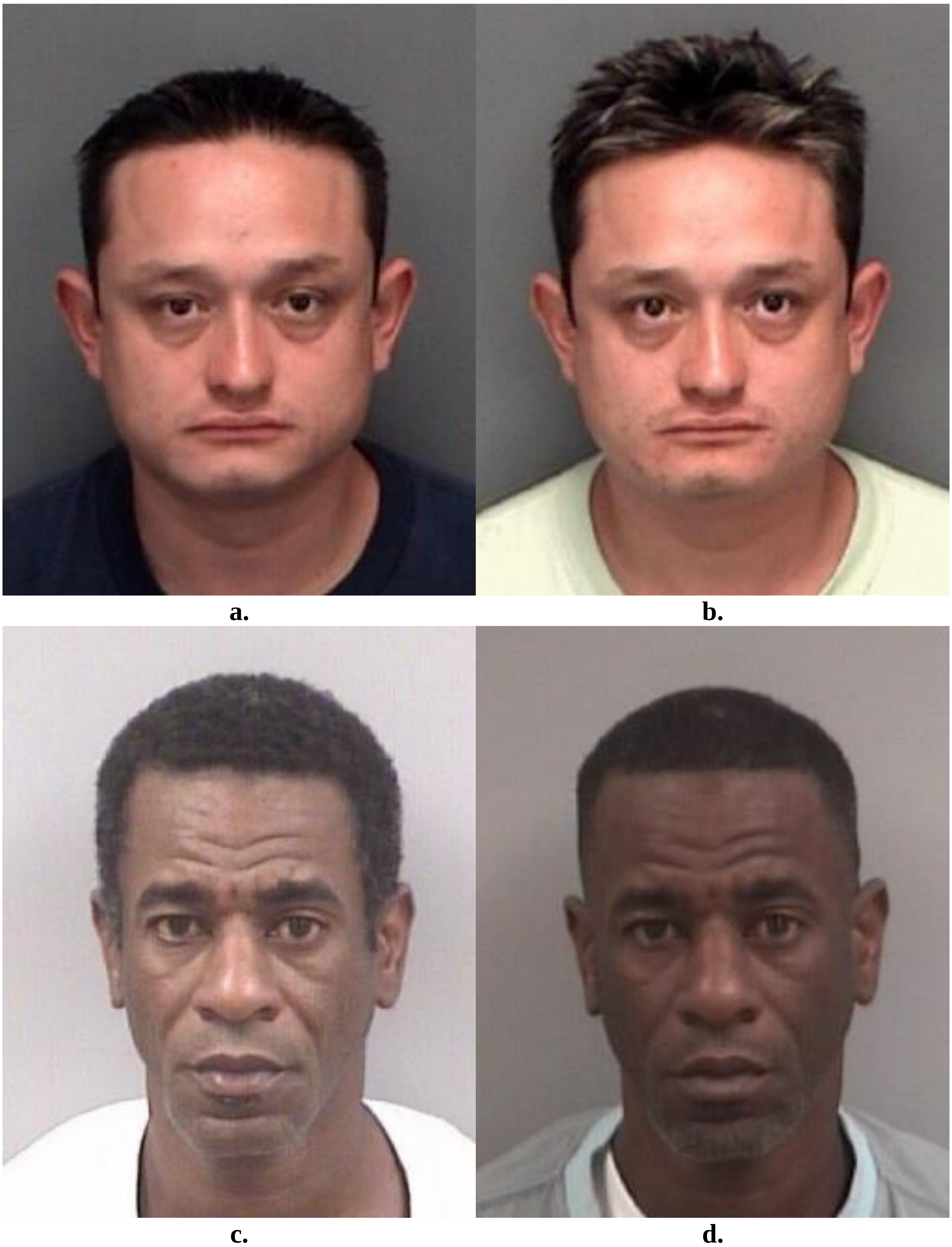

**Fig. 5.5 Examples of successful face match where rank-1 scores represent a genuine match.**
**a. Probe image, age 31.0 years for face example 1; b. Rank-1 gallery image, age 28.6 years for face example 1.**
**c. Probe image, age 44.5 years for face example 2; d. Rank-1 gallery image, age 43.0 years for face example 2.**

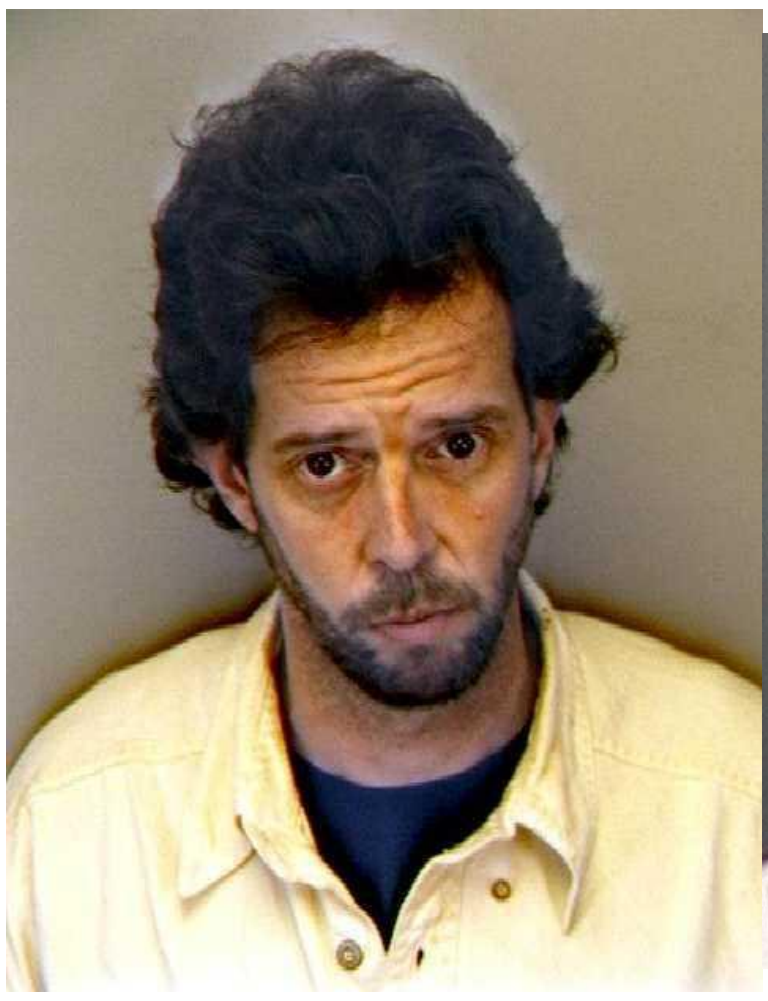
**a.**

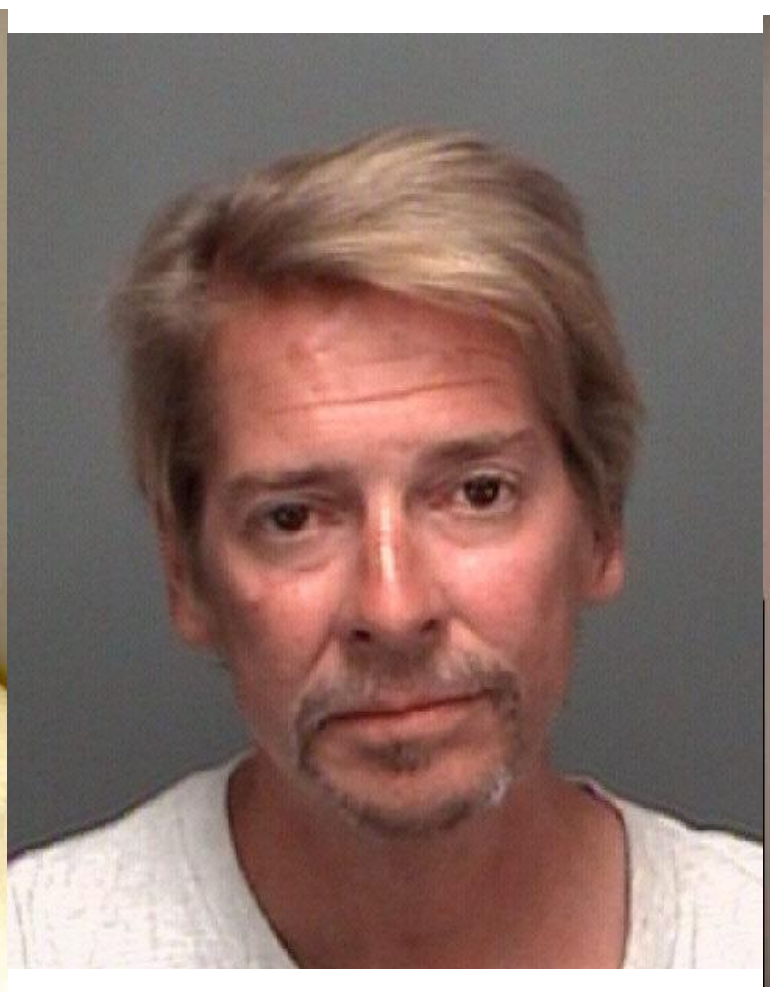
**b.**

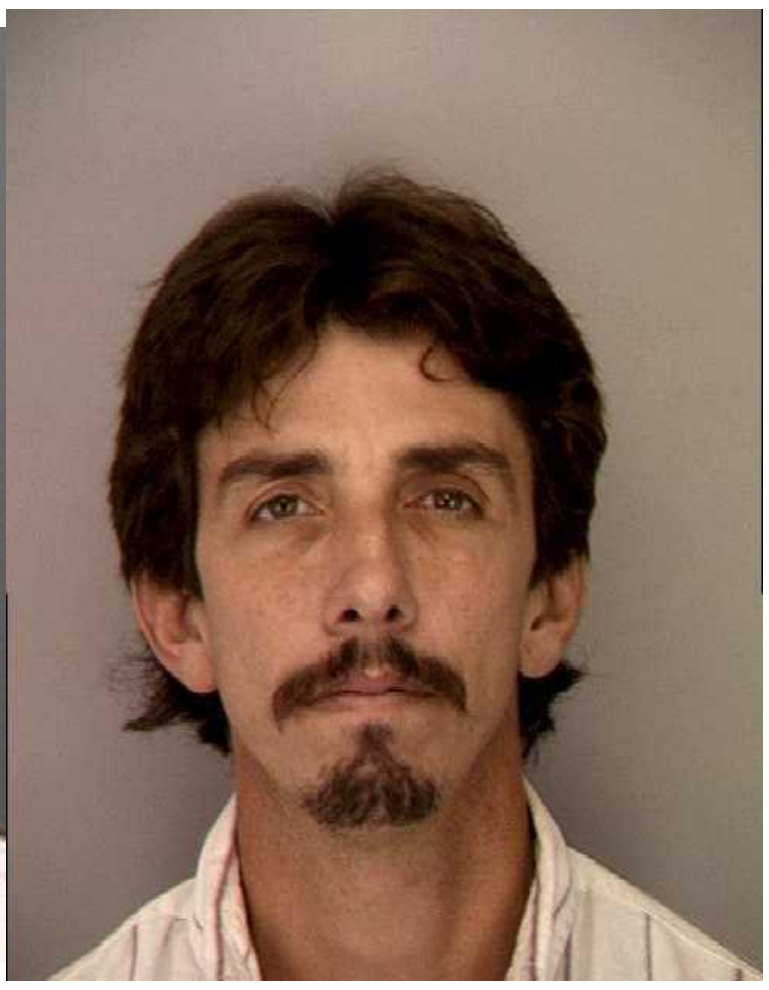
**c.**

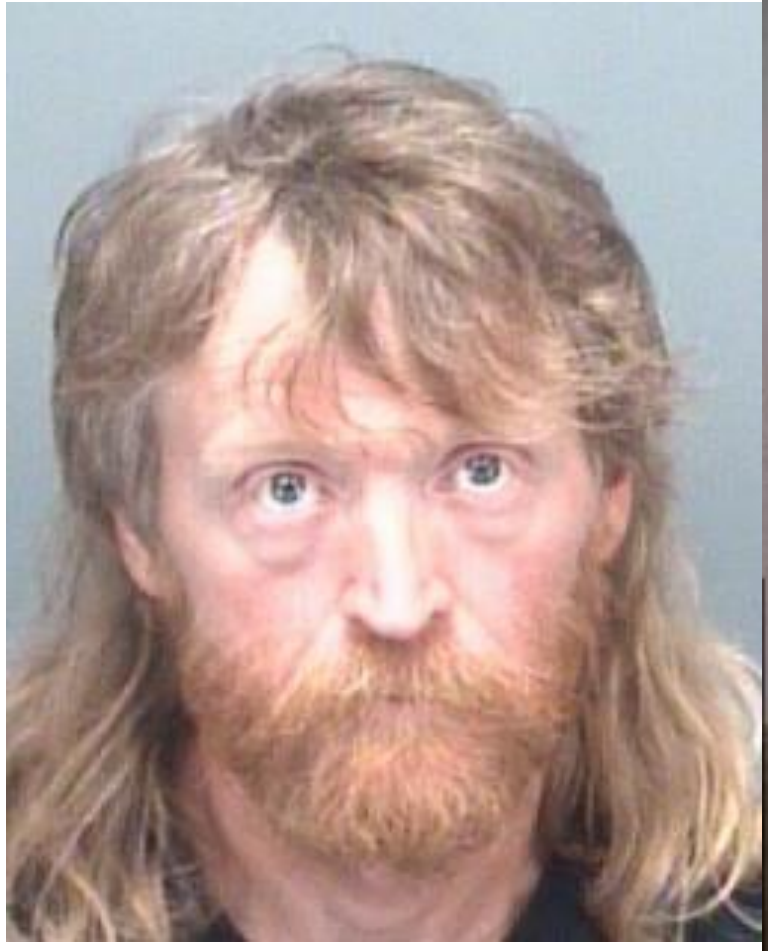
**d.**

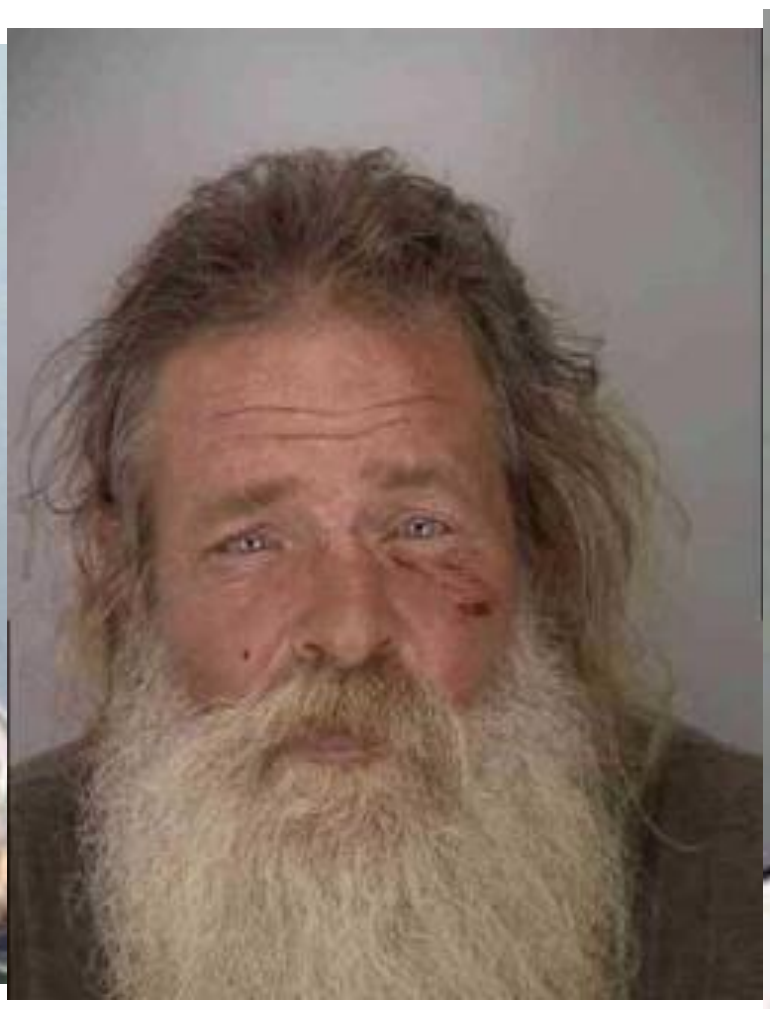
**e.**

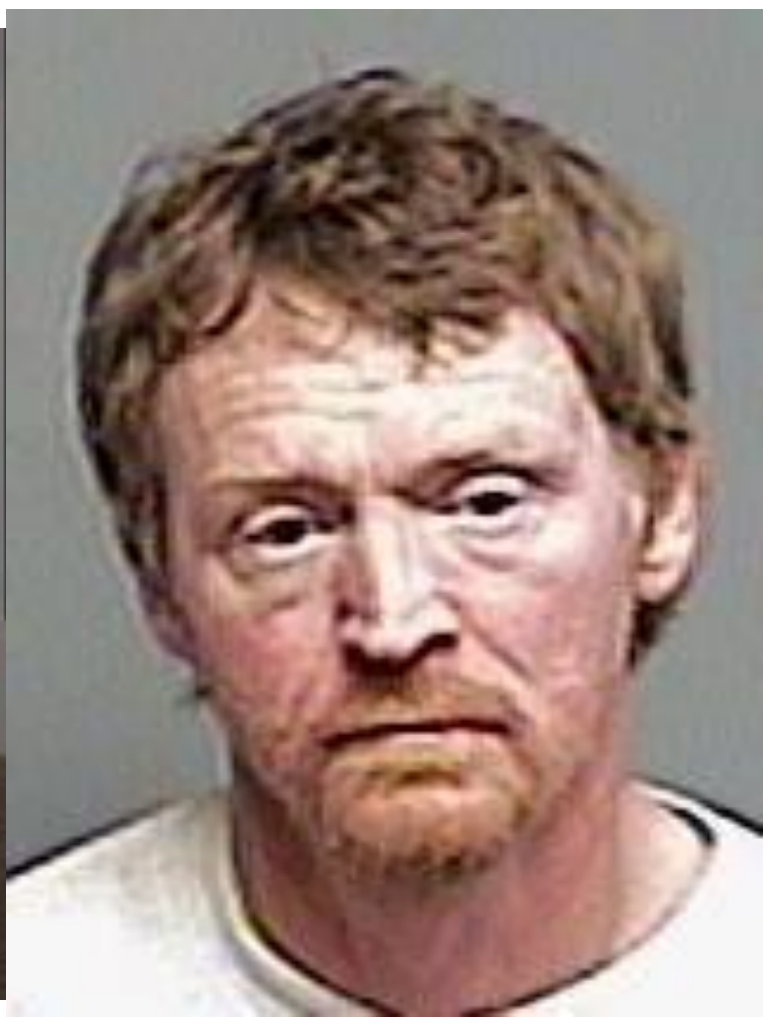
**f.**

**Fig. 5.6 Examples where face match alone is not sufficient.**
**a. Probe image, age 41.1 years for face example 3; b. Rank-1 gallery image, age 43.0 years for face example 3; c. Gallery image of genuine subject, age 38.9 years for face example 3 retrieved at rank 18,657.**
**d. Probe image, age 47.6 years for face example 4; e. Rank-1 gallery image, age 47.4 years for face example 4; f. Gallery image of genuine subject, age 44.5 years for face example 4 retrieved at rank 16,372.**

## 5.4. Fusion of Biometric and Biographical Information

As already discussed, some values for biometric matching scores are upper outliers, making them exceedingly large as compared to the rest of the distribution. This causes min-max normalization to not produce good results, because outliers cause the rest of the scores to be reduced to a very small range. To avoid upper outlier scores from compressing a majority of the biometric score distribution to a small range, z-score normalization was used. The normalized scores may be computed as

$$s' = \frac{s - \mu}{\sigma} \quad (22)$$

where *s'* is the normalized similarity score, *s* is the original raw score, $\mu$ and $\sigma$ are, respectively, the mean and standard deviation of the distribution of *s*, either known or estimated from the data [70]. Fusion of scores for all scenarios discussed in this and the subsequent sections has been performed on z-score normalized scores by employing the sum fusion rule [62].

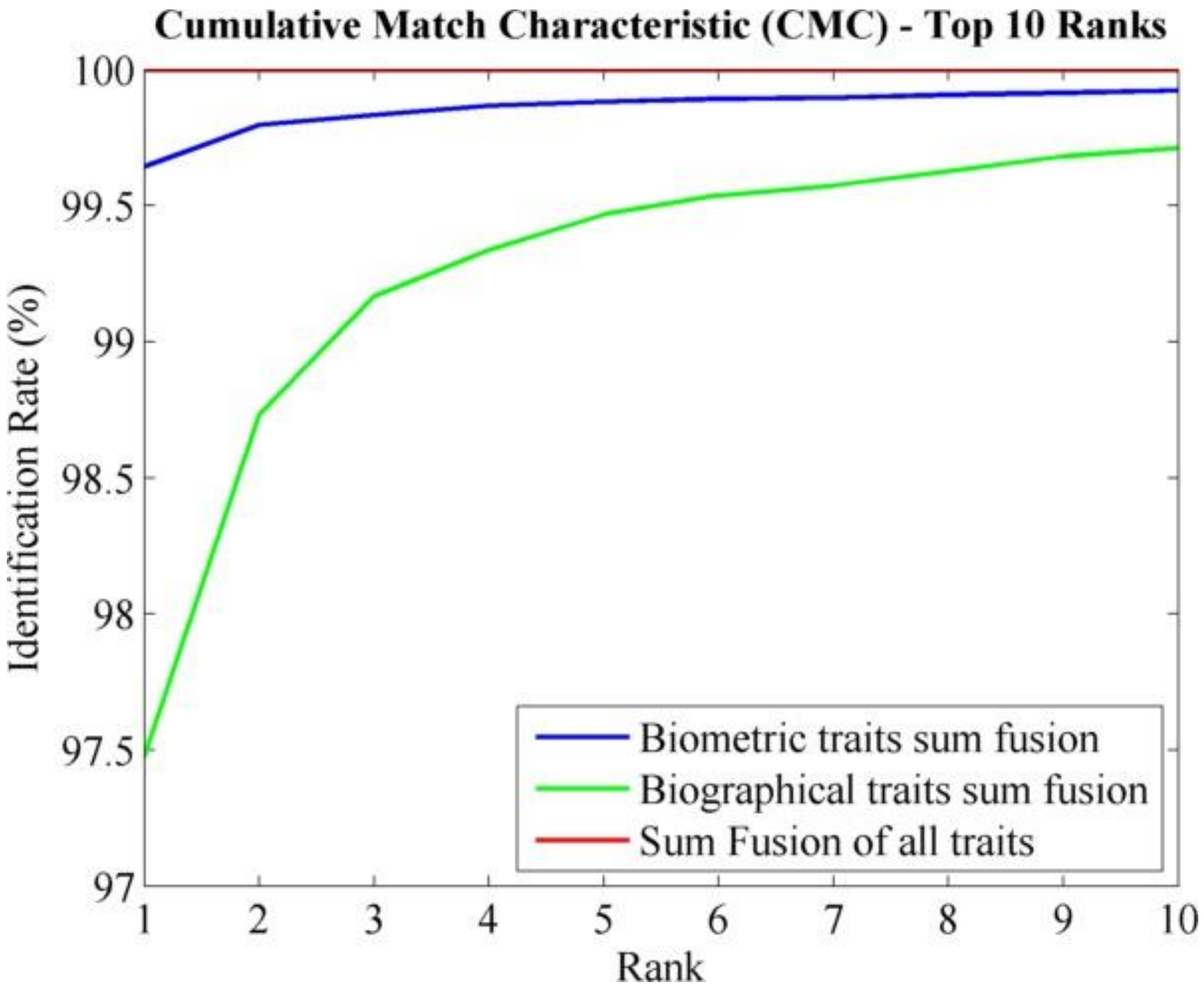

**Fig. 5.7 CMC curve for biometric and biographical fusion.**

Several person recognition systems rely solely on either biographical information or on biometric information of the subjects for de-duplication. A comparative evaluation, where only the biographical information or the biometric information is fused is shown in the cumulative match curve in Fig. 5.7.

| | |
|---|---|
| **LARRY GRYANTHAN**<br>**WARREN GRYANTHAN** | **LARRY GRANTHAN**<br>**WARREN GRANTHAN** |
| **a.** | **b.** |
| **JFFREY RUSSELL**<br>**DEOUGLAS RUSSELL** | **JEFFREY RUSSELL**<br>**DOUGLAS RUSSELL** |
| **c.** | **d.** |

**Fig. 5.8 Examples of successful biographical match where rank-1 scores represent a genuine match.**
**a. Probe name and father's name for biographical example 1; b. Rank-1 gallery name and father's name for biographical example 1.**
**c. Probe name and father's name for biographical example 2; d. Rank-1 gallery name and father's name for biographical example 2.**

Two examples where the rank-1 biographical match (fusion of name and father's name) is genuine are shown in Fig. 5.8, and an example where biographical matching does not retrieve the genuine match at rank-1 is shown in Fig. 5.9.

| | | |
|---|---|---|
| **R SMIT**<br>**M SMTH** | **CARL SMITH**<br>**MARK SMITH** | **ROSEMARIE SMITH**<br>**MICHAEL SMITH** |
| **a.** | **b.** | **c.** |

**Fig. 5.9 Example where biographical match alone is not sufficient.**
**a. Probe name and father's name for biographical example 3; b. Rank-1 gallery name and father's name for biographical example 3; c. Gallery name and father's name of genuine subject for biographical example 3 retrieved at rank 211.**

Person recognition systems often use a single biometric trait along with biographical information of the subjects. The cumulative match characteristic curves for comparative analysis of unimodal and multi-modal biometric fusion with biographical information is shown in Fig. 5.10.

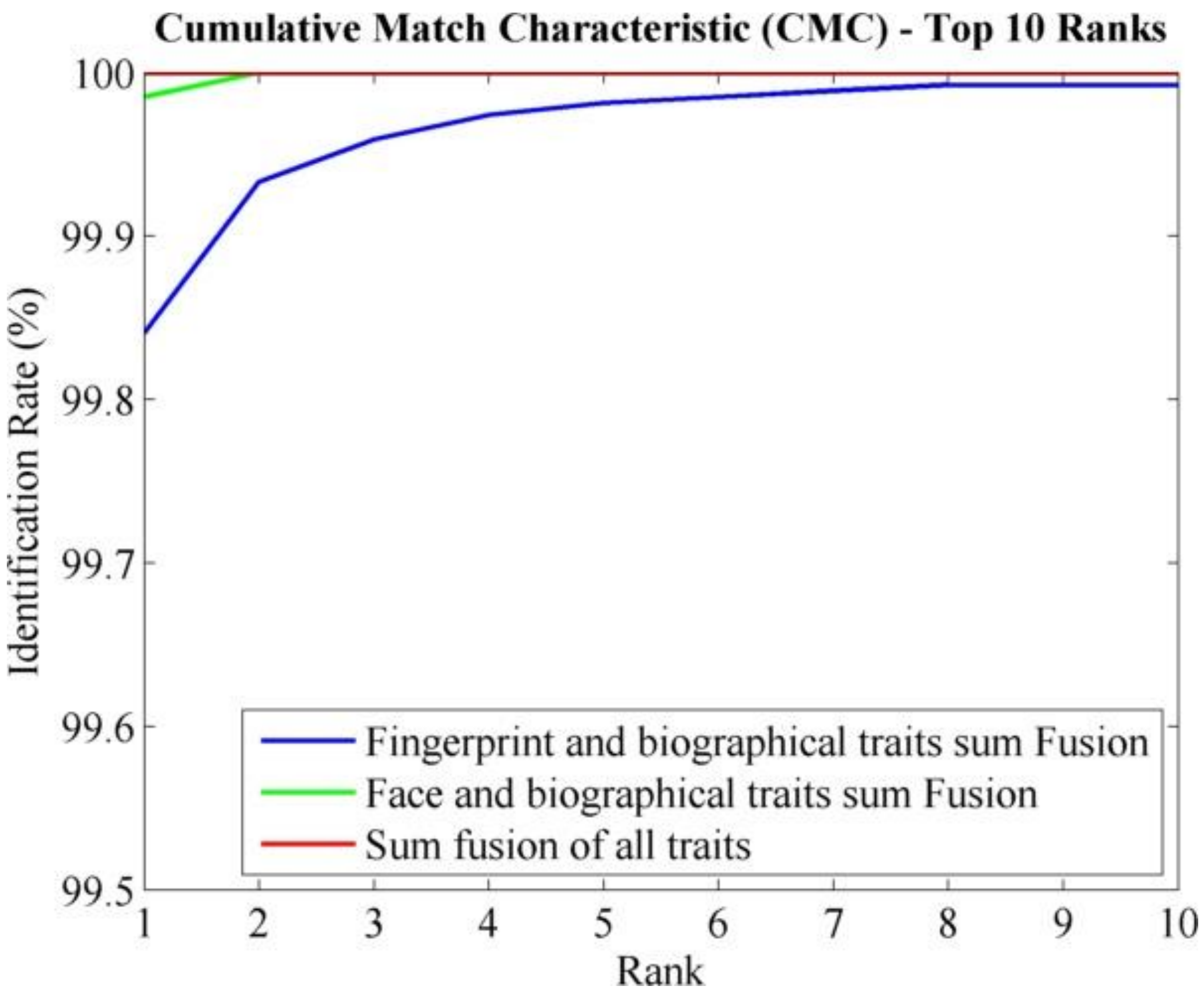

**Fig. 5.10 CMC curve for biometric and biographical fusion.**

## 5.5. Adaptive Fusion of Identifying Information

The 27,000 subjects in the database were randomly partitioned into three subsets (9,000 in each subset) for 3-fold cross validation. One of the 3 subsets was retained for testing the model by turn, and the remaining 2 subsets are used as training data. This 3-fold cross-validation process is repeated 3 times, with each subset used as testing data exactly once. For the present experiment, this three-fold cross validation produced the same results for each fold, so the variance is zero. In general, however, the average results of the three cases may be used for the purposes of reporting. The ensemble consisted of $m$ = 100 classifiers, with top $k$ = 5 highest scores being supplied to each of the classifiers as input.

Based on the results of best discriminability from the training set, the traits were chosen in the sequence of fingerprint followed by face and then finally biographical information. The standard statistical test [141] described in the appendix section A.3 was also implemented for comparison.

The results of the comparative evaluation in terms of efficiency of fusion are summarized in Table 5.1 and also illustrated graphically in graphically in Fig. 5.11, indicating the number of subjects that required the various identifying information to be considered by different fusion algorithms.

**Table 5.1 Comparative Evaluation of Efficiency of Fusion Algorithms**

| Fusion Method | Rank-1 Accuracy | Face Matching and Fusion Required | Biographical Information Matching and Fusion Required |
|---|---|---|---|
| Fusion of all traits | 100.0% | 100.0% | 100.0% |
| Adaptive fusion with single outlier detection using standard statistical test [141] | 100.0% | 47.29% | 18.64% |
| Proposed adaptive fusion algorithm | 100.0% | 36.82% | 8.13% |

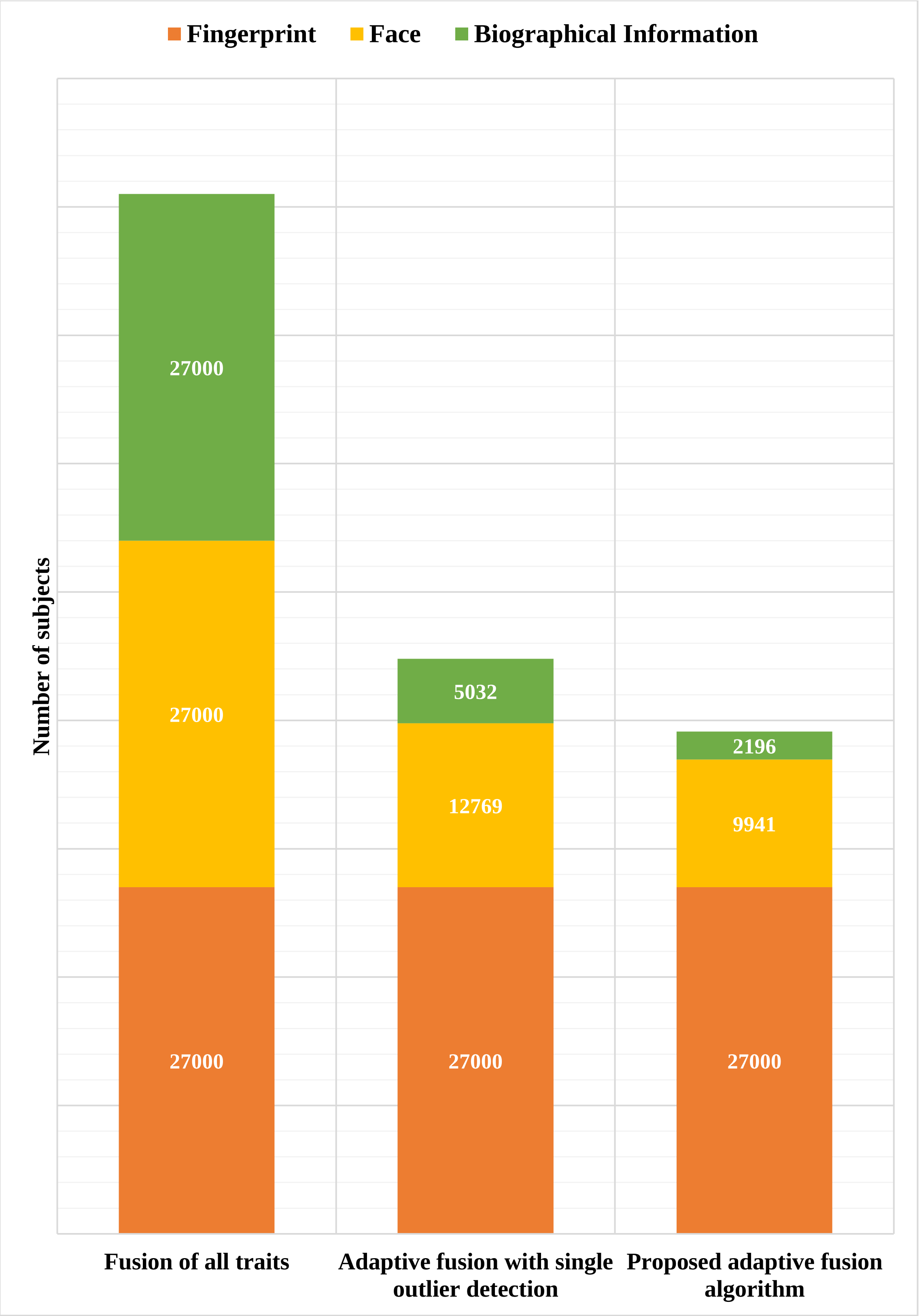

**Fig. 5.11 Comparative evaluation of number of subjects that require matching and fusion of various biometric traits and biographical information, of a total of 27,000 subjects, with a 100% rank-1 accuracy in all cases.**

## 5.5.1. Quality Based Adaptive Fusion

An adaptive fusion approach based on the quality of the fingerprint image was also implemented. The NFIQ algorithm [34] defines five quality levels for fingerprint images, which are interpreted as follows:

i. Level 1: Excellent
ii. Level 2: Very Good
iii. Level 3: Good
iv. Level 4: Fair
v. Level 5: Poor

The lower levels of quality indicate that the matching results may not be reliable. The quality based adaptive fusion approach, therefore, considers additional traits (face and biographical information in this case) for matching and fusion, only if the quality of the acquired probe fingerprint image is below a certain NFIQ level.

It is experimentally observed that of the 27,000 subjects available in the dataset, while quality based adaptive fusion does not require additional identifying information beyond fingerprint to be considered for all subjects, resulting in saving of computational effort, it is not able to achieve this without a deterioration in the accuracy.

**Table 5.2 Comparison of Rank-1 Accuracy and Error at Different Levels of Fingerprint Image Quality in Adaptive Fusion Based on Quality**

| NFIQ Level At and Below Which Additional Identifying Information Beyond Fingerprint is Required | Rank-1 Accuracy | Matching and Fusion of Additional Identifying Information Beyond Fingerprint Required |
|---|---|---|
| Level 1 (Excellent) | 100.0% | 100.0% |
| Level 2 (Very Good) | 99.73% | 68.58% |
| Level 3 (Good) | 99.70% | 63.93% |
| Level 4 (Fair) | 98.03% | 16.71% |
| Level 5 (Poor) | 97.73% | 11.25% |

Table 5.2 illustrates the rank-1 accuracy and the percentage of subjects that require additional information beyond fingerprint to be matched and fused.

While it is evident from these results that a majority of the subjects that require matching and fusion of additional identifying information beyond fingerprint are those for whom the quality of the acquired fingerprint probe image is not of a desirable quality, quality based adaptive fusion has not been considered for the purpose of comparison as it does not afford any computational savings without a deterioration in the rank-1 accuracy.

It is well accepted that for the de-duplication application, accuracy is the predominant objective that may not be compromised for an improvement in computational efficiency.

## 5.6. Predicted Effort to Error Trade-off Curve

While the subject of biometric fusion has widely been studied, there has not been a systematic study of the trade-off between the effort required in determining and fusing matching scores for various traits and the benefit extended by such fusion through reduction in error rates.

A new tool for analysis, the predicted effort to error tradeoff curve to study the relative efficiency of fusion algorithms is therefore proposed. The curve charts the predicted effort as the percentage of subjects where fusion of additional information was predicted to be required, against the error as the percentage of subjects where additional information was predicted to be not required but the rank-1 score did not represent a genuine match.

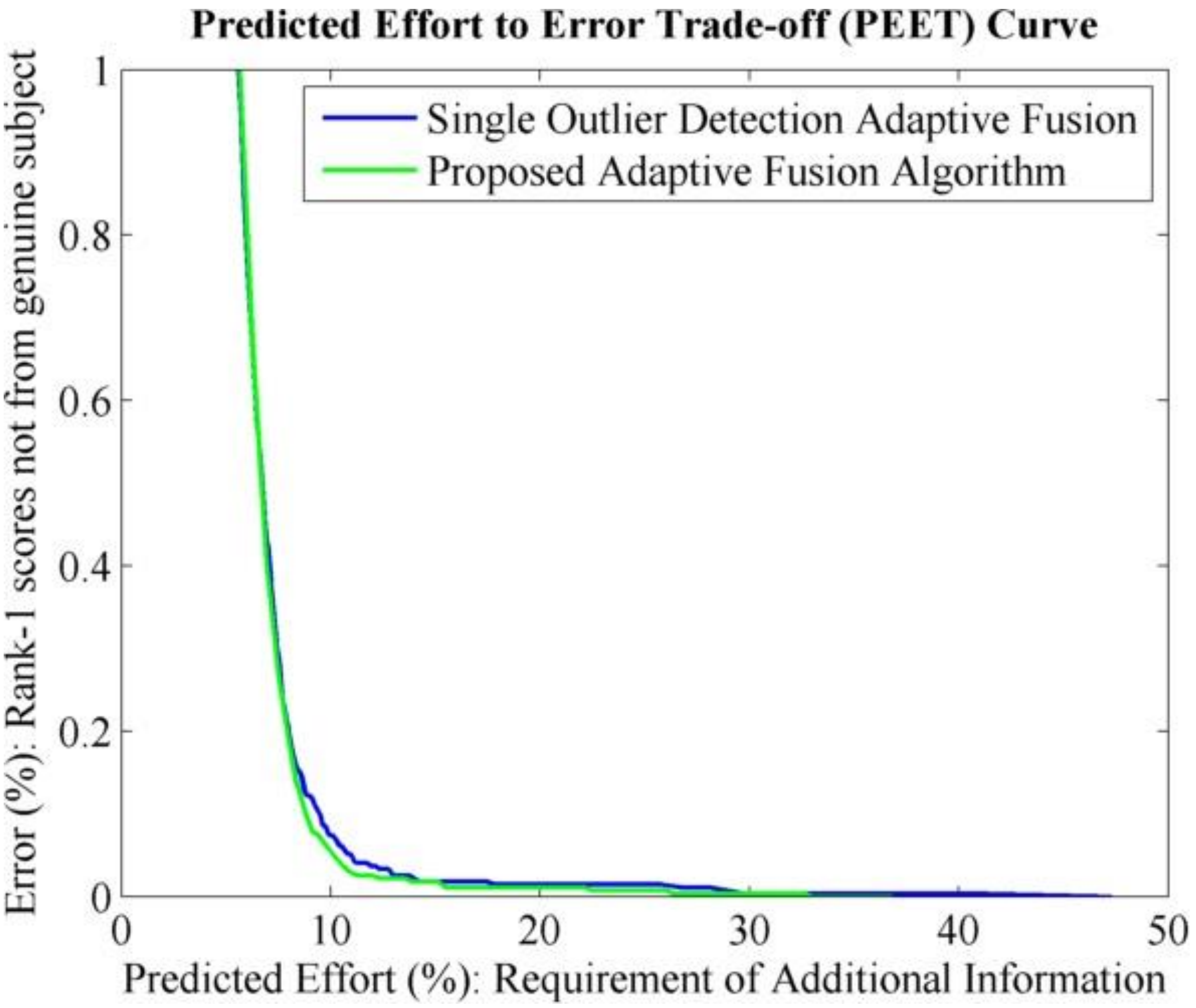

**Fig. 5.12 PEET curve after first stage (fingerprint) comparing the proposed adaptive fusion algorithm with single outlier detection based fusion [141].**

The predicted effort to error trade-off comparing the efficiency of the proposed adaptive fusion algorithm to the adaptive fusion algorithm with single outlier detection using standard statistical test [141] is shown in Fig. 5.12 after the first (fingerprint) stage.

The predicted effort to error trade-off curve after the second (fusion of face with fingerprint) stage is shown in Fig. 5.13.

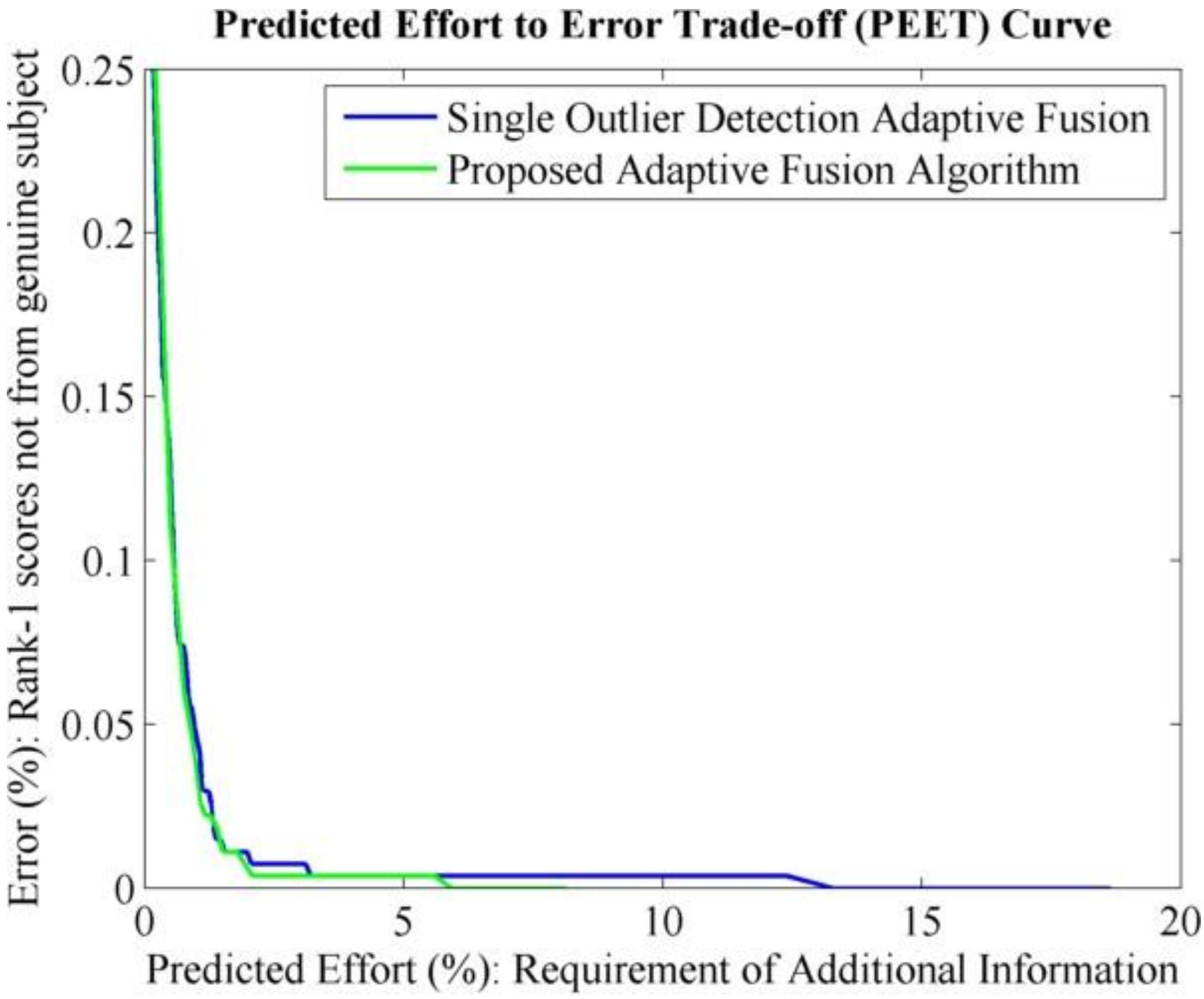

**Fig. 5.13 PEET curve after second stage (fusion of fingerprint and face) comparing the proposed adaptive fusion algorithm with single outlier detection based fusion [141].**

Both Fig. 5.12 and 5.13 indicate that the proposed algorithm converges to the minimum error faster in comparison to the adaptive fusion algorithm based on the detection of a single upper outlier [141].

## 5.7. Summary

A comparative evaluation of different works pertaining to fusion of biometric and biographical information, including the principal results of the present study was summarized in Table 1.1, along with the review of other identity de-duplication systems in section 1.7.

This detailed results of the evaluation of the de-duplication framework and the soft computing model that were proposed in subsections 4.3.2 and 4.3.3 respectively on a large virtual multi-modal dataset of 27,000 subjects, consisting of duplicate instances of fingerprint, face, name and father’s name information for each subject have been described in this chapter. These results include matching accuracy for various unimodal identifiers and multi-modal systems have also been considered. These include fingerprint, face, name and father’s name as the available identifiers, and subsequent fusion by using:

i. only the biometric information,
ii. only the biographical information,
iii. fingerprint with biographical information,
iv. face with biographical information,
v. all biometric and biographical traits, and
vi. the proposed adaptive fusion algorithm

A summary of the rank-1 accuracy for the various identifiers alone and their combinations is presented in Table 5.3.

**Table 5.3 A comparison of identification accuracies of individual biometric traits and biographical information and various combinations.**

| Identifier or Fusion | Rank-1 Identification Rate |
|---|---|
| Fingerprint | 94.93% |
| Face* | 95.56% |
| Name | 89.00% |
| Father’s name | 89.02% |
| Fingerprint + face | 99.64% |
| Name + father’s name | 97.47% |
| Fingerprint + name + father’s name | 99.84% |
| Face + Name + Father’s name | 99.98% |
| Fingerprint + face + name + father’s name | 100.00% |
| Proposed adaptive fusion algorithm | 100.00% |

*** Some face images in the PCSO** [29] **dataset are mislabeled, but have been used to replicate a realistic scenario.**

An evaluation of the algorithm for matching of biographical information using the average impact of edit distances proposed in the previous chapter suggests a marked improvement in comparison to the use of Levenshtein distance.

The results of the adaptive fusion algorithm demonstrate the tremendous savings in computational effort with no deterioration in accuracy.

A quality based adaptive fusion algorithm based on the NFIQ quality levels has also been studied. While this provides useful insights into the conditions when fingerprint may not be a sufficient unimodal identifier, the algorithm is unable to provide savings in computational effort without a deterioration in accuracy.

Finally, a predicted effort to error trade-off curve is proposed. This curve affords a scientific approach for illustrating and comparing the efficiency of fusion algorithms, in relation to the accuracy. A comparison between the proposed adaptive fusion algorithm with the standard statistical single outlier detection based adaptive fusion establishes the superiority of the proposed algorithm.

# CHAPTER 6

# CONCLUSION AND FUTURE WORK

## 6.1. Conclusion

In this thesis, an adaptive fusion framework has been proposed, along with a soft computing based algorithm that considers matching and fusion of additional information available in a multi-biometric system, along with textual or biographical information.

The proposed algorithm does not explicitly require computation of additional characteristics (such as biometric sample quality) and uses information that is inherently computed as part of the de-duplication process. The algorithm is shown to be not only computationally more efficient than adaptive fusion based on discovery of single upper outlier, but also has a higher accuracy over quality based adaptive fusion.

The evaluation of the proposed algorithm in this research has been performed on a reasonably large virtual multi-modal database consisting of two instances of fingerprint, face and biographical information for each of the 27,000 subjects, derived from publicly available datasets. In particular, the proposed system correctly predicts that for 63.18% of the queries, only fingerprint is sufficient. For an additional 28.69% of the queries, both fingerprint and face scores need to be fused and biographical information is needed for only 8.13% of the queries.

With several ongoing attempts towards improvement of sensing devices, advancements in extraction and representation of feature sets and better matching algorithms, it can reasonably be expected that the proposed algorithm would require only a unimodal match for an increasing fraction of the queries. This is evident from the computational savings in the experimental analysis.

The tools and models created as part of this research are intended to be made available as open source resources to facilitate further exploration and development of algorithms towards error-free unique identification of individuals.

## 6.2. Future Work

The study may be extended further using identification databases involving a larger number of subjects (which may not necessarily be in the public domain) for specific applications. Operational databases, such as the Aadhaar Project of India, typically have ten-print fingerprints, along with other biometric traits such as both irises and face besides biographical information [13].

Another avenue for further study would be to incorporate biometric quality in the proposed fusion algorithm. For example, the sequence in which the identifiers are considered for a subject may also be based on the quality of the individual identifiers captured for that particular subject, instead of the globally most discriminating identifier.

### 6.2.1. Soft Computing Techniques for Security of Biometric Systems

The security of biometric data itself is a major concern as unlike possession based or knowledge based tokens, compromised biometric templates may not be revoked or re-issued [146]. The study of enhancement of template security, including creation of cancelable biometric templates with the use of soft computing techniques, is thus another application of paramount importance.

### 6.2.2. Reliable Authentication over a Distributed Network

The performance and accuracy of making the biometric security system available over the network so that a distributed or centralized database can be accessed from several locations is yet another area of research that needs to be addressed. While this would only require a sensor and some means for connecting to the network, such systems would be useful for several applications such as airports, buildings with multiple entrances, ATM machines, etc. An effective solution for this application would need to integrate advancements in both biometrics as well as data networks.

There have been some prior attempts towards addressing this application domain [147]. However, it would be of interest to consider newer and innovative solutions, considering the rapid pace of progress not only in technology development, but also in the establishment of best practices in both biometrics as well as networks.

Since such authentication applications demand accuracy of recognition in addition to real-time turnaround, it would be worthwhile to explore soft computing approaches towards achieving an optimal solution.

The Active Authentication program of the Defense Advanced Research Projects Agency (DARPA) seeks to address authentication over a network "by developing novel ways of validating the identity of the person at the console that focus on the unique aspects of the individual through the use of software based biometrics" [148].

The program initially seeks to research biometrics that do not require the installation of hardware sensors, but instead use “cognitive fingerprint”, based on users behavioral pattern when interacting with the computer, such as the way user handles the mouse and combine available biometrics at a later stage. “The combinatorial approach of using multiple modalities for continuous user identification and authentication is expected to deliver a system that is accurate, robust, and transparent to the user’s normal computing experience” [148]. Since such authentication mechanisms will rely on approximate matches, soft computing approaches would be natural candidates towards solutions in this domain.

Continuous and uninterrupted availability of the network is yet another challenge in the network authentication application, where soft computing techniques may be applied to enhance availability and optimize the network bandwidth, such as through the use of software defined cognitive networks [149].

## 6.3. Summary

The conclusions from research study have been presented in this chapter, with emphasis on the computational savings that the proposed soft computing approach based integrated model for information fusion affords. A comparison drawn from the evaluation of the proposed algorithm against other adaptive fusion approaches has also been reiterated.

The chapter also presents a discussion on possible extension of the study. While a general framework for soft computing based adaptive fusion has been proposed in this study, and evaluated on publicly available benchmark datasets, the framework may be adapted for specific applications, based on the requirements and the available data for specific applications. The integration of additional information that may be obtained from raw biometric samples, e.g., quality, is also proposed to be integrated.

The use of soft computing techniques in related areas has also been proposed, such as for the security of biometric templates and reliable person authentication over a network. The emerging paradigms of active authentication using cognitive fingerprinting, and software defined cognitive networks may also benefit from the application of soft computing approaches.

# APPENDIX

## A.1. Introduction

This appendix provides background information, explanations, reasons and description of heuristics employed in the research, with the intent to elucidate the material presented in the thesis.

## A.2. System Design Considerations

The identity de-duplication system presented in the thesis has a few possible alternatives, leading to the necessity of making some rational design choices, based on a perspective understanding of the problem, as well as empirical estimations.

One of the primary problems that has been addressed in this thesis is to improve the efficiency of the de-duplication process, without any loss in accuracy. Towards achievement of this objective, it is imperative that the system has the capability to predict that at any stage in the de-duplication process, whether the rank-1 score is a duplicate. A framework for achieving this, and a comparative evaluation of the proposed method have been presented in the fourth and the fifth chapters respectively. This section discusses the rationale for choosing an ensemble of veto-wielding logistic regression classifiers over other soft computing paradigms that have been described in the third chapter. The adaptations made in the logistic regression classifiers and the bootstrap aggregating ensemble, to meet the requirements of the application scenario have also been discussed in subsections A.2.1 and A.2.2 respectively.

The process of de-duplication of identities generally consists of several stages of matching and fusing various pieces of evidence that are available as biometric or biographical information. The decision on whether the system has sufficient confidence to terminate matching and fusion of additional information after any stage has been considered as a classification problem.

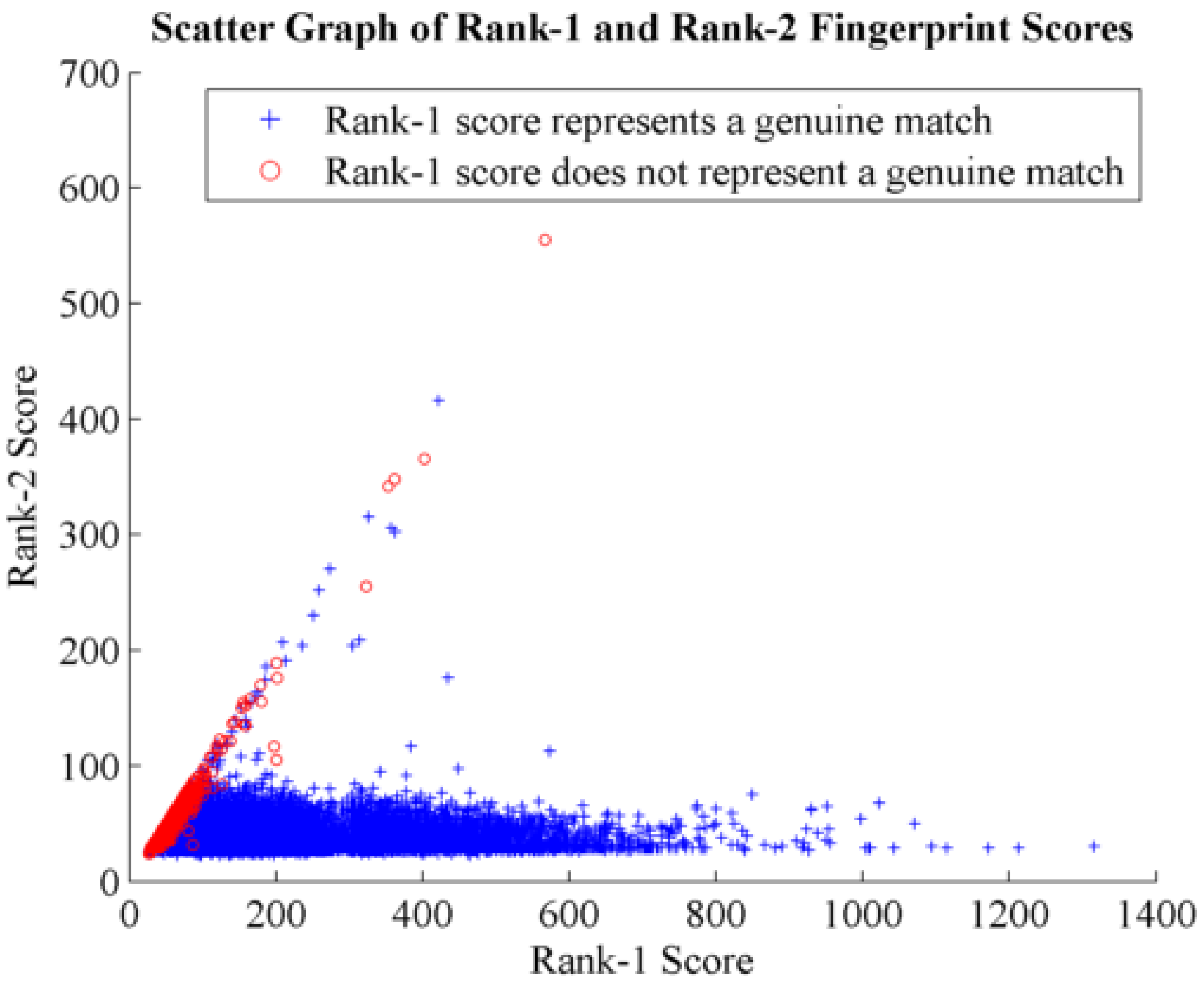

**Fig. A.1 A scatter graph of representation of rank-1 and rank-2 fingerprint scores.**

For a visual representation of a classification problem, the rank-1 and rank-2 fingerprint scores for every individual probe have been presented on a set of orthogonal axes as a scatter graph and the two classes representing whether or not the rank-1 score represents a genuine match have distinctly been marked. While only the top-2 scores have been displayed here for ease of graphical representation, the top-*k* scores, in general, may be used as inputs for the classification problem.

The two-fold objective of improving efficiency while not compromising on the accuracy requires that additional information be considered for matching and fusing if the rank-1 score does not represent a genuine match, while providing preference to accuracy over efficiency. This has been achieved in the proposed method by making suitable modifications to the standard logistic regression classifier and bootstrap aggregating ensemble.

### A.2.1. Adapted Logistic Regression Classifier

A detailed description of the logistic regression classifier and its implementation in the proposed method has been presented in subsections 3.2.1 and 4.3.3 respectively.

When the logistic regression classifier is used for two-class classification, the predicted class is considered to be the one for which the classifier yields a higher probability. Since the class membership in such problems are mutually exhaustive (simultaneous membership of both classes is not permissible) and also mutually exhaustive (the data point must be a member of one of the two classes), this is generally interpreted as the output class being the one for which the probability output is greater than 0.5, thus automatically making the probability of membership in the other class as less than 0.5, as the two probabilities must add up to unity. In other words, the decision threshold in the two-class classification problem when using logistic regression, therefore, is 0.5.

This notion of class assignment in logistic regression has been modified in the proposed method by replacing the decision threshold of 0.5 by a judiciously chosen very low value for the decision threshold $\eta$, to ensure that the process of matching and fusion of additional information is not prematurely terminated, unless the confidence of the classifier that the rank-1 score represents a genuine match is extremely high.

The choice of this adaptation of the logistic regression classifier has also been guided by the general principles of selection of appropriate soft computing algorithm, viz., consideration of simplest model that is least opaque, and provides maximum interpretability. The lack of any hidden layers and a complete interpretation for all inputs and outputs in the model, therefore, make it an ideal candidate for the problem.

Besides, this adapted classifier also provides for better control by allowing tuning of the parameters to suit the requirements of the algorithm for particular de-duplication scenarios, e.g., choice of the operating point in the efficiency to accuracy trade-off through tuning of the decision threshold $\eta$, as described above. The lack of interpretability and the inability to tune operating conditions, therefore, make other soft-computing algorithms less suitable propositions for the robust de-duplication problem.

It was also empirically determined that the adapted logistic regression classification (on whether additional traits are required to be considered), followed by simple sum fusion performs the best, in comparison to other soft computing approaches described in the third chapter.

### A.2.2. Adapted Bootstrap Aggregating Ensemble

The bootstrap aggregation (bagging) meta-algorithm [112] has been described in section 3.3 and the implementation of this technique in the context of the proposed method for adaptive fusion has been discussed in subsection 4.3.3. The choice of the bootstrap aggregating ensemble and the adaptations made to the original meta-algorithm to suit the requirements of a robust de-duplication scenario are discussed here.

It has already been established through prior studies [112, 129] that the use bootstrap aggregation for a soft computing paradigm based systems improves the stability and robustness – both of which are essential ingredients in a large-scale deduplication scenario. In such a scenario, the conflicting requirements of efficiency and accuracy need to be satisfied simultaneously. When multiple biometric identifiers along with biographical information are required to be matched for every query against a large and increasing gallery (pre-enrolled subjects), increasing the efficiency requires some of the computational effort to be reduced, while the accuracy requirement demands that no individual be allowed to enroll more than once in the system, even if that leads to an increase in the computational effort.

For most de-duplication scenarios, as stated in subsection 4.3.3, accuracy is a higher objective over efficiency and a bootstrap aggregating ensemble of adapted logistic regression classifiers (described in subsection A.2.1) has been included in the design of the proposed system as the desirable qualities imparted by bagging suitably fulfill the requirements of stability and robustness. The results and comparative analysis presented in the fifth chapter also validate this notion, as the proposed method has been demonstrated to achieve a significant improvement in efficiency, without any loss in accuracy.

In general, when bagging [129] is used for regression problems, the final outcome may be based on the average value of the regression output, and when this meta-algorithm is used for the two-class classification problem, the class that has been predicted by a majority of the classifiers is considered as the output of the ensemble. This usage of bagging has been modified and adapted for the proposed method to keep accuracy as a higher objective over efficiency. Hence, instead of considering the majority opinion on whether additional information is required to be considered, the process of matching and fusion for additional identifiers is terminated only if all classifiers are unanimous in their opinion that the rank-1 score conclusively represents a genuine match. Thus, every individual classifier in the ensemble wields a *veto* [139], and the composite may therefore appropriately be referred to as a bootstrap aggregating ensemble of veto-wielding logistic regression classifiers.

## A.3. Deterministic Adaptive Fusion based on Upper Outlier Detection

A deterministic adaptive sequential fusion strategy has been proposed [141] based on the detection of presence of an "upper outlier" in the similarity score distribution, originally in the context of latent fingerprint matching, under the assumption that these scores follow an exponential distribution.

An outlier is an observation (matching score in the current setting) which appears to deviate markedly from other members of the sample. An upper outlier is a significantly higher value than the rest of the distribution. An outlier may be an "extreme manifestation of the random variability inherent in the data" or "may be the result of gross deviation from prescribed experimental procedure or an error in calculating or recording the numerical value" [150]. In statistical analysis, while outliers are usually rejected for the latter reason, an upper outlier may also be considered to be the outcome of the former phenomenon, and hence, a strong indicator that the matching score is from a true mate.

Intuitively, the presence of a single upper outlier is a strong indication of a true mate because of the abysmally low probability of two events occurring simultaneously, videlicet, a false match generates such high matching score that it is far removed from the rest of the similarity score distribution, and the true match generates such low score that it is within the distribution.

It is known that biometric matchers have predefined similarity functions, that may easily be determined by fitting a parametric curve to the score distribution histograms [151, 152] and a goodness-of-fit to the distribution may then be evaluated using the chi-square test [153]. Such parametric distributions generally have well defined methods for determination of outliers.

After calculating the matching scores for a certain trait (say, right index finger) against background database, a decision on whether or not additional pieces of evidence may be required is based on the presence of a single upper outlier in the similarity score distribution [141].

Many de-duplication systems use ten-print fingerprints, possibly with other biometrics and along with soft-biometric and biographical information [39, 42, 43, 44]. It has been determined in [141] that fingerprint matching scores from various COTS usually follow an exponential distribution. The decision on whether additional traits are required to be considered may therefore be based on the following stopping criteria, based on the presence of an upper outlier in the score distribution using order statistics [141].

i. Sort the matching scores in descending order

ii. Subtract the second highest score from the highest score and divide this difference by the sum of top $n$ scores, where, $n$ is indicative of approximation of the score distribution to the exponential similarity function. This is the test statistic z.

iii. Calculate the critical (threshold) value $z(\alpha)$ as follows:

$$z(\alpha) = 1 - \alpha^{\frac{1}{n-1}} \tag{23}$$

where $\alpha$ is the significance level, the probability of a non-outlier being incorrectly labelled as an outlier.

iv. Further pieces of evidence are not required if $z > z(\alpha)$, as the test statistic $z$ indicates if the highest score is an upper outlier in the score distribution for that trait.

For achieving computational speed-up in evaluating against this stopping criteria and also for better approximation, $n$ may be made large to include all scores, reducing the sorting the sorting step above to two searches of the highest and the second highest scores.

The significance level $\alpha$ decides the trade-off between computational effort and accuracy. A smaller value of $\alpha$ raises the bar on the stopping criteria, requiring a greater number of traits to be merged, leading to increased computational expense in calculating and fusing matching scores for additional traits.

The traits are considered starting with the one believed to be the most reliable for a large population, followed by the next most reliable, and so on, in that sequence. Starting with the most reliable biometric has the advantage that for a large section of the population, second or subsequent traits may not be required to be considered. However, the biometric trait that is the most reliable for a large population may not be the best for every individual. Therefore, as additional traits are considered for certain individuals, when the first trait fails to provide sufficient evidence.

A caveat in this method, however, is that it is not always possible to reliably determine a parametric distribution for the scores. Further, not all parametric score distributions are amenable to determining a single upper outlier.

A flowchart of a de-duplication algorithm based on the detection of a single upper outlier in the match score distribution is shown in Fig. A.2.

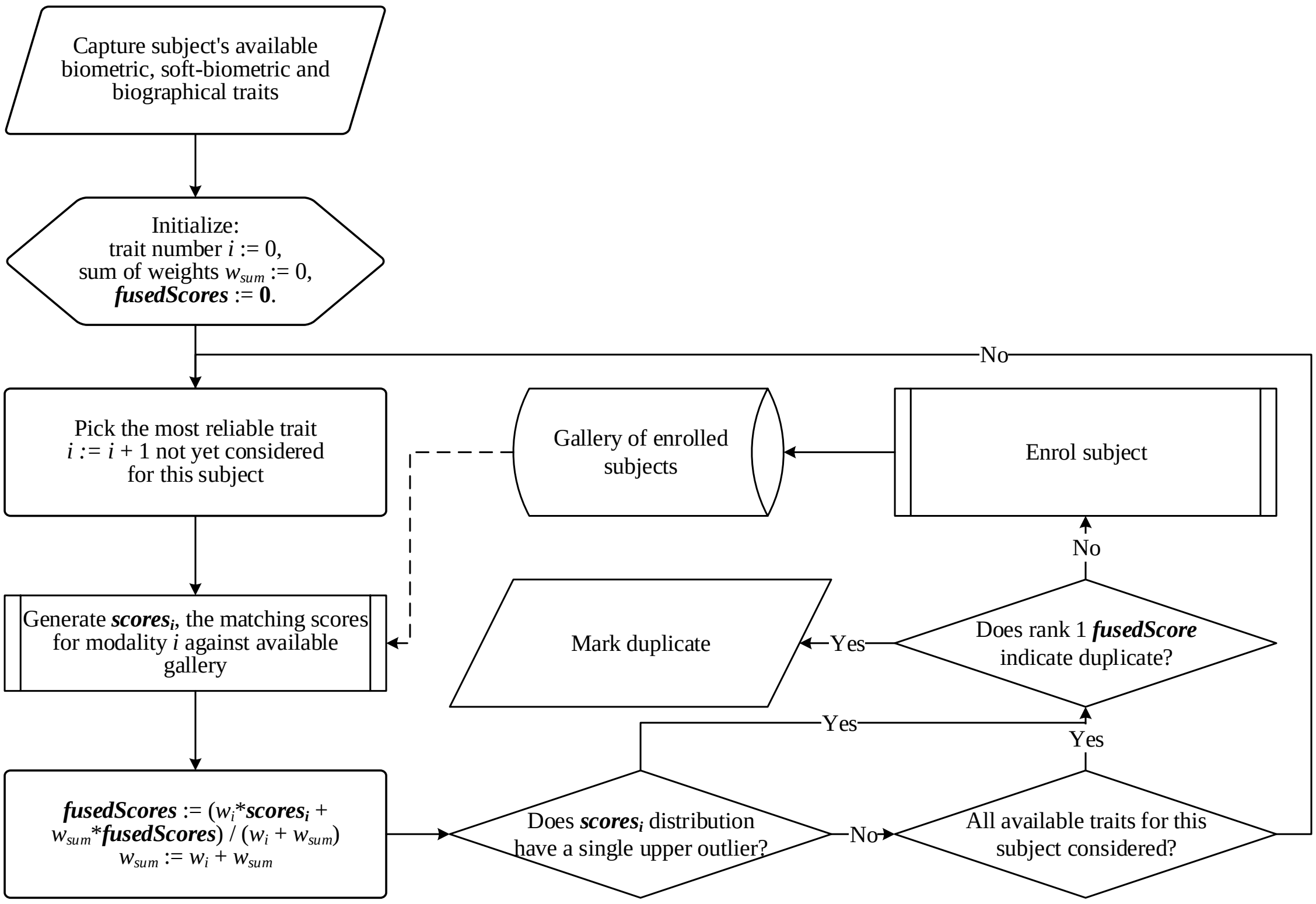

**Fig. A.2 A de-duplication algorithm based on the detection of a single upper outlier in match score distribution.**

## A.4. Evaluation and Comparative Analysis

A comprehensive discussion on the evaluation of the adaptive integrated soft computing model proposed in the fourth chapter has been presented in the fifth chapter, not only in terms of accuracy of the proposed system, but also in terms of computational savings. Some additional discussions on computational complexity and robustness of the proposed model are presented here.

### A.4.1. Computational Complexity

The proposed method improves the time required for fusion, by reducing the number of queries for which scores for additional traits are required to be computed and fused. The improvement in computational efficiency is because of reduction in this time required for fusion. Even in the worst case, when the matching scores for all traits for all users need to be computed and fused, the time complexity is of the order of number of traits times the number of enrolled users, which is the same as the time complexity of approaches that the proposed approach has been compared to. Even for the general scenarios (better than the worst case), the efficiency of the proposed approach continues to be maintained, because of the reduction in the number of queries for which scores for additional traits are required to be computed and fused.

From an analysis perspective, the proposed approach, which determines whether additional traits are required, followed by fusion does not increase the time complexity in comparison to the other approaches (that compute and fuse matching scores for all traits for all users). It may be argued that the time complexity of the order of number of enrolled subjects achieved by the proposed approach is lower than the time complexity for any known de-duplication system, as has also been illustrated in Table 1.1.

To emphasize the importance of efficiency in fusion of biometric and biographical information, a new metric (Predicted effort to error trade-off curve) is has been proposed in section 5.6, and the performance of the proposed system shown in Fig. 5.12 and Fig. 5.13.

### A.4.2. System Robustness

Multi-biometric systems, in general, are expected to be more reliable due to the presence of multiple, independent pieces of evidence. These systems deter spoofing and impart fault tolerance to biometric applications, because while it may be easy to deceive the system through spoofing one biometric trait by creating an imitation or a replica, this becomes increasingly difficult when multiple pieces of evidence are involved.

The adaptive fusion proposed in this thesis should be used with caution, as the method may not consider all pieces of evidence, and may terminate the process of matching of additional identifying information and fusion, potentially even after the first identifier has been considered. It is, therefore, not recommended that adaptive fusion be used without considering the associated pitfalls for the specific application scenario.

The method, however, poses minimal negative consequences for the identity de-duplication scenario that it has been proposed for. In the process of de-duplication during enrolment, the motivation for the individual subject is to not be identified as pre-enrolled or duplicate, as that would jeopardize their chances for obtaining benefits from public welfare schemes or for escaping harsher punishment as a repeat offender. The lack of a conceivable incentive for the subject to spoof the system in a de-duplication scenario renders the proposed method suitable for this application. It is also noteworthy that the method terminates the matching and fusion process only when a duplicate is positively determined as a rank-1 match, and continues with the matching and fusion of all available identifiers otherwise.

Bootstrap aggregation (bagging) improves the stability and accuracy of soft computing algorithms and may be used for both classification and regression [129]. Bagging also reduces variance and helps to avoid overfitting, this affording robustness to the system. An adapted version of this meta-algorithm, described in subsection A.2.2 has therefore been used to ensure a robust system design. The logistic regression classifier has also been adapted through a judicious choice of the decision threshold, in favor of accuracy over efficiency, thus imparting even greater robustness to the system.

## A.5. Publications

A comparative review of information fusion approaches for multi-biometric security systems, and a succinct summary of the framework, algorithm, significant results from this research and their comparative analysis have been published as peer-reviewed journal articles in [154, 155].